\documentclass[10pt,twocolumn,letterpaper]{article}

\usepackage[pagenumbers]{cvpr} 

\usepackage{microtype}

\renewcommand{\paragraph}[1]{\vspace{.5em}\noindent\textbf{#1.}}

\usepackage{multirow}
\usepackage{pgfplots,pgfplotstable}
\usepgfplotslibrary{groupplots}
\pgfplotsset{compat=1.18}
\usepackage{xcolor}
\usetikzlibrary{patterns,patterns.meta}
\usepackage{amsmath,array,booktabs,longtable,url}
\usepackage{tcolorbox}
\usepackage{subcaption}
\usepackage{siunitx}
\usepackage{float}
\usepgfplotslibrary{groupplots}

\newcommand{\xsmall}{\fontsize{7}{8}\selectfont}

\definecolor{cvprblue}{rgb}{0.21,0.49,0.74}
\usepackage[pagebackref,breaklinks,colorlinks,allcolors=cvprblue]{hyperref}

\def\confName{CVPR}
\def\confYear{2027}

\title{Off the Driven Path: Transferable Road Understanding in Driving VLAs
}

\author{
Lena Wild$^{1,4}$ 
\quad Christian Helten$^{2}$
\quad Katie Z Luo$^{3,6}$
\quad Jakob Thumm$^{3,5}$ \\
\quad Rafael Valencia$^{4}$
\quad Patric Jensfelt$^{1}$
\quad Marco Pavone$^{3,7}$  \vspace{5pt}\\
$^{1}$KTH Royal Institute of Technology ~~\quad $^2$Technical University of Munich ~~\quad $^3$Stanford University \\ $^4$TRATON ~~\quad $^5$scaledrive.ai~~\quad $^6$Waymo~~\quad $^7$NVIDIA
}

\begin{document}
\definecolor{_lightteal}{HTML}{ABD4D4}
\definecolor{_teal}{HTML}{52A1A3}
\definecolor{_lightgreen}{HTML}{ADDED0}
\definecolor{_green}{HTML}{50B99B}
\definecolor{_lightred}{HTML}{EA7C93}
\definecolor{_red}{HTML}{AF1D3C}
\definecolor{_lightyellow}{HTML}{FAE097}
\definecolor{_yellow}{HTML}{F3B816}
\definecolor{_lightorange}{HTML}{F69C7F}
\definecolor{_orange}{HTML}{D23F0F}
\definecolor{_lightpurple}{HTML}{B58FBB}
\definecolor{_purple}{HTML}{84488E} 
\maketitle
\vspace{-2mm}
\begin{abstract}
End-to-end driving policies increasingly operate without explicit HD maps at inference, raising the question of what road knowledge they actually internalize.
To probe this knowledge through model outputs, we introduce Road-KSU, a framework that automatically converts HD maps and actor annotations into a road-centric scene graph.
From this, we derive road-focused question answering and 
reasoning targets, complemented by map-based trajectory compliance metrics. 
Pairing Road-KSU with geographically disjoint splits yields a $2 \times 2$ evaluation crossing output modality (language vs. action) with place familiarity (seen vs. unseen roads).
On NAVSIM, 95\% of test samples lie within 10\,m of a training sample, and scores of 13 published planners on turns rise steeply with the number of training drives through the test location, pointing to memorized road geometry. 
Retraining on geographically disjoint splits and probing with Road-KSU confirm that action-space performance degrades most on unfamiliar turns, with failures dominated by geometric compliance. Language accuracy, however, drops on road-centric VQA on straight and curved roads alike, and remains stable on dynamic scene content.
Scaling the training data further reveals a double dissociation: more drives through the same places improve driving on unseen roads but leave road question answering unchanged, whereas spreading a fixed data budget over more places increases road question answering with little change in driving performance. 
These findings show that improved driving scores need not imply broader road understanding on unseen roads, and identify geographic diversity as key to transferable road knowledge.
\end{abstract}    
\vspace{-8mm}
\section{Introduction}
\label{sec:intro}
End-to-end driving models, which jointly optimize perception and planning, have made substantial progress on open-loop driving benchmarks~\cite{casas2021mp3, hu2023planning, navsim, bojarski2016end}. 
Recent advances in vision-language models (VLMs) \cite{liu2023visual, Qwen3-VL} have extended this approach to driving vision-language-action (VLA) models, which reason about the scene in language space and directly return driving trajectories~\cite{qwendrive, nvidia2025alpamayo}.
At the same time, many end-to-end models no longer receive explicit road representations, most prominently high-definition (HD) maps, at inference time \cite{hu2023planning,jiang2023vad}. While such maps provide detailed road geometry, lane topology, and traffic rules, relying on them during deployment requires maintaining accurate maps wherever the vehicle operates~\cite{bateman, wild2025argotweak}. Removing this inference-time dependency offers practical advantages but raises a fundamental question: \textit{Can a model infer the road structure it needs from the current driving scene alone?}

Whether end-to-end driving models acquire such internal road representations is inherently difficult to establish, since implicit knowledge cannot be directly inspected.
An alternative is to study \emph{road understanding} through model behavior, which we define as the \textit{observable} ability of a model to identify, describe, and act upon properties of the road.
Driving VLAs~\cite{nvidia2025alpamayo, qwendrive} provide a particularly useful setting, as their language outputs reveal which road properties they can report, while their predicted actions show whether their driving complies with road geometry and rules.

These two \emph{output modalities} --- language and action --- define the first dimension of our study. The second is \emph{place familiarity}: at locations seen during training, a model may memorize geometry or use location-specific priors, whereas geographically held-out places test its ability to infer road properties in unseen scenes \cite{lilja2024localization, yuan2024streammapnet}. Crossing \emph{language versus action} with \emph{familiar versus unfamiliar roads} yields a $2\times2$ evaluation space that lets us ask where road understanding appears, how it transfers to new places, and how each output modality responds to training data and supervision.

To this end, we introduce \textbf{Road-KSU}, a framework that brings HD maps as a source of explicit road information back into the development pipelines of driving VLAs. In training, our map-derived supervision encourages models to internalize road information. Map-derived evaluation subsequently allows us to assess if driving VLAs identify, express, and act upon it, without requiring HD map coverage at deployment. 
At its core, Road-KSU automatically transforms HD maps and actor annotations into a road-centric scene graph that links geometry, topology, traffic rules, and actors to a common set of road entities. From this shared representation, we derive two language probes: \textbf{KNOW} probes the broader road context through VQA, including surrounding road structure and maneuvers beyond the current navigation instruction; \textbf{SAY} provides reasoning targets focused on road properties relevant to the immediate driving task. In action space, \textbf{USE} scores whether predicted trajectories comply with road geometry and rules, extending the rule-based EPDMS score \cite{navsimv2} with speed-compliance computed against the same HD maps. To control for place familiarity, we introduce geographically disjoint train and test splits to distinguish performance on familiar roads from transfer to unseen locations. We additionally use an answer-balanced VQA protocol to test whether models recognize less common road configurations rather than relying on memorizing the most frequent answers.

Our experiments reveal that geographic familiarity affects language and action differently. On NAVTEST~\cite{navsim}, where 95\% of test samples lie within 10\,m of a training sample, the scores of 13 published planners correlate with the number of training drives through the test location on turns, while the human driver's scores barely change. Retraining on our geographically held-out splits and evaluating with Road-KSU shows the same pattern, with failures dominated by departures from the drivable area. The other terms, including USE’s speed checks, fail at similar rates on seen and unseen roads, pointing to road geometry rather than traffic rules as what models fail to transfer. 

Road question answering, on the other hand, degrades on both straight and curved roads, and even more under answer-balanced evaluation, while dynamic-scene accuracy remains largely unchanged. Notably, the sensitivity to answer balancing is not specific to our models: human annotators recognize rare road configurations as reliably as common ones, whereas all tested general-purpose and driving VLMs lose accuracy under answer balancing.
Together, these results indicate that models may drive successfully yet fail to identify broader road properties or non-standard road configurations that could matter beyond nominal driving.

To further investigate whether language and action benefit from different training data, we perform controlled scaling experiments. 
Increasing the number of drives through the same training locations improves driving performance on unseen roads, but yields little gain in road-centric VQA, whereas both improve with repetition on seen roads. Conversely, spreading a fixed data budget over more places improves VQA performance on unseen roads while leaving driving performance largely unchanged. Together, these results identify geographic diversity as an important factor in transferable road understanding, showing that improvements in driving scores need not imply corresponding gains in broader road knowledge. Our contributions are threefold:
\begin{itemize}
\item We introduce \textbf{Road-KSU}, a framework that grounds road question answering (\textbf{KNOW}), task-focused road descriptions (\textbf{SAY}), and trajectory-compliance evaluation (\textbf{USE}) in shared HD-map and scene annotations, enabling the study of both immediately driving-relevant knowledge and broader road understanding across language and action.
\item We establish geographically held-out and answer-balanced evaluation protocols for transferable road understanding, to distinguish performance on seen roads from generalization to new locations and test recognition of diverse road configurations beyond frequent-answer priors.
\item Through controlled data-scaling experiments, we characterize how road understanding develops across language and action, identifying visitation count and geographic coverage as complementary axes of data scaling.
\end{itemize}

\section{Related work}
\label{sec:related_work}

\paragraph{End-to-end driving and VLAs}
End-to-end driving models replace modular pipelines with a single network that maps sensor input to actions~\cite{bojarski2016end,chitta2023transfuser,chen2024end}.
Increasingly operating without HD maps at inference~\cite{hu2023planning,jiang2023vad}, they must infer road structure from sensor observations.
Building on the end-to-end paradigm, driving VLAs add a VLM backbone that emits language alongside actions~\cite{hwang2025emma,renz2025simlingo,nvidia2025alpamayo,qwendrive}, while a parallel line of work uses auxiliary language or scene supervision only during training and reports planning gains~\cite{pan2024vlp,hegde2025distilling,xu2025vlm,schmidt2025graphpilot,chen2026geo}.
Yet the language VLAs emit need not inform their actions, as generated explanations can diverge from their trajectories~\cite{foutter2026vision,nguyen2026vladrivebench} and VLAs without textual reasoning remain competitive~\cite{rawal2026nord,song2026more,li2026recogdrive}. Rather than assuming that one output modality explains the other, we treat
language and action as separate probes of the road understanding a model has
internalized.

\paragraph{Benchmarking driving VLAs}
Prior work probes and supervises driving knowledge in language space through VQA and reasoning targets that are written by annotators~\cite{marcu2024lingoqa}, generated by LLMs or VLMs~\cite{park2025nuplanqa,xu2025vlm}, or instantiated from scene annotations~\cite{qian2024nuscenes,nie2024reason2drive,wang2025embodied}.
Road-specific benchmarks find VLMs weak on static road structure~\cite{cao2024maplm,liu2026roadscenebench,zhang2026roadbench,chen2026are, bridging2026wild}, but evaluate language only.
In action space, nuPlan and NAVSIM aggregate road compliance, progress, and comfort into rule-based scores such as the predictive driver model score (PDMS) and its extension EPDMS~\cite{nuPlan,dauner2023parting,navsim,navsimv2}.
Joint benchmarks evaluate both output modalities~\cite{sima2024drivelm,wang2025omnidrive,nuReasoning,jia2026bench2drivevl}, yet cover general driving knowledge and score language and action against different sources~\cite{xu2026drivehierarchy,chembu2026causaldrivebench}. 
No real-world benchmark investigates road-specific knowledge across language and action space, deriving road questions, reasoning targets, and trajectory compliance from a common ground truth.

\paragraph{Geographic generalization and data scaling}
Common driving benchmarks contain repeated traversals of the same roads in training and test
data~\cite{caesar2020nuscenes,Argoverse2,nuPlan,navsim,nuReasoning}. This overlap is exploitable: on NAVSIM, a planner that retrieves features
from nearby training drives nearly matches camera-based methods~\cite{lowens2026driving}, revealing that NAVSIM demands little interaction with other actors. What such overlap means for the road knowledge models acquire has been
studied only for online mapping, where geographically disjoint splits
roughly halve accuracy on nuScenes~\cite{lilja2024localization,yuan2024streammapnet}.
On the scaling side, generalization in reinforcement and imitation learning grows with the number of distinct training environments~\cite{cobbe2019quantifying,li2023metadrive,lin2025data}, whereas driving scaling studies vary data volume~\cite{naumann2025data,zheng2026data,codevilla2019exploring,dimlioglu2026scaling} and do not separate revisiting a road from visiting a new one.
We fill these gaps by proposing geographically held-out splits, then vary visitation and geographic coverage.

\section{The Road-KSU framework}
\label{sec:method}

To probe and improve road understanding, we construct \textbf{Road-KSU}, a framework that automatically derives road-centric supervision and evaluation in both action and language space from shared ground-truth scene annotations (\Cref{fig:method}). To this end, we operationalize HD maps and actor annotations into a road graph (\Cref{sec:graph_construction}), from which we generate VQA pairs (\textbf{KNOW}) and reasoning targets (\textbf{SAY}). For action space (\textbf{USE}), we extend rule-based driving metrics \cite{navsim, navsimv2} for speed compliance (\Cref{sec:graph_execution}). We then instantiate Road-KSU on nuPlan \cite{nuPlan}, and propose geographically disjoint splits and an answer-balanced VQA protocol in \Cref{sec:instantiation}.

\subsection{Graph construction}\label{sec:graph_construction}
The graph representation at the heart of \textbf{Road-KSU} serves two purposes: It provides the shared ground truth for KNOW, SAY and USE, and bridges two descriptions of the driving scene that do not naturally align: geometric primitives of the HD map and actor annotations with their numerical and vectorized representation, and the road concepts expressed in natural language. A \texttt{roadblock} or \texttt{lane\_group}, \eg, is a vectorized HD map primitive, whereas a language expression may refer to \emph{``the road ego is on''} or \emph{``the lane to my left''}. 
Querying the map directly would leave each probe to resolve these references independently. 
The graph resolves such references into semantically meaningful nodes and their mutual edge relations aligned with natural language instead, with links to their map primitives as evidence. Reusing these resolved entities and the HD map across probes, \textbf{Road-KSU} provides a consistent basis for generating and evaluating road knowledge in both language and action space.

\paragraph{Graph definition} We define a \emph{decision point} at $T=0$ as the time from which on a driving model is asked to predict the future trajectory. At each decision point, the graph builder $\mathcal{B}$ constructs a directed, attributed scene graph,
\begin{equation}
    G_0=\mathcal{B}(\vec{p}_\text{ego},\{\vec{p}_\text{i}\}_{i=1}^{n_\text{actor}},\{L_\text{j}\}_{j=1}^{n_\text{tl}}, M, \nu_0)=(V_0,E_0). 
\end{equation}
Here, $\vec{p}_\text{ego}=\{x_\text{ego},y_\text{ego},\theta_\text{ego},v_\text{ego}\}_{T_{-n}}^{T_{0}}$ is the current ego state and its historical values over the past $n$ timesteps, $\{\vec{p}_\text{i}\}_{i=1}^{n_\text{actor}}$ is the set of actor annotations over the same horizon, while $\{L_\text{j}\}_{j=1}^{n_\text{tl}}$ is the set of traffic-light states at $T=0$. $M$ is the HD map in a window of $[(-\Delta x/2, +\Delta x/2),(0, +\Delta y)]$ around the ego pose in forward heading direction and $\nu_0$ is the ego route. 
In $G_0$, each node $v\in V_0$ has a type drawn from a fixed node-vocabulary $\mathcal V$, and each edge $(u,r,v)\in E_0\subseteq V_0\times\mathcal R\times V_0$ represents a typed relation $r$ in a fixed edge-vocabulary $\mathcal R$. Note that $G_0$ is a snapshot at $T{=}0$: it includes the current state of signals and the current kinematics of actors, but no future states -- the $n$ timesteps of position history provided to $\mathcal{B}$ are used to resolve map-matching ambiguities.

\input{figures/method}
\paragraph{From map units to road concepts}
The graph builder $\mathcal{B}$ transforms HD map primitives and actor annotations into ego-centric road nodes and their semantic relations. Graph construction proceeds in three steps:
\begin{enumerate}
\item \emph{Anchor.} We identify the ego’s current lane using polygon containment, heading agreement, and recent motion history, thereby establishing the graph's reference node.
\item \emph{Expand.} We traverse the local HD map $M$ to identify road nodes from map primitives (\eg., intersections) and their properties (\eg, control type), while traffic controls and annotated actors are added with their current states and kinematics. Each node retains type-specific attributes and references to its source geometry.
\item \emph{Relate.} Finally, we combine map connectivity, geometry, and spatial context to derive topological (\eg, \texttt{LEADS\_UP\_TO}), spatial (\eg, \texttt{IS\_WITHIN}), and functional (\eg, \texttt{GOVERNED\_BY}) edges that are meaningful to a driver but are not necessarily explicit map elements. 
\end{enumerate}
The node- and edge-vocabularies $\mathcal{V}$ and $\mathcal{R}$ are closed sets fixed within $\mathcal{B}$. Their definitions and the complete construction rules are provided in Appendix \cref{sec:graph_construction_details}.

\subsection{Graph execution}\label{sec:graph_execution}
Once the graph is constructed, KNOW, SAY, and USE apply component-specific rules and generate questions, reasoning targets, and compliance metrics, respectively.

\paragraph{KNOW: Visual Question Answering}
For testing the models' knowledge about relevant road properties, we follow \cite{bridging2026wild} and define graph queries $\mathcal{Q}_i$ over $G_0$, 
\begin{equation}
\begin{aligned}
     \mathcal{Q}_i &= \{\mathcal{S}_i, \mathcal{T}_i^q,\mathcal{T}_i^a\}, \\
    q_i &= \mathcal{T}_i^q(\mathcal{S}_i(G_0)), \quad
    a_i = \mathcal{T}_i^a(\mathcal{S}_i(G_0)),
\end{aligned}
\end{equation}
where the selection operator $\mathcal{S}_i$ retrieves the set of target nodes and edges, the question template $\mathcal{T}_i^q$ maps the targets to a question in natural language $q_i$, and the answer template $\mathcal{T}_i^a$ returns the correct answer $a_i$ from the graph. In total, we define 40 graph queries covering lane geometry and topology, permitted maneuvers, intersections and traffic controls, crossings, and nearby actor positions; the complete set is provided in Appendix \cref{sec:question_templates_know}.

\paragraph{SAY: Road-Grounded Reasoning}
For task-focused road descriptions, we define a set of $c$ ordered clauses $\mathcal{C}_j$ over $G_0$ and ego route $\nu_0$,
\begin{equation}
\begin{aligned}
    \mathcal{C}_j &= \{\mathcal{S}_j,\mathcal{T}_j\},\\
    R_0 &= \operatorname{concat}_{j=1}^{c}\mathcal{T}_j(\mathcal{S}_j(G_0,\nu_0)).
\end{aligned}
\end{equation}
Here $\mathcal{S}_j$ returns the relevant graph evidence, or an empty sequence if the clause does not apply; $\mathcal{T}_j$ renders the selected facts as sentences. Concatenation of all clauses gives the reasoning target $R_0$. Clauses describe nearby actors and crossings, lane shape and permitted movements, intersections and traffic controls, and how to execute the navigation instruction, with the full set of clauses in Appendix \cref{sec:clauses_templates_say}.

\paragraph{USE: Road Compliance}
To evaluate whether predicted trajectories obey road rules, we build on
EPDMS~\cite{navsimv2}, which checks collisions, drivable-area and
driving-direction compliance, traffic-light compliance, lane keeping, ego
progress and comfort directly against the HD map. EPDMS does not, however,
check whether ego respects stop signs, slows down when approaching a red traffic light, or obeys posted speed limits, leaving a gap to what our road graph contains. We therefore add a speed-compliance term
computed from the same map and traffic-light states: stop signs and red
traffic lights impose a zero-speed target at their stop lines, which we
propagate backward along the route's lanes and connectors as
$v_\text{stop}(d)=\sqrt{2a_\text{plan}d}$, where $d$ is the along-road
distance and $a_\text{plan}$ the assumed braking deceleration. Combining this ceiling with posted speed limits yields the speed field $v_\text{max}$, and for a trajectory state $s_t=(x_t,y_t,\theta_t,v_t)$ we penalize
\begin{equation}
  c_\text{speed}(s_t,t)=\left[v_t - v_\text{max}(x_t,y_t,t)\right]_+^2,
\end{equation}
integrated along the predicted trajectory to capture the magnitude and
duration of speed violations (see Appendix \cref{sec:details_use_field}). 
Each cause (posted limit, stop sign, red light) is integrated and thresholded separately into a binary compliance factor, which multiplies EPDMS to give EPDMS+USE.
\section{Instantiation and data partitioning}
\label{sec:instantiation}

\begin{table*}[t]
\centering
\caption{Share (\%) of decision points in val / test within 10, 50 and 100\,m of a training sample, and their median distance in m to the nearest training sample. Density: mean number of distinct drives through each 10\,m cell and 90$^\circ$ heading bin, with train, val and test pooled. }
\label{tab:train_test_overlap}
\begin{tabular}{@{}lcccccccccc@{}}
\toprule
Dataset & Cities & Train & Val/test & $<$10\,m & $<$50\,m & $<$100\,m  & Median & Density  \\
\midrule
Argoverse 2$^{\dag}$~\cite{Argoverse2} & 6 & 22.6k & 4.8k / 4.8k & 50.4 / 49.0 & 71.4 / 71.9 & 80.2 / 84.7 & 9.7 / 10.6 & 1.3 \\
nuScenes$^{\dag}$~\cite{caesar2020nuscenes} & 2 & 28.1k & 6.0k / 6.0k & 82.8 / 89.9 & 93.9 / 95.3 & 96.5 / 97.5 & 1.3 / 0.7 & 2.0 \\
nuReasoning$^{\dag, \ddag}$\cite{nuReasoning} & 2 & 396.6k & 1.0k / 1.0k & 92.9 / 91.7 & 97.5 / 96.3 & 98.7 / 97.3 & 0.5 / 0.5 & 4.4 \\
nuPlan$^{\dag}$~\cite{nuPlan} & 4 & 7.07M & 678.3k / 719.0k & 100 / 99.9 & 100 / 100 & 100 / 100 & 0.0 / 0.0 & 111.8 \\
NAVSIM~\cite{navsim} & 4 & 85.1k & 18.2k / 12.1k & 91.8 / 95.0 & 97.9 / 99.0 & 99.6 / 99.7 & 0.5 / 0.4 & 6.2\\

\midrule
Buffer zone & 4 &   & 3.1k & 0.5 & 32.3 & 84.9 & 61.8 & 6.4 \\
GEO split (ours) & 4 & 106.3k & 3.7k & 0.0 & 0.0 & 0.0 & 185.5 & 6.4 \\
\bottomrule
\end{tabular}
  \par\vspace{2pt}
    {\raggedleft\scriptsize
    $^{\dag}$ sampled at 2\,Hz; $^{\ddag}$ computed on the 60\% of currently released dataset\par}
\vspace{-0.5cm}
\end{table*}

We instantiate Road-KSU on nuPlan's~\cite{nuPlan} NAVSIM \cite{navsim} derivation. Not only is this a commonly used driving dataset and EPDMS its native metric, its repeated traversals and dense geographic coverage also support studying road understanding across familiarity and generalization.\footnote{All code, the Road-KSU graph builder, the KNOW/SAY generation pipeline, the USE metric, and our geographic splits will be publicly released upon acceptance.}

\paragraph{Road-KSU on nuPlan}
We construct the scene graph $G_0$ from nuPlan's HD maps, ego poses, tracked actor annotations, and traffic-light states. We set the ego-aligned window to $\Delta x=\Delta y= \ $\qty{50}{\metre}, and use $n= \ $\qty{3}{\second} of historical positions to resolve ambiguous lane assignments. Our node vocabulary $\mathcal{V}$ contains 18 node types, while the edge vocabulary $\mathcal{R}$ comprises 26 labels. Executing these graphs through our 40 KNOW-queries, our framework produces around 19 QA pairs per decision point on average (min: 5, max: 44), for a total of 2.09 million pairs, while SAY's reasoning traces have an average of 5.5 clauses (min: 3, max: 9). Details are in Appendix \cref{sec:roadksu_nuplan}. To assess VQA quality, two annotators independently judged whether each of 400 randomly sampled questions was reasonable from camera images and navigation instruction, and agreed on 96.0\% of the questions (Gwet's AC1=0.96). A third annotator resolved the 16 disagreements, leaving 387 of the 400 questions (96.8\%) judged reasonable; 10 of the 13 remaining errors stem from answers not visible in the camera images or from incorrect HD-map annotations. Additional details on the quality assurance for both VQA and CoT are provided in Appendix \cref{sec:quality_checks}.

\paragraph{Geographic splits}
As \Cref{tab:train_test_overlap} shows, widely used driving datasets exhibit substantial geographic overlap between training and test locations, limiting their ability to assess generalization to unfamiliar roads. We therefore create a geographically held-out split within nuPlan, preserving coverage across cities and road types. We discretize ego trajectories into \qty{10}{\metre} cells, connect consecutively traversed cells, and assign each cell to its nearest eligible intersection by graph distance, merging anchors with shared cells. We select complete intersection-centered regions covering 15--20\% of occupied cells per city, including at least two regions and both traffic-light and stop-sign intersections where available. 
Selection minimizes the recorded driving distance excluded from training by test regions and discarded buffers, which we introduce to separate train and test areas by at least \qty{100}{\metre}. 
Finally, we assign NAVSIM points whose full input history and prediction horizon lie within a zone to GEOTRAIN, GEOTEST, or the discarded buffer, excluding points that span zone boundaries (\Cref{tab:train_test_overlap} and Appendix \cref{sec:partitioning_details}). 

\paragraph{Natural vs. balanced test VQA selection}
Geographic splits test transfer of road understanding to unfamiliar roads, but models may still exploit common road configurations through answer-frequency priors, since roads share regular configurations (\eg, most roads have one lane in each direction). Hence, we additionally construct natural and answer-balanced KNOW test sets. The \emph{natural} set processes frames in random order and selects the least-used applicable query, spreading question types as evenly as possible without explicitly balancing answers. The \emph{balanced} set uses the same frames and per-query question counts, but distributes answers as evenly as availability permits, greedily assigning frames to the applicable (query, answer) pair furthest below its target count. A blind prior predicting the most frequent training answer shows how much the natural set rewards answer frequencies: it achieves 57.6\% accuracy on the natural but only 38.9\% on the balanced set (see Appendix \cref{sec:nat_bal_stats}).

\section{Experiments}
\label{sec:experiments}
To organize our experiments around the $2\times2$ evaluation space of output modality and place familiarity, we first examine action-space performance on seen vs. unseen roads (\Cref{sec:dose_response}), then assess the corresponding language outputs (\Cref{sec:language_drop_main}). Finally, we vary visitation count and geographic coverage to study how both modalities respond to the training distribution (\Cref{sec:place_scaling}). We evaluate driving VLAs in both output spaces and include vision-only planners in the action-space analysis. 

To isolate the effects of Road-KSU supervision, we additionally finetune Qwen3-VL-4B-Instruct \cite{Qwen3-VL} on identical decision points, with the only difference being the target text. \textbf{RoadVLA-0} predicts trajectories; \textbf{RoadVLA-SAY} generates task-focused road descriptions before the trajectory; and \textbf{RoadVLA-KNOW} predicts the trajectory, then answers a KNOW question. Following the configuration of \cite{qwendrive}, each model receives past and current camera images, ego-motion history, and a navigation instruction, and predicts the next \qty{5}{\second} trajectory as waypoints in text. Input specifications, trajectory encoding, training hyperparameters and details on compute resources are provided in Appendix \cref{sec:training_details}.

\begin{figure}
    \centering
  \input{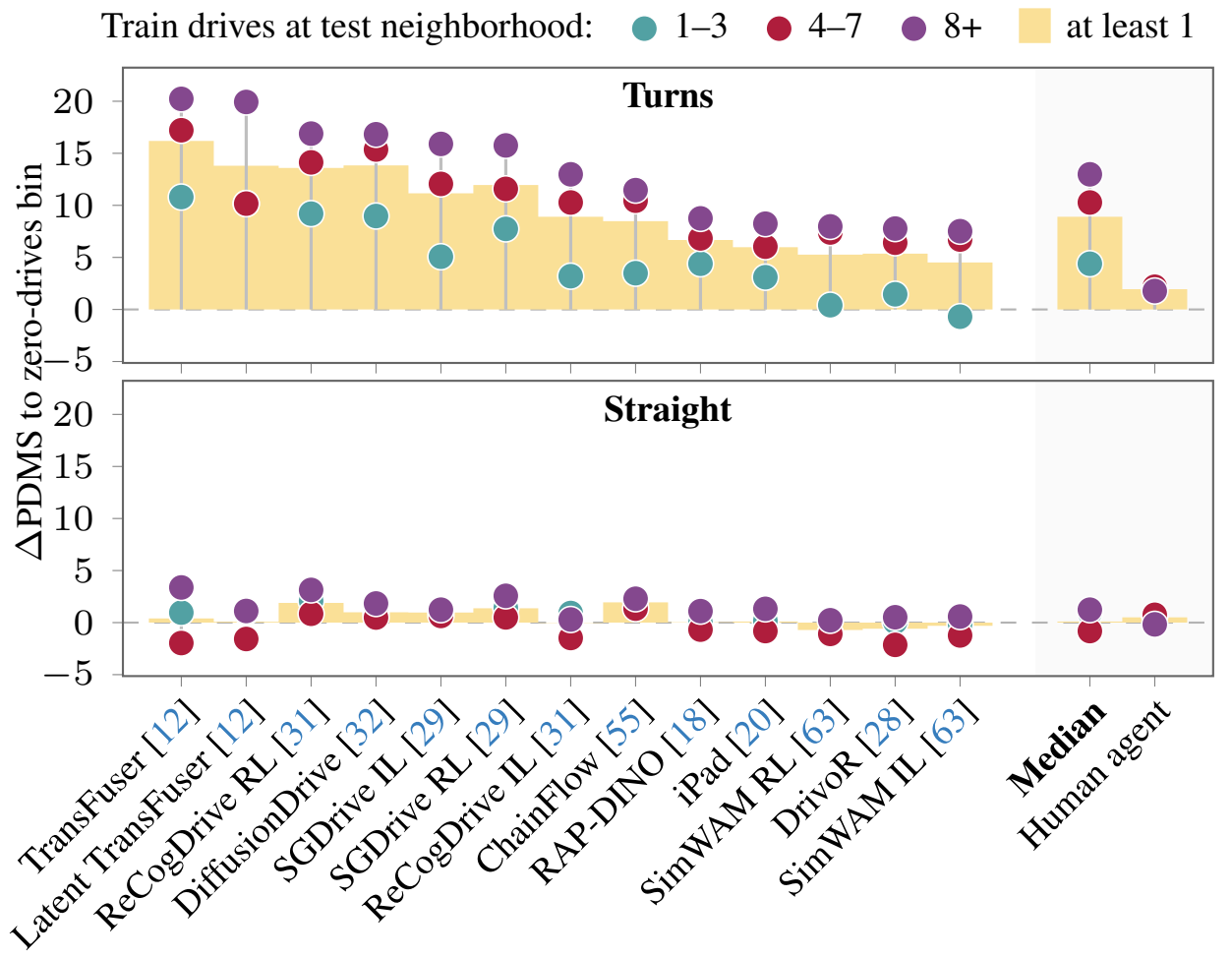}
  \caption{Driving performance follows training exposure on NAVTEST. PDMS differences relative to zero-exposure locations, for turns and straight driving.}
  \vspace{-0.3cm}
    \label{fig:dose_response}
\end{figure}

\subsection{Action-space road understanding}\label{sec:dose_response}
\paragraph{Training exposure on NAVSIM} To ground our study in the current state of the field, we begin with an exploratory analysis of NAVSIM, a widely used benchmark on which current planners achieve strong performance. Its test split substantially overlaps with training locations: 95\% of its decision points lie within \qty{10}{\metre} of a training point (\cf \Cref{tab:train_test_overlap}), allowing us to investigate whether planners score better at places visited more often during training.

\begin{table*}[t]
\centering
\caption{Driving performance on NAVTEST (seen) and GEOTEST (unseen roads), EPDMS+USE evaluation. Gaps use raw, equal-city, and NAVTEST-city weighting; the mean gap averages the three.}\label{tab:retrain_leakage}
\begin{tabular}{@{}l cc cc cc|cc@{}}
\toprule
& \multicolumn{2}{c}{NAVTEST (seen)} & \multicolumn{2}{c}{GEOTEST (unseen)} & \multicolumn{2}{c|}{raw / city avg.\ / NAVTEST avg.} & \multicolumn{2}{c}{Mean $\Delta$} \\
\cmidrule(lr){2-3}\cmidrule(lr){4-5}\cmidrule(lr){6-7}\cmidrule(l){8-9}
& straight & turn & straight & turn & $\Delta$ straight & $\Delta$ turn & straight & turn \\
\midrule
Latent TransFuser \cite{chitta2023transfuser} & 83.9 & 77.7 & 78.5 & 65.2 & 5.4\,/\,2.4\,/\,3.7 & 12.5\,/\,9.4\,/\,10.1 & 3.8 & 10.6 \\
ReCogDrive IL \cite{li2026recogdrive} & 87.8 & 74.3 & 82.2 & 64.9 & 5.6\,/\,3.7\,/\,4.2 & 9.4\,/\,7.7\,/\,8.5 & 4.5 & 8.5 \\
DrivoR \cite{kirby2026drivor} & 89.7 & 84.4 & 87.4 & 74.3 & 2.3\,/\,1.4\,/\,2.4 & 10.1\,/\,7.0\,/\,7.4 & 2.0 & 8.2 \\
RoadVLA-0 & 86.7 & 78.3 & 83.3 & 68.6 & 3.4\,/\,3.9\,/\,5.3 & 9.7\,/\,10.2\,/\,11.9 & 4.2 & 10.6 \\
\midrule
RoadVLA-SAY & 87.5 & 79.2 & 82.9 & 68.7 & 4.5\,/\,4.0\,/\,4.8 & 10.5\,/\,10.8\,/\,12.2 & 4.4 & 11.1 \\
RoadVLA-KNOW & 87.5 & 78.9 & 83.5 & 70.2 & 4.0\,/\,5.0\,/\,6.6 & 8.7\,/\,8.6\,/\,9.7 & 5.2 & 9.0 \\
\midrule
Human agent (PDMS) & 94.8 & 93.9 & 93.8 & 92.4 & 1.0\,/\,$-$0.1\,/\,0.2 & 1.5\,/\,0.6\,/\,0.8 & 0.4 & 1.0 \\
\bottomrule
\end{tabular}
\end{table*}
To stratify performance by training exposure, we define the \emph{neighborhood} of a test sample by its city, a \qty{5}{\metre} radius and a heading cone of $\pm$45$^\circ$ around the ego, distinguishing travel directions and junction approaches. Its \emph{training exposure} is the number of distinct training logs containing a decision point in that neighborhood. We group NAVTEST into exposure bins of 0, 1--3, 4--7, and $\geq$8 logs, and evaluate published baselines whose action heads were trained exclusively on NAVTRAIN's train logs (85k samples). For each model, we then compute mean PDMS per bin relative to zero-exposure samples within each city, and average these differences across Boston and Pittsburgh (Singapore lacks heavily seen turns, while Las Vegas has no never-seen turns). We score the human agent as a reference for scene difficulty, and analyze turns and straight driving separately, using a $15^\circ$ net heading-change threshold over the recorded \qty{4}{\second} future.

\Cref{fig:dose_response} reveals a strong association between training exposure and PDMS on turns, which is obscured when pooling turns and straight driving. Across the three nonzero exposure bins, the median baseline scores 4.4, 10.3 and 13.0 points above never-seen places, and every baseline improves in the highest-exposure bin. Straight driving shows no clear trend. Notably, the pattern persists under alternative spatial definitions, turn thresholds, and exposure bins (Appendix \cref{sec:dose_response_details} for details). These results suggest that familiarity with road structure may boost planner scores primarily on turns, where predicted trajectories depend strongly on road geometry and hence benefit from memorizing the road layout.

\paragraph{Geographically held-out retraining}\label{sec:geo_drop_main}
The training exposure in the above analysis rests on the local, direction-specific definition of a neighborhood (\qty{5}{\metre} radius, $\pm$45$^\circ$ heading). Hence, a test sample with zero-exposure may still have seen the same intersection from another leg, or traversed a road in the opposite direction during training. To test whether the observed pattern persists on completely unfamiliar roads, we retrain Latent TransFuser~\cite{chitta2023transfuser}, ReCogDrive IL~\cite{li2026recogdrive}, DrivoR~\cite{kirby2026drivor} and RoadVLA-0 on NAVSIM and our geographically disjoint splits, respectively, and compare turning and straight-driving performance between test sets. 
\begin{figure}
    \centering
  \input{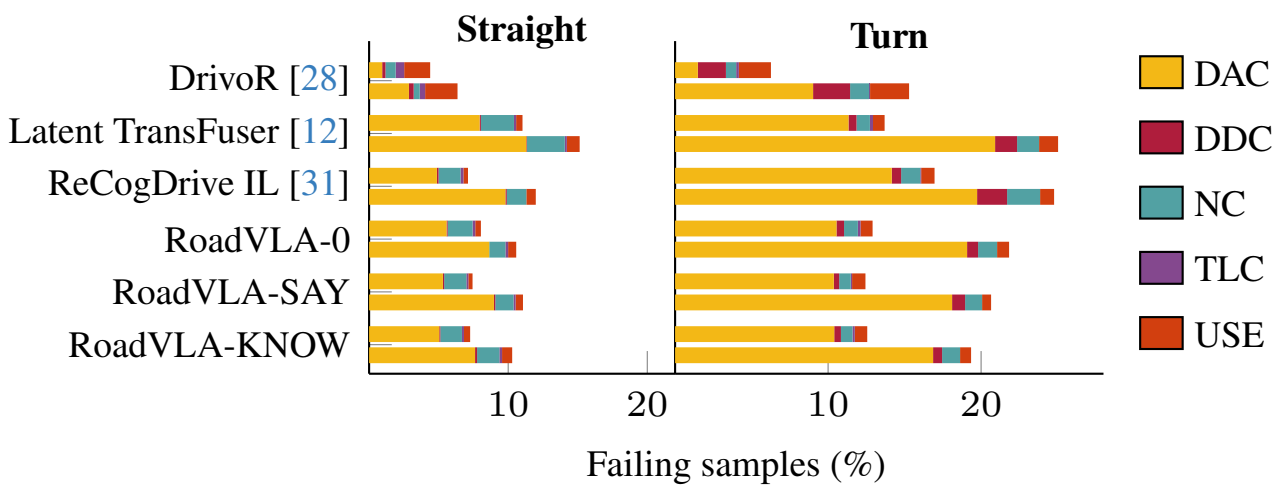}
  \caption{First violated EPDMS+USE term per failing sample on seen (NAVTEST, top) vs. unseen (GEOTEST, bottom) roads.}
    \label{fig:failures}
\end{figure}
For the external baselines, we follow their original recipes and match their original training-sample counts with GEOTRAIN. We report three different results: the EPDMS+USE gap computed across the complete test sets irrespective of composition, the gap computed per city and averaged over cities to control for differences in per-city effects, and the gap computed per city and reweighted to the NAVTEST split to account for distribution differences. Notably, since NAVTEST mixes low- and high-exposure locations, this comparison likely underestimates the performance difference between familiar and unfamiliar roads (see Appendix \cref{sec:geo_drop_addition} for a stratified analysis).

As \Cref{tab:retrain_leakage} shows, the pattern from \Cref{fig:dose_response} persists across all weightings and models, with performance again dropping most on held-out turns.
The failure breakdown localizes this gap (\Cref{fig:failures}): rule-based terms (collisions NC, traffic light compliance TLC, and USE's stop-sign, red-light, and speed-limit checks) fail at comparable rates on seen and unseen roads. The additional failures on unseen roads come almost entirely from drivable-area compliance (DAC). Familiarity thus improves geometric compliance, not rule compliance.  Scoring our models' full 5\,s predictions widens the gap further on turns, from 9.0--11.1 at the 4\,s horizon to 15.9--18.1 points of 5\,s EPDMS+USE (Appendix \cref{sec:geo_drop_addition}).
Broader road-context supervision through KNOW modestly improves held-out turning performance, while task-focused SAY supervision provides no improvement, and neither closes the geographic gap, indicating that explicit road information in language does not automatically yield geometrically compliant trajectories.
\begin{figure}
    \centering
  \input{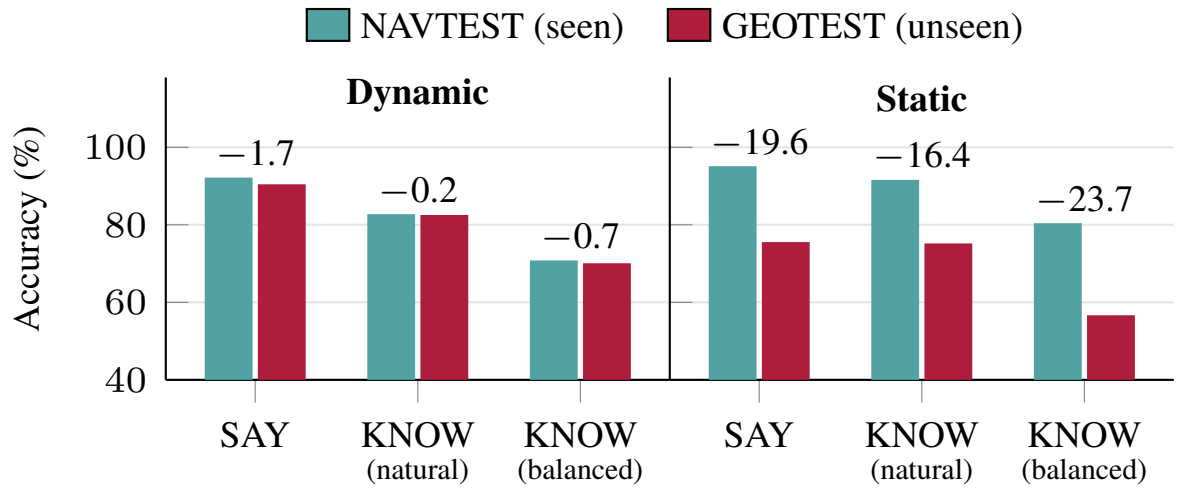}
  \caption{Comparison of language-space accuracy.}
    \label{fig:language_drop}
    \vspace{-0.2cm}
\end{figure}

\subsection{Language-space road understanding}\label{sec:language_drop_main}
The limited benefit of explicit road supervision leaves open how much road understanding successful trajectory prediction requires. Models may already encode the properties needed for nominal driving, while broader road knowledge may contribute little to the evaluated trajectories. To complement the narrow action-space readout of a model's road knowledge, we turn to KNOW and SAY as language-based probes and test whether models can identify road properties beyond the immediate driving task. 
\begin{figure*}[t]
    \centering
  \input{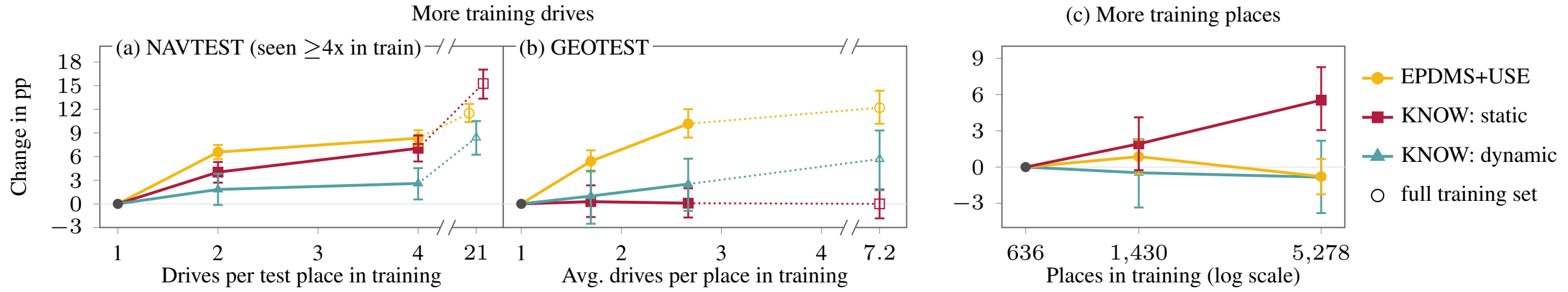}
 \caption{Effects of repeated visitation on familiar (a) and held-out roads (b), and geographic coverage at a fixed frame budget (c). Changes are relative to one drive per place (a,b) and smallest geographical coverage (c). 95\% CI from a paired cluster bootstrap over test places.}
    \label{fig:place_scaling}
\end{figure*}

\paragraph{Road memorization in language} We compare RoadVLA-KNOW and RoadVLA-SAY trained on GEOTRAIN with their NAVTRAIN-trained counterparts. To separate static (\eg,``Is there a lane to your left running in the same direction?'') from scene-dependent road understanding (\eg, ``Is there a VRU in the vicinity of the next crosswalk?''), we group the KNOW queries into two families. For SAY, we score the quality of the predicted reasoning by coarse factual agreement: for the static road, (i)~whether ego is approaching the next intersection or already inside it, (ii)~that intersection's control (ego must stop, traffic light, ego has right of way, uncontrolled), (iii)~the shape of the road ahead (straight, slight bend or bend, and its direction), and (iv)~whether a marked crosswalk crosses ego's path; for the dynamic scene, (v)~lead vehicle presence, and (vi)~the status of the traffic light governing ego's route. We score KNOW by exact match after normalization averaged per query template, with over 99.8\% of predictions being exact matches of one of the template's answers.

As \Cref{fig:language_drop} shows, accuracy drops sharply on static road properties at unseen locations, while dynamic controls stay flat, since they are not place dependent. This is consistent with location-specific learning in both task-relevant descriptions (SAY) and broader road knowledge (KNOW). Unlike action performance, static-language accuracy deteriorates on both straights and turns: drops are respectively 23.0/17.1 points for SAY, 16.7/15.7 for natural KNOW, and 24.5/20.9 for balanced KNOW, while dynamic accuracy changes by at most 3.8 points. Together, these results suggest broader losses in road knowledge beyond the geometry failures on turns observed in action space, providing a potential explanation of why improving language-level road knowledge alone need not yield geometrically compliant trajectories.
\begin{table}[t]
    \centering
    \caption{QA accuracy, natural vs balanced answer distributions.}    \label{tab:vqa_investigation}
    \begin{tabular}{@{}l ccc@{}}
        \toprule
         & Natural & Balanced & $\Delta$ \\
        \midrule
         Human expert$^\dag$ &  93.8 \scriptsize{$\pm$2.7} & 93.8 \scriptsize{$\pm$2.6} & 0.0\\
         Chance & 32.5 & 32.5 & 0.0 \\
         Blind prior & 57.6 & 38.9 & 18.7 \\
        \midrule
        Qwen3-VL-4B \cite{Qwen3-VL} & 53.1 & 48.6 & 4.5 \\
        ReCogDrive \cite{li2026recogdrive} & 52.9 & 43.8  & 9.1 \\
         Qwen3.5-4B \cite{Qwen3-VL, qwen3.5} & 57.1 & 47.9 & 9.2 \\
         InternVL3-2B \cite{zhu2025internvl3}& 50.6  & 39.7  &10.9 \\
        Cosmos-Reason2-8B \cite{azzolini2025cosmosreason1} & 63.2 & 52.0 & 11.2 \\
          Qwen-Drive-1.0 \cite{qwendrive} & 63.4 & 50.2 & 13.2 \\
         RoadVLA-KNOW$^\ddag$ & 80.5 & 66.2 & 14.3 \\
        \bottomrule
    \end{tabular}
    \par\vspace{2pt}
    {\raggedleft\scriptsize
    $^\dag$2 raters, 250 QA / set. 94.8\% agreement; $^\ddag$constrained to the option set\par}
    \vspace{-0.4cm}
\end{table}

\paragraph{Answer priors} Beyond geographic familiarity, the answer distribution also affects measured road knowledge. In the experiment above, RoadVLA-KNOW shows a natural--balanced gap on both known and held-out roads, consistent with reliance on frequent road configurations. To contextualize this sensitivity beyond our specific training setup, we pose the held-out KNOW questions as up-to-four-option multiple choice to human annotators, who see the same camera input as the models, and to general-purpose and driving VLMs
(\Cref{tab:vqa_investigation}). Humans reach 93.8\% on both sets, with no natural--balanced gap ($0.0\pm3.7$, $250$ QA/set), showing that rare configurations are as recognizable from the images as common ones. All models, by contrast, perform worse on the balanced set, showing that reliance on answer priors is widespread among models both with and without driving-specific training (see Appendix \cref{sec:external_baselines_prompt} for evaluation details and human expert study). Taken together, these results indicate that language-space road knowledge is less transferable than driving performance suggests: it degrades on unfamiliar roads regardless of road shape, and part of what remains reflects answer statistics rather than the scene.

\subsection{Scaling behavior of road understanding}\label{sec:place_scaling}
Lastly, we test if language and action space performance respond to the same changes in the training distribution. We use RoadVLA-KNOW throughout and define a \emph{place} as a \qty{10}{\metre} cell and one of four \ang{90} heading bins within a city.

\paragraph{Repeated visitation} We first vary the number of training drives per place while keeping all places: 6,328 in NAVTRAIN and 5,278 in GEOTRAIN. For each place, we randomly draw at most $K \in \{1, 2, 4\}$ drives and keep all their frames there. Smaller $K$ thus yields fewer distinct frames; to hold the number of training samples fixed (33k for GEOTRAIN, 37k for NAVTRAIN), we pair each frame with additional KNOW questions, matching the per-query question count of $K=4$. Additional details on the training set construction are provided in Appendix \cref{sec:place_scaling_details}.

On seen roads (\Cref{fig:place_scaling}, a), we score the frames at places driven at least four times in NAVTRAIN (69\% of NAVTEST), so the runs saw exactly $K$ drives through each test place. More drives raise EPDMS+USE ($+$6.6 and $+$8.3 at $K{=}2$ and 4), static KNOW accuracy ($+$4.0 and $+$7.0) and dynamic accuracy ($+$1.8 and $+$2.6) together. On GEOTEST (\Cref{fig:place_scaling}, b), the corresponding GEOTRAIN runs improve driving similarly (EPDMS+USE $+$5.4 and $+$10.2, from a lower start: 65.8 vs.\ 75.8 at one drive), but static KNOW accuracy does not move ($+$0.3 and $+$0.1, 95\% CIs [$-$1.7, $+$2.3] and [$-$1.7, $+$2.0] from a paired cluster bootstrap over test places). Repeated drives improve static road knowledge only at the places that were driven, consistent with memorization of the local road, while additional drives improve driving in a way that transfers to unseen places.

\paragraph{Geographic diversity} In a complementary experiment in the geographically disjoint training regime, we fix the training budget and increase the number of places it covers, \ie, the geographic diversity. The least diverse set takes the most-driven places first (636 places, 9.4 drives each on average); a second draws places in random order (1,430 places, 3.9 drives each), which also guards against road-type bias in the most-driven places; the most diverse set is the $K{=}1$ subset above, with one drive through each of the 5,278 places (11,820 decision points). All sets contain the same number of samples per city and turn class, and the VQA question mix is matched as far as the decision points allow.

Increasing the number of places from 636 to 1,430 and 5,278 raises static KNOW accuracy on GEOTEST by 1.9 and 5.5 points (\Cref{fig:place_scaling}, c), while dynamic accuracy ($-$0.5, $-$0.8) and EPDMS+USE ($+$0.9, $-$0.8) stay within their intervals. Together with the repetition experiment, this yields a double dissociation: more drives through the same places improve driving on unseen roads but not static road knowledge, whereas broader geographic coverage improves static road knowledge but not driving. 
\section{Discussion and outlook}\label{sec:discussion}
We introduced the Road-KSU framework, which allowed us to study road understanding through language probes (KNOW, SAY) and map-based action scoring (USE) from a shared underlying ground truth, comparing seen and unseen roads. Our findings have three practical implications. 
First, training–test overlap inflates driving scores through memorized road geometry. Benchmarks should therefore report geographically disjoint results, and complement trajectory evaluation with probes of road knowledge beyond the immediate maneuver.
Second, even on held-out roads, common configurations mask what models understand. In action, straight driving remains relatively stable while static-road language accuracy drops substantially, indicating that nominal open-loop driving requires little road understanding. In language, accuracy partly reflects answer priors. Benchmark construction should therefore deliberately include non-standard road configurations where road understanding becomes decisive.
Third, visitation count and geographic coverage are complementary axes of data scaling, and data collection for driving VLAs should treat geographic diversity as a target in its own right rather than a by-product of volume.

\paragraph{Limitations}
Our probes measure observable behavior, not internal representations or the causal role of language in action prediction. Geographic separation reduces local overlap but does not establish full generalization to new cities. Moreover, our controlled scaling results concern one specific training configuration and need not transfer fully to other architectures. Short-horizon NAVSIM evaluation does not establish closed-loop safety or performance during emergency maneuvers. Future work should apply Road-KSU to test these scaling patterns across architectures and datasets, and examine whether broader road knowledge supports alternative maneuvers under changed instructions, obstructions, or reactive traffic.

\section*{LLM Usage Statement}
We used Claude (Anthropic) and ChatGPT (OpenAI) to edit text written by the authors for improved clarity, across all sections of the paper. Suggestions were reviewed, revised and selectively adopted by the authors; all claims, analyses and numbers were produced and written by the authors. We used Claude Code and Codex to generate parts of the code for graph construction, question and reasoning-template generation, USE metric computation, training and evaluation scripts, and plotting. The extent of use was drafting individual functions and boilerplate from author specifications. All generated code was reviewed and tested by the authors, and the outputs of the KNOW pipeline were additionally checked through the manual review described in \Cref{sec:instantiation}. LLMs were not used to generate data, labels, results, references, or the research ideas and experimental design. LLMs evaluated as subjects of study (\Cref{tab:vqa_investigation}) are part of the experimental methodology and are described in the main text.

{
    \small
    \bibliographystyle{ieeenat_fullname}
    \bibliography{main}
}

\clearpage
\maketitlesupplementary

\section{Details of graph construction}
\label{sec:graph_construction_details}

Road-KSU represents a decision point by a directed graph $G_0$ whose entities connect the numerical annotations of nuPlan~\cite{nuPlan} to the road concepts used by KNOW, SAY, and USE (see \Cref{sec:graph_execution}). This section specifies the construction implemented by the graph builder $\mathcal{B}$. 

\subsection{Inputs, spatial support, and coordinates}
\label{sec:suppl_graph_inputs}
The inputs to $\mathcal{B}$ are the nuPlan \cite{nuPlan} HD map $M$ in a window of $[(-\Delta x/2,+\Delta x/2),(0, +\Delta y)]$ around the ego pose in forward heading direction, the driven route $\nu_0$ expressed as an ordered sequence of nuPlan \texttt{roadblock} IDs, the current ego pose and speed $\vec{p}_\text{ego}$, current tracked actors $\{\vec{p}_\text{i}\}_{i=1}^{n_\text{actor}}$ and traffic-light states $\{L_\text{j}\}_{j=1}^{n_\text{tl}}$, and the preceding $n=3$\,s of ego and actor positions. Map layers supply lane and connector polygons, their baselines and connectivity, road segments, intersection polygons, crosswalks, walkways, parking areas, and stop polygons. Lanes explicitly annotated as bicycle lanes, and connectors entering or leaving them, are removed from the motor-vehicle lane topology since we find annotations in nuPlan to be inconsistent across cities. Cyclists remain represented as actors. The standard graph window's origin is an estimated driver position 1.9\,m ahead of nuPlan's rear-axle pose along the ego heading, corresponding to the driver's seat. Historical ego poses are shifted in the same way for localization. This offset defines the reference for the graph. Map features are generally selected by intersection with the window, whereas actors are selected by intersection of their oriented boxes with it. Consequently, a retained object's center or part of its geometry can lie outside the window. 
Construction uses map coordinates. After association, positions are translated and rotated into the driver-centered frame. The graph stores representative positions as \texttt{local\_x} and \texttt{local\_y}; actor \texttt{yaw} is relative to the ego heading. Dimensions and distances are in meters, speeds in meters per second, and angles in radians.

\subsection{Ego anchoring}
\label{sec:suppl_graph_anchor}
We first test which map polygons contain the ego, considering lane polygons only when no connector contains it. Heading agreement is the cosine of the difference between ego heading and the nearest baseline tangent. If at least one candidate has nonnegative agreement, candidates running against ego are discarded. Among multiple candidates, the largest number of contained historical positions determines the assignment, with heading agreement breaking remaining ties. Without informative history, heading agreement determines the assignment. When no containing lane or connector can be found, the decision point is not exported to the training or evaluation set and discarded.

The ego anchor includes two additional checks. If the containing element runs against ego, the nearest lane running in ego's direction within 10\,m is used when available. When route context is supplied, a route-consistent element containing the rear axle can replace a driver-point assignment that is absent or inconsistent with that route. Route information can also resolve a tied historical overlap between connector candidates.

\subsection{Longitudinal and lateral lane expansion}
Starting from this anchor, the builder follows directed lane--connector--lane connectivity and adds lateral neighbors sharing lane boundaries. Road-segment membership identifies opposite-direction lanes and groups lanes into \text{road} nodes. A spatial fallback includes lane groups not reached by this expansion, and missing connector successors are added when they intersect the window. 

Lane width is approximated by polygon area divided by baseline length, and posted speed limits are read for lanes and connectors where the map provides them. To describe road shape, we follow the heading of lanes and connectors along ego's forward path (\cf Appendix \cref{sec:suppl_graph_route}) for the next 50\,m. A turn begins once the heading has swung more than $15^\circ$ from its initial direction and ends when it swings back by more than $15^\circ$ from the furthest point reached; its angle $\Delta\theta$ is that furthest swing, or the net heading change if the heading never swings back. KNOW and SAY read the road shape from $\kappa=\Delta\theta/\ell$, with $\ell$ the measured length: straight for $|\kappa|\le0.0025$\,rad/m, a slight bend up to $0.005$\,rad/m (about $7^\circ$ and $14^\circ$ over a 50\,m path), and a curve beyond, to the left for positive $\kappa$. Baselines shorter than 8\,m or with fewer than three vertices are not measured and count as straight.

\subsection{Junctions, crosswalks, traffic controls}
\label{sec:suppl_graph_junctions}
Every nuPlan junction polygon overlapping with the graph window becomes a graph node, yet not all become \texttt{intersections}. Signalized, stop-sign-controlled and uncontrolled junctions use type \texttt{intersection}; lane branches, lane merges, pass-through regions, and untyped map aggregates use distinct types, since they do not correspond to what ``intersection'' is commonly referred to in natural language. Connectors are linked to their aggregate junction through the map's intersection identifier by \texttt{IS\_WITHIN}, subject to the spatial window. This preserves lane topology without treating every topological event in the map as an \texttt{intersection} node.

Since nuPlan junction polygons are sometimes labeled as \texttt{default}, \ie their control type is not known, we additionally use map evidence along the ego route forward walk to promote such untyped junctions to \texttt{stop-controlled} or truly \texttt{uncontrolled} intersections: a stop sign on ego's approach, a turn-stop marking on a movement from ego's road, or a stop sign or turn-stop marking on a cross approach. Promotion also requires another incoming road, and is declined when a traffic light governs one of the considered ego movements. Without sufficient evidence, the junction remains \texttt{unknown\_control}. Promoted nodes retain the promotion reason and their original type.

Stop polygons are linked to the lanes, connectors, and crossings they govern or protect. Connector-specific signal groups retain their current traffic-light state. Intersection attributes summarize stop and yield requirements across the represented approaches, distinguishing ego's obligations from those of cross traffic (\eg, \texttt{ego\_approach\_is\_stop\_controlled}, \texttt{cross\_traffic\_must\_yield}). Crosswalks are linked through map identifiers and connector overlap, with crossing positions stored along the connector baseline.

\subsection{Actors, pedestrian groups, roadside features}
\label{sec:suppl_graph_actors}
Vehicles, bicycles, and pedestrians retain their current kinematics and dimensions. Pedestrians and cyclists are first associated with walkways or crosswalks, and vehicles with parking areas. Actors without such an association are matched to represented lanes or connectors using containment, heading, and recent positions. Nearby pedestrians are grouped by proximity within 5\,m into a \texttt{pedestrian\_group}; the group carries their shared polygon location while individual members remain in the graph.

Walkways and parking areas are also related to ego's road by their lateral position. Parking areas overlapping a lane are linked to that lane, while vehicles in curbside parking strips can receive an alongside relation based on alignment and available passing space. The complete entity and relation vocabularies are given in \Cref{tab:graph_node_types} and \Cref{tab:graph_relations}.

\subsection{Route context and ego path}
\label{sec:suppl_graph_route}
To include route information in the graph, the Road-KSU pipeline additionally supplies navigation context obtained from the recorded future: a sequence of \texttt{roadblock} identifiers and a navigation maneuver. This context can affect the ego-anchor tie-breaks described above and selects the continuation at route-dependent forks, and is used to add additional information to the ego node.

After graph construction, a forward-path computation concatenates the selected lane and connector baselines from ego's projected position, with an along-path budget of 50\,m. At a fork it uses the navigation maneuver, then route membership, then topology. The ego node stores this path's curvature profile, length, element identifiers, junction-entry distances, and termination reason. Additional attributes separate the first road-shape segment from the segment after the first junction: when ego is outside, the first segment is its approach; when inside, it is the current junction's complete connector movement. When including route-dependent knowledge (\eg, curvature along the ego route), KNOW and SAY read these shared values rather than independently resolving the road ahead.

\subsection{Graph representation and conventions}
\label{sec:suppl_graph_schema}
The implementation uses a simple directed graph with one relation per ordered node pair. A shared insertion helper adds the declared inverse, or the reverse edge with the same label for symmetric relations (\eg \texttt{lane}$\rightarrow$\texttt{CONTAINS}$\rightarrow$\texttt{ego} $\leftrightarrow$ \texttt{ego}$\rightarrow$\texttt{IS\_IN}$\rightarrow$\texttt{lane}). 
\providecommand{\gfield}[1]{\nolinkurl{#1}} 
Map nodes retain a feature identifier \texttt{fid}; lane, connector, and boundary identifiers allow geometry to be retrieved from the source map. All types carry \gfield{local\_x}, \gfield{local\_y}, and \gfield{world\_z}. Most map nodes and all individual actors also carry \gfield{closest\_x}, \gfield{closest\_y}, \gfield{furthest\_x}, and \gfield{furthest\_y}, expressed in the ego frame. For polygons these are boundary points; the closest and farthest point are selected among outline vertices. Missing source values remain unset where supported, and optional fields in the table are not guaranteed on every instance. 
\Cref{tab:graph_node_types} gives an overview of the node vocabulary $\mathcal{V}$ and its attributes; \Cref{tab:graph_relations} lists the relation vocabulary $\mathcal{R}$ instantiated by the builder $\mathcal{B}$, including the inverses.

\providecommand{\gfield}[1]{\nolinkurl{#1}}
\begin{table*}[tbp]
\centering
\footnotesize
\caption{Node vocabulary of the graph builder $\mathcal{B}$. Type-specific fields are given in addition to the shared fields described in the main text. Source refers to nuPlan map features or actor tracks.}
\label{tab:graph_node_types}
\begin{tabular}{@{}>{\raggedright\arraybackslash}p{\dimexpr.18\textwidth-1.333\tabcolsep\relax}>{\raggedright\arraybackslash}p{\dimexpr.32\textwidth-1.333\tabcolsep\relax}>{\raggedright\arraybackslash}p{\dimexpr.50\textwidth-1.334\tabcolsep\relax}@{}}
\toprule
\textbf{Node type} & \textbf{Meaning and source} & \textbf{Map properties and derived attributes} \\
\midrule
\gfield{ego} & The driver-centered reference entity, derived from the current ego state. & \gfield{speed}, \gfield{heading}, \gfield{driving_side} (left/right traffic); optional \gfield{connector_progress} (fraction along the current connector); \gfield{route_forward_curvature} (describes the path along the ego route).  \\

\gfield{lane} & A motor-vehicle lane polygon and its baseline. & \gfield{lane_group_fid}, \gfield{lane_index}, \gfield{speed_limit_mps}, \gfield{left_boundary_fid}, \gfield{right_boundary_fid}, \gfield{width_m}, \gfield{direction}, \gfield{curvature} values; optional \gfield{one_way}. \\

\gfield{lane_connector} & A directed movement joining a predecessor lane to a successor lane. & \gfield{turn_type} (straight/left/right/uturn), \gfield{speed_limit_mps}, \gfield{lane_group_connector_fid}, \gfield{downstream_lane_group_fid}, \gfield{intersection_fid}, \gfield{curvature} values. \\

\gfield{road} & A source road-segment polygon grouping lanes through its lane-group IDs. It can contain both travel directions. & No additional fields beyond map identity and shared geometry fields. \\

\gfield{intersection} & A signalized, stop-controlled, or uncontrolled road junction. & \gfield{intersection_type}, \gfield{stop_control}, \gfield{total_approach_lanes}, \gfield{stop_controlled_approach_lanes}, \gfield{ego_approach_is_stop_controlled}, \gfield{yield_control}, \gfield{ego_approach_must_yield}, \gfield{cross_traffic_must_yield}; optional \gfield{promoted_from}, \gfield{promoted_reason}. Stop/yield control is none/partial/all\_way or unset. \\

\gfield{lane_branch} & A junction-layer polygon labeled as a lane split. & No additional fields beyond map identity and shared geometry fields. \\

\gfield{lane_merge} & A junction-layer polygon labeled as a lane merge. & No additional fields beyond map identity and shared geometry fields. \\

\gfield{pass_through} & A junction-layer polygon labeled as pass-through. & No additional fields beyond map identity and shared geometry fields. \\

\gfield{unknown_control} & A junction from nuPlan's untyped/default intersection category, without sufficient evidence for promotion to real \texttt{intersection}. & No additional fields beyond map identity and shared geometry fields. Upon promotion, its type and ID prefix become \gfield{intersection} and promotion metadata are added. \\

\gfield{stop_polygon} & A deduplicated cluster of annotated stopping regions. & \gfield{control_type} (list drawn from crosswalk, stop\_sign, traffic\_light, turn\_stop, yield). The smallest cluster \gfield{fid} and its geometry are retained. \\

\gfield{traffic_light} & The signal group controlling one connector. & \gfield{status} (current source state, or unknown). Its \gfield{fid} is the controlled connector ID; the node ID prefix is \gfield{traffic_light}. \\

\gfield{crosswalk} & A mapped pedestrian crossing. & \gfield{is_marked} (Boolean). \\

\gfield{walkway} & A mapped pedestrian walkway, instantiated for actor grounding or as a roadside feature. & No additional fields beyond map identity and shared geometry fields. \\

\gfield{carpark_area} & A mapped parking polygon, instantiated for vehicle grounding, lane overlap, or roadside context. & No additional fields beyond map identity and shared geometry fields. \\

\gfield{vehicle} & A tracked vehicle at $T=0$ whose oriented box intersects the window. & \gfield{category}, \gfield{speed}, \gfield{width}, \gfield{length}, \gfield{yaw}; optional \gfield{lane_free_left_m}, \gfield{lane_free_right_m} when lane clearance can be measured. \\

\gfield{bicycle} & A tracked cyclist at $T=0$ whose oriented box intersects the window. & \gfield{category}, \gfield{speed}, \gfield{width}, \gfield{length}, \gfield{yaw}. \\

\gfield{pedestrian} & A tracked pedestrian at $T=0$ whose oriented box intersects the window. & \gfield{category}, \gfield{speed}, \gfield{width}, \gfield{length}, \gfield{yaw}. \\

\gfield{pedestrian_group} & A cluster of at least two pedestrians at pairwise linkage distance at most 5 m tracked at $T=0$. & \gfield{member_count}; representative position is the mean of member centers.  \\
\bottomrule
\end{tabular}
\end{table*}

\begin{table*}[tbp]
\centering
\caption{Relations instantiated by the graph builder $\mathcal{B}$. The 17 forward labels below expand to 26 distinct directed labels after inverse completion. Symmetric labels are stored in both directions.}
\label{tab:graph_relations}
\footnotesize
\begin{tabular}{@{}>{\raggedright\arraybackslash}p{\dimexpr.27\textwidth-1.5\tabcolsep\relax}>{\raggedright\arraybackslash}p{\dimexpr.20\textwidth-1.5\tabcolsep\relax}>{\raggedright\arraybackslash}p{\dimexpr.22\textwidth-1.5\tabcolsep\relax}>{\raggedright\arraybackslash}p{\dimexpr.31\textwidth-1.5\tabcolsep\relax}@{}}
\toprule
\textbf{Forward relation} & \textbf{Inverse relation} & \textbf{Source and target types} & \textbf{Construction evidence} \\
\midrule

\gfield{IS_IN} & \gfield{CONTAINS} & Ego or actor $\to$ lane/connector. & Containment-based localization; ego receives the anchoring checks described in the text. \\

\gfield{CONNECTS_TO} & \gfield{IS_CONNECTED_FROM} & Lane $\to$ connector; connector $\to$ lane. & Directed source-map connectivity. \\

\gfield{IS_ADJACENT_LEFT_OF} & \gfield{IS_ADJACENT_RIGHT_OF} & Lane $\to$ lane; walkway/parking area $\to$ road. & Shared lane boundaries, or the side of a roadside feature relative to ego heading. \\

\gfield{IS_IN_OPPOSITE_DIRECTION_OF} & same label & Lane $\leftrightarrow$ lane. & Lanes in distinct lane groups grouped by a road segment. Symmetric. \\

\gfield{IS_PART_OF} & \gfield{CONTAINS} & Lane $\to$ road; pedestrian $\to$ group. & Map-derived road membership or proximity-cluster membership. \\

\gfield{IS_WITHIN} & \gfield{FEATURES} & Connector $\to$ junction aggregate. & Source intersection identifier; aggregates include real and topology-only junction types. \\

\gfield{GOVERNED_BY} & \gfield{GOVERNS} & Lane/connector $\to$ stop polygon; connector $\to$ signal group. & Source control attachments and connector-specific signal association. \\

\gfield{PROTECTED_BY} & \gfield{PROTECTS} & Crosswalk $\to$ stop polygon. & The stop polygon lists that crosswalk among the elements it protects. \\

\gfield{CROSSES} & \gfield{CROSSED_BY} & Crosswalk $\to$ lane. & Source crosswalk-to-lane identifier list. \\

\gfield{INTERSECTS} & same label & Crosswalk $\leftrightarrow$ connector. & Geometric overlap; \gfield{progress} and \gfield{entry_m} locate the crossing along the connector. Symmetric. \\

\gfield{IS_ON} & \gfield{CONTAINS} & Actor/group $\to$ walkway/crosswalk; vehicle $\to$ parking area; parking area $\to$ lane; crosswalk $\to$ intersection. & Surface association by containment, parking-to-lane area overlap, or source intersection IDs. \\

\gfield{IS_NEAR} & \gfield{IS_APPROACHED_BY} & Pedestrian/bicycle/group $\to$ crosswalk. & Outside the crossing but within 2 m; group association aggregates member evidence. \\

\gfield{IS_ALONGSIDE} & \gfield{HAS_ALONGSIDE} & Vehicle $\to$ lane. & Curbside parking-strip geometry, alignment, clearance, and adjacency checks. \\

\gfield{CONNECTS_STRAIGHT_AHEAD_TO} & none & Road $\to$ intersection. & Road-to-junction vector lies within $45^\circ$ of the ego-forward axis. \\

\gfield{CONNECTS_AT_LEFT_OF} & none & Road $\to$ intersection. & Road connects on the junction's left in the ego heading frame. \\

\gfield{CONNECTS_AT_RIGHT_OF} & none & Road $\to$ intersection. & Road connects on the junction's right in the ego heading frame. \\

\gfield{CONNECTS_BEHIND_OF} & none & Road $\to$ intersection. & Road-to-junction vector lies within $45^\circ$ of the ego-backward axis. \\

\bottomrule
\end{tabular}
\end{table*}

\section{Queries for the KNOW axis}
\label{sec:question_templates_know}

KNOW applies 40 registered queries to the scene graph. Following the decomposition in the main text, each query combines a selection operator $\mathcal S_i$ with question and answer templates $\mathcal T_i^q$ and $\mathcal T_i^a$. 
A selection operator retrieves the relevant graph entities, checks applicability, and derives the facts needed to fill the templates. It can return no instance, one instance, or several instances, for example one per available turn movement, sometimes producing more than one QA pair per query. 
\Cref{tab:know_queries} specifies these three for every query.

\paragraph{Shared selection rules}
``Ego's lane'' requires ego to be on a lane, while queries defined on its current location can also operate on a connector. Forward queries use the shared path and route resolution described in the graph-construction section. Real-intersection queries exclude lane branches, merges, pass-through regions, and unresolved junctions. Asking ranges are 50\,m for junctions, vehicles, and signals, 40\,m for crosswalks and vulnerable road users (VRUs), and 30\,m for junction-movement details. Lane-choice queries additionally require a distance of at least 5\,m left until the intersection entry. The nearest-vehicle query searches all vehicle nodes in the graph, and the sidewalk-presence query uses represented adjacent walkways.

\paragraph{Answer conventions}
Distance answers use the bands $<$5, 5--10, 10--20, 20--30, and over 30\,m. Queries reporting a distance are omitted within 1\,m of the 5\,m boundary and within 2\,m of the 10, 20, and 30\,m boundaries, to avoid asking ambiguous questions. Lead-actor distances follow the road to the actor's box; nearest-vehicle and nearest-crosswalk distance answers use the closest stored boundary point in a straight line. Intersection distances use path-entry distance when available. Relative-speed answers use a 5\,km/h tolerance around ego's speed. A vehicle is stopped at speeds at most 0.3\,m/s; stopped-state and relative-speed questions omit the ambiguous crawling interval $(0.3,5/3.6)$\,m/s.
\providecommand{\gfield}[1]{\nolinkurl{#1}}
\begin{table*}[tp]
\centering
\caption{The 40 registered KNOW queries, decomposed into selection $\mathcal S_i$, question rendering $\mathcal T_i^q$, and answer rendering $\mathcal T_i^a$. Repeated direction and maneuver variants share a row; braces show alternatives, with $d$ the travel direction, $m$ the movement, and $s$ the lateral side. Question wording and answers are abbreviated.}
\label{tab:know_queries}
\footnotesize
\setlength{\tabcolsep}{4pt}
\begin{tabular}{@{}>{\raggedright\arraybackslash}p{\dimexpr.34\textwidth-1.333\tabcolsep\relax}>{\raggedright\arraybackslash}p{\dimexpr.40\textwidth-1.333\tabcolsep\relax}>{\raggedright\arraybackslash}p{\dimexpr.26\textwidth-1.334\tabcolsep\relax}@{}}
\toprule
\textbf{Selection operator $\mathcal S_i$} & \textbf{Question template $\mathcal T_i^q$} & \textbf{Answer template $\mathcal T_i^a$}\\
\midrule

\textbf{K01} Current or continuing lane; same-direction neighbors; outside junctions. & Are there adjacent same-direction lanes? & Both sides / left / right / neither.\\

\textbf{K02} Forward shape, split at junctions when available. & Is the road ahead straight or curving? & Straight / slight bend / bend; side and segment.\\

\textbf{K03} Ego's road; opposite-direction lanes. & Is this carriageway one-way? & Yes iff no opposite-direction lanes exist.\\

\textbf{K04} Current lane; same-direction neighbors. & Is ego in the leftmost/rightmost lane, both, or neither? & Leftmost / rightmost / only lane / neither.\\

\textbf{K05} Lateral lane sweep; multiple lanes required. & Which lane is ego in, counting from the left? & Lane index out of total.\\

\textbf{K06} Current lane or connector; posted limit $v$. & What is the current speed limit? & $v$ mph.\\

\textbf{K07--08} Ego's road; nonempty lane set in direction $d$. & How many lanes run in \{ego's / opposite\} direction? & Lane count in direction $d$.\\

\textbf{K09} All lanes of ego's road, by direction. & How many lanes are on ego's roadway? & Total and directional counts.\\

\textbf{K10--12} Direct approach, 5--30 m; connectors for $m$. & Can ego \{turn left / turn right / go straight\}? & Yes iff $m$ is available from ego's lane.\\

\textbf{K13} Current connector inside a real junction. & What maneuver is ego currently performing? & Straight / left / right / uturn.\\

\textbf{K14} Next junction within 30 m; reachable movements. & Which movements does ego's lane permit? & Available turn types.\\

\textbf{K15} First split/merge involving ego's path, within 50 m. & Does ego's lane split or merge ahead? & No, or split count / merging lanes.\\

\textbf{K16--18} Direct approach, 5--30 m; lanes offering $m$. & Is ego in the correct lane to \{turn left / turn right / go straight\}? & Yes, with alternatives; or no, with viable lane.\\

\textbf{K19} First real intersection within 50 m; ego outside a junction. & What kind of intersection is ahead? & Control type and approach obligations.\\

\textbf{K20} As K19; include distance or absence. & Is an intersection within 50 m; what type and distance? & None, or type and distance band.\\

\textbf{K21} Intersection within 30 m; all approach lanes; side roads. & Can ego's road enter the road \{left / across / right\}? & Yes iff an approach lane offers entry.\\

\textbf{K22} Route/lane signals within 50 m; consistent known state. & What color is the signal governing ego's route? & Signal color.\\

\textbf{K23} Movements within 30 m; resolved controls. & How is each available movement controlled? & Movements and control requirements.\\

\textbf{K24} Available unsignalized left/right turn within 30 m. & If ego turns \{left / right\}, must it yield? & No, or yes with yield/stop explanation.\\

\textbf{K25} Nearest marked path crossing within 40 m. & How far ahead is the nearest crosswalk? & None, or distance band.\\

\textbf{K26} First marked path crossing; protecting controls. & Is that crosswalk protected by a light or stop sign? & Unprotected, or control types.\\

\textbf{K27} First marked path crossing; on/near VRUs. & Is a VRU on or near that crosswalk? & Yes iff eligible VRUs are present.\\

\textbf{K28} Movements from ego's approach road; marked crossings. & Would \{going straight / turning left / right\} cross a crosswalk? & Movement unavailable / no crossing / count and entry/exit locations.\\

\textbf{K29} Available movements, including U-turn; crossings/VRUs. & Would that movement cross a crosswalk with VRUs? & Yes iff a crossing has eligible VRUs.\\

\textbf{K30} Nearest nonparked lead vehicle within 50 m. & Is there a lead vehicle in ego's lane; how far ahead? & None, or yes and distance band.\\

\textbf{K31}$^\dag$ Nearest bicycle-type actor on ego's path within 40 m. & Is a bicycle, motorcycle or tricycle ahead in ego's lane? & Absent, or yes and distance band.\\

\textbf{K32} Nearest vehicle by straight-line boundary distance. & How far away is the closest vehicle? & None, or distance band.\\

\textbf{K33} Stopped vehicles along ego's path; parking or lane-edge evidence. & Are vehicles street-parked along ego's route? & Yes iff any qualify as street-parked.\\

\textbf{K34--35} Sweep lanes on side $s$; first lead vehicle within 50 m. & Is there a lead vehicle in the lane to the \{left / right\}? & No lane / no vehicle / distance and lanes over.\\

\textbf{K36} Adjacent walkways and their VRUs. & Are pedestrians or cyclists on the sidewalk? & None, or sides and actor categories/counts.\\

\textbf{K37} Lead vehicle selected in K30; known speeds; exclude crawling. & Is the lead vehicle faster, slower, or at ego's speed? & Faster / slower / about the same.\\

\textbf{K38} Lead vehicle selected in K30; known speed; exclude crawling. & Is the lead vehicle stopped? & Yes iff speed is at most 0.3 m/s.\\

\textbf{K39--40} Vehicle in the adjacent lane on side $s$; exclude crawling. & Is the vehicle to the \{left / right\} faster or slower? & Faster / slower / about the same.\\

\bottomrule
\end{tabular}
 \par\vspace{2pt}
    {\raggedleft\scriptsize
    $^\dag$excluded from evaluation, as it is almost always negative (only 16 GEOTEST and 32 NAVTEST instances are positive, around 0.4\%)\par}
\end{table*}

\section{Clauses for the SAY axis}
\label{sec:clauses_templates_say}

SAY renders task-focused road descriptions from ten ordered clause types $\mathcal{C}_j$. As detailed in \Cref{sec:graph_execution}, each clause combines a selection operator $\mathcal S_j$ over the scene graph and navigation context with a rendering template $\mathcal T_j$. \Cref{tab:say_clauses} summarizes the ten types in their rendering order. An inapplicable clause contributes no text; an applicable clause may contain several sentences, for example to describe multiple crossings or both the current and next junction. Concatenation finally joins their sentences with spaces.

\paragraph{Shared facts and SAY-specific rendering}
SAY uses the same road, actor, crossing, and control resolvers as KNOW (see Appendix \cref{sec:question_templates_know}). The corresponding asking ranges are 50\,m for junctions, vehicles, and signals, 40\,m for crossings and VRUs, and 30\,m for movement details. Distances use coarse bands, and road shape uses straight/slight-bend/bend descriptions. Unlike KNOW's distance questions, SAY does not omit statements near band boundaries.  Stop-sign stopping distance is stated in whole meters. 

The navigation clause uses the supplied instruction and its intersection references along the ego route. Wording such as ``in sight'' refers to the intersection being represented on the selected path within the 50\,m asking range and not a camera-visibility test, since SAY supervision contains privileged map information.

\providecommand{\gfield}[1]{\nolinkurl{#1}}
\begin{table*}[tbp]
\centering
\caption{The ten SAY clauses in rendering order, decomposed into selection $\mathcal S_j$ and rendering $\mathcal T_j$. Bracketed phrases are placeholders or alternatives; wording and branch descriptions are abbreviated. A clause can render several sentences.}
\label{tab:say_clauses}
\footnotesize
\begin{tabular}{@{}>{\raggedright\arraybackslash}p{\dimexpr.45\textwidth-\tabcolsep\relax}>{\raggedright\arraybackslash}p{\dimexpr.55\textwidth-\tabcolsep\relax}@{}}
\toprule
\textbf{Selection operator $\mathcal S_j$} & \textbf{Rendering template $\mathcal T_j$}\\
\midrule

\textbf{S01} \gfield{lead_vehicle}. Nearest nonparked lead vehicle along ego's path within 50 m; otherwise select street-parked vehicles. & ``There is a lead vehicle in my lane, [distance][state].'' Otherwise state that the lane is clear of other vehicles, mentioning street parking when present.\\

\textbf{S02} \gfield{crosswalk_vru}. All marked crossings on the selected path within 40 m, nearest first; distinguish VRUs on each crossing from those nearby. & ``[A/Another] marked crosswalk crosses my path [distance][intersection side], [occupancy].'' One sentence per crossing: occupied, VRUs nearby, or clear.\\

\textbf{S03} \gfield{lane_curvature}. Shared forward-path shape, separating the approach/current intersection and following segment when available. & ``The road ahead [runs straight / bends slightly / curves] [side].'' With an intersection, describe the available segments before/through/after it.\\

\textbf{S04} \gfield{adjacent_lanes}. Current/continuing lane outside real intersections; immediate same-direction neighbors. & ``There [is a lane to my left/right / are lanes on both sides] running my way.'' Otherwise: ``Mine is the only lane running my way.''\\

\textbf{S05} \gfield{intersection_ahead}. Current real intersection and subsequent intersections within 50 m; approach-specific control and connector progress. & State whether passing through/leaving an intersection, or approaching one at [distance], with ego/cross-traffic obligations. May also name a following controlled intersection.\\

\textbf{S06} \gfield{complete_stop}. Moving ego outside a real intersection; nearest stop-sign line ahead on its approach path, 1--50 m away. & ``I need to slow down and come to a complete stop in [whole meters] meters.'' Omit when already stopped or no eligible line exists.\\

\textbf{S07} \gfield{traffic_light}. Consistent signal groups governing the route movement, otherwise ego's lane, within 50 m. & ``The traffic light governing my [subject] is [status].'' Omit absent or conflicting signal evidence.\\

\textbf{S08} \gfield{permitted_movements}. Next real intersection within 30 m from ego's lane; known movement types on its approach connectors. & ``My lane [only permits / permits] me to [movement list] at the next intersection.'' Movements: straight, left, right, U-turn.\\

\textbf{S09} \gfield{route_stop_line}. Resolved navigation movement with an in-range turn-stop requirement; exclude signals and other non-crosswalk control types. & ``[Upon turning left/right / Upon making a u-turn / Crossing this intersection], I need to yield to cross-traffic.''\\

\textbf{S10} \gfield{nav_execution}. Supplied instruction and intersection binding; current/next intersection state, path range, and lanes offering the maneuver. & State the maneuver through the current intersection, at the next one, or after leaving. Mention a required side-lane change, unavailable movement, or that no intersection is in sight.\\

\bottomrule
\end{tabular}
\end{table*}

\section{Details on USE terms}
\label{sec:details_use_field}
USE supplements EPDMS with speed-compliance costs derived from the same shared HD map inputs. We consider three terms: posted speed limit, stop-sign approach, and red-light approach. Each applicable term defines a nonnegative cost along a candidate trajectory. 

\subsection{Evaluation context and spatial support}
All states are expressed in the ego frame at $T=0$, with the rear axle at the origin, $x$ forward, and $y$ left. USE builds a separate graph with the same construction routines as KNOW and SAY, over a corridor following map successors for a 130\,m travel budget, to account for longer predicted trajectories when traveling at higher speeds. The lanes and connectors along the route are buffered by 25\,m laterally. This extends coverage beyond the language queries' window.

\subsection{Speed-compliance fields}
\paragraph{Speed ceilings}
Each speed term $k$ has its own ceiling $\bar v_k(x,y,t)$. For a target speed $v_\star$ at along-road distance $d$ ahead, the backward extension is
\begin{equation}
\bar v_k(d)=\sqrt{v_\star^2+2a_{\mathrm{plan}}d},
\qquad a_{\mathrm{plan}}=1.5\,\mathrm{m/s^2}.
\end{equation}
A zero-speed target therefore requires progressively slower approaches to a stop line. Propagating the stop constraint backwards operates on reachable lanes and connectors, including same-direction lateral neighbors. It does not follow opposite-direction relations or continue beyond a U-turn connector. Ceilings are propagated backward through connectivity until reaching the posted speed limit of that lane or connector. Stop signs impose a zero-speed target at the mapped stopping region. Mapped traffic-light stop lines use the lane end and bind only while every outgoing connector is red. Thus, a red turn arrow does not slow an approach that also offers a green movement. 

\paragraph{Cost and stop release}
The unweighted cost for each speed cause is
\begin{equation}
 c_k(s_t,t)=\bigl[v_t-\bar v_k(x_t,y_t,t)\bigr]_+^2,
 \quad [z]_+=\max(z,0).
\end{equation}
Different causes are scored separately, so exceeding both a posted limit and a stopping ceiling contributes to both terms. A cause contributes zero where no ceiling is defined. Stop-sign requirements are released after ego reaches at most 0.3\,m/s within 8\,m before or 2\,m beyond the line, measured along the lane. A qualifying stop during the preceding 10\,s also counts, \eg, if the history already reached zero-speed. Once released, that stop line does not constrain the rest of the scored path. This requires memory of earlier trajectory states, distinguishing a vehicle pulling away after stopping from one rolling through. Red lights and posted limits are never released by stopping.

\subsection{USE scoring and combination with EPDMS}
Trajectories $\gamma$ are given as waypoints in $[x,y]$, with speed estimated by central differences over a 1\,s baseline. Each term's weighted cost is integrated over the horizon $H$ by the trapezoidal rule,
\begin{equation}
 S_k[\gamma]=w_k\int_0^H c_k(s_t,t)\,dt,
\end{equation}
with $w_k=0.1$ for the posted limit and $w_k=1$ for stop-sign and red-light approaches. A term fails when $S_k>5$; for example, when the trajectory exceeds a stopping ceiling by 2.3\,m/s for 1\,s, or the posted limit by 3.6\,m/s throughout a 4\,s horizon.
EPDMS+USE multiplies EPDMS \cite{navsimv2} by one compliance factor per term. When defining $\gamma_{\mathrm{human}}$ as the recorded human trajectory,
\begin{equation}
\begin{aligned}
 &C_k(\gamma)=
 \begin{cases}
 1,& S_k[\gamma]\leq 5 \ \text{or}\\
   & \bigl(k\neq \mathrm{stop}\ \land\
       S_k[\gamma_{\mathrm{human}}]>5\bigr),\\
 0,&\text{otherwise},
 \end{cases}\\
 &(\mathrm{EPDMS}+\mathrm{USE})(\gamma)=
 \mathrm{EPDMS}(\gamma)\prod_k C_k(\gamma).
\end{aligned}
\end{equation}
As in EPDMS's human filter, the speed-limit and red-light terms are waived for every agent at decision points where the recorded human fails them too. The stop-sign term $k=\mathrm{stop}$ is not, since a recorded driver who does not stop at a stop sign still violates the rule. Applied to our test sets, we find 47 NAVTEST and 19 GEOTEST decision points where the recorded human effectively performs rolling stops and fails to come to a complete stop. Overall, the effect on the results reported in \Cref{sec:geo_drop_main} remains small. Waiving the term as well would raise mean EPDMS+USE by 0.2--0.5 points across models.  EPDMS and USE are scored over the same horizon of $H=4$\,s. Since a term charges nothing where no ceiling is defined (and not all nuPlan HD maps carry posted limits for all elements), missing map coverage can hide a violation but cannot create one.
\section{Details on Road-KSU on nuPlan}
\label{sec:roadksu_nuplan}

We instantiate Road-KSU on the decision points of NAVSIM~\cite{navsim}, \ie, the difficulty-filtered 2\,Hz decision points of nuPlan~\cite{nuPlan} logs listed in NAVSIM's scene filters. We consider both the original NAVTRAIN/NAVTEST splits and our own GEOTRAIN/GEOTEST repartitioning (see Appendix \cref{sec:partitioning_details} for details).

\paragraph{KNOW}
All 40 queries (\cf Appendix \cref{sec:question_templates_know}) fire on every split. On GEOTRAIN, a decision point yields 19.7 question--answer pairs on average (5 to 44), 2.09 million in total. Of these, 62\% concern the road layout and traffic rules (static VQA) and 38\% actors and traffic lights (dynamic VQA). Query frequencies differ widely. Queries about ego's own lane apply at virtually every decision point, while the rarest, yielding to oncoming traffic when turning, yields 0.12 questions per decision point. Some queries yield several instances, for example one per available turn movement. Training uses one question per decision point (see Appendix \cref{sec:training_details}), as do the natural and balanced test sets.

\paragraph{SAY}
The lead-vehicle, road-shape and navigation clauses appear at every decision point, and the other seven clauses only when their facts hold (\Cref{tab:roadksu_nuplan}). On GEOTRAIN, a reasoning covers 5.5 clause types (3 to 9) in 68 words on average (29 to 152).

\paragraph{Distributions across splits}
Both NAVTRAIN and GEOTRAIN draw on the same pool of NAVSIM frames, and their Road-KSU content is nearly identical. The query mix of GEOTRAIN differs from NAVTRAIN's by a total variation distance of 0.005, and no clause frequency differs by more than 1.1 points. The GEOTEST regions differ more from NAVTEST (total variation 0.07 for the query mix). They are centered on intersections and weight the cities differently, so they contain more crosswalk (53\% \vs 42\%) and stop-line clauses (14\% \vs 7\%) and fewer traffic-light clauses (12\% \vs 30\%) (see Appendix \cref{sec:partitioning_details} for how we mitigate these differences when reporting our experimental results). 

\paragraph{Decision points}
The split sizes in \Cref{tab:train_test_overlap} and Appendix \cref{sec:partitioning_details} count the decision points each split provides; on the geographic split, these are the decision points whose 1.5\,s input history and 5\,s trajectory target lie within a single interval of their side. \Cref{tab:roadksu_nuplan} instead counts the decision points that Road-KSU processes, which additionally requires ego to be localized on the HD map and the graph to yield at least one KNOW question and a SAY reasoning. This keeps 103,073 of NAVTRAIN's and 12,137 of NAVTEST's decision points, and 106,106 of GEOTRAIN's and 3,743 of GEOTEST's decision points, at least 99.8\% in every split. NAVTRAIN here includes the 18.2k decision points that NAVSIM's NAVTRAIN filter takes from nuPlan's validation logs (\cf \Cref{tab:train_test_overlap}).

Several examples of generated KNOW QAs and SAY clauses are provided in \Cref{fig:qualitative_know} and \Cref{fig:qualitative_say}. 

\begin{table}[t]
\centering
\caption{Road-KSU on nuPlan per split: decision points, KNOW question--answer pairs and share of decision points with each SAY clause.}
\label{tab:roadksu_nuplan}
\small
\begin{tabular}{@{}lrrrr@{}}
\toprule
& \multicolumn{2}{c}{GEO split} & \multicolumn{2}{c}{NAVSIM split} \\
\cmidrule(lr){2-3}\cmidrule(l){4-5}
& Train & Test & Train & Test \\
\midrule
Decision points & 106,106 & 3,743 & 103,073 & 12,137 \\
KNOW pairs avg. & 19.7 & 19.1 & 19.6 & 20.3 \\
\quad range & 5--44 & 5--41 & 5--44 & 5--43 \\
SAY clause types & 5.5 & 5.6 & 5.5 & 5.6 \\
\quad range & 3--9 & 3--9 & 3--9 & 3--9 \\
\midrule
\multicolumn{5}{@{}l}{\emph{Decision points with the SAY clause (\%)}} \\
Lead vehicle & 100.0 & 100.0 & 100.0 & 100.0 \\
Road shape & 100.0 & 100.0 & 100.0 & 100.0 \\
Navigation & 100.0 & 100.0 & 100.0 & 100.0 \\
Adjacent lanes & 77.4 & 72.4 & 77.1 & 77.4 \\
Intersection ahead & 69.1 & 70.8 & 69.3 & 68.6 \\
Crosswalk & 36.4 & 52.6 & 37.2 & 41.9 \\
Permitted movements & 28.4 & 30.0 & 28.4 & 30.8 \\
Traffic light & 28.1 & 12.0 & 27.0 & 30.2 \\
Complete stop & 5.2 & 5.4 & 5.2 & 4.8 \\
Stop line on route & 5.0 & 14.1 & 5.5 & 7.0 \\
\bottomrule
\end{tabular}
\end{table}

\providecommand{\qualknowdir}{supplementary/figures/KNOW_qualitative}

\begin{figure*}[h]
\centering
\newlength{\qkgap}\setlength{\qkgap}{0.1pt}
\newlength{\qkh}
\setlength{\qkh}{\dimexpr(\linewidth-3\qkgap)*63/278\relax}
\newlength{\qkfrontw}
\setlength{\qkfrontw}{\dimexpr\qkh*16/9\relax}
\newsavebox{\qkcameras}
\newsavebox{\qkmap}

\tcbset{
  qk bubble/.style={
    boxrule=0.35pt,
    arc=3pt,
    outer arc=3pt,
    boxsep=0pt,
    left=3pt,right=3pt,top=1pt,bottom=0.1pt,
    before skip=0pt,after skip=0pt
  },
  qk question/.style={
    qk bubble,
    colback=gray!9,colframe=gray!30,
    overlay={
      \path[fill=gray!9,draw=gray!30,line width=0.35pt]
        ([yshift=-4pt]frame.north west) --
        ([xshift=-3pt,yshift=-6pt]frame.north west) --
        ([yshift=-8pt]frame.north west);
    }
  },
  qk answer/.style={
    qk bubble,
    colback=teal!6,colframe=teal!30!gray,
    overlay={
      \path[fill=teal!6,draw=teal!30!gray,line width=0.35pt]
        ([yshift=-4pt]frame.north east) --
        ([xshift=3pt,yshift=-6pt]frame.north east) --
        ([yshift=-8pt]frame.north east);
    }
  }
}

\newcommand{\qa}[2]{%
  \par\noindent
  \hspace*{3pt}
  \begin{minipage}[t]{\dimexpr0.65\linewidth-6pt\relax}
    \vspace{0pt}
    \begin{tcolorbox}[qk question]
      \scriptsize\raggedright\strut #1\strut
    \end{tcolorbox}%
  \end{minipage}\hfill
  \begin{minipage}[t]{\dimexpr0.35\linewidth-6pt\relax}
    \vspace{0pt}%
    \begin{tcolorbox}[qk answer]
      \scriptsize\raggedright\strut\textit{#2}\strut
    \end{tcolorbox}%
  \end{minipage}%
  \hspace*{3pt}
  \par\vspace{1pt}}

\sbox{\qkcameras}{%
  \raisebox{-0.5\height}{\includegraphics[width=0.5\qkfrontw]{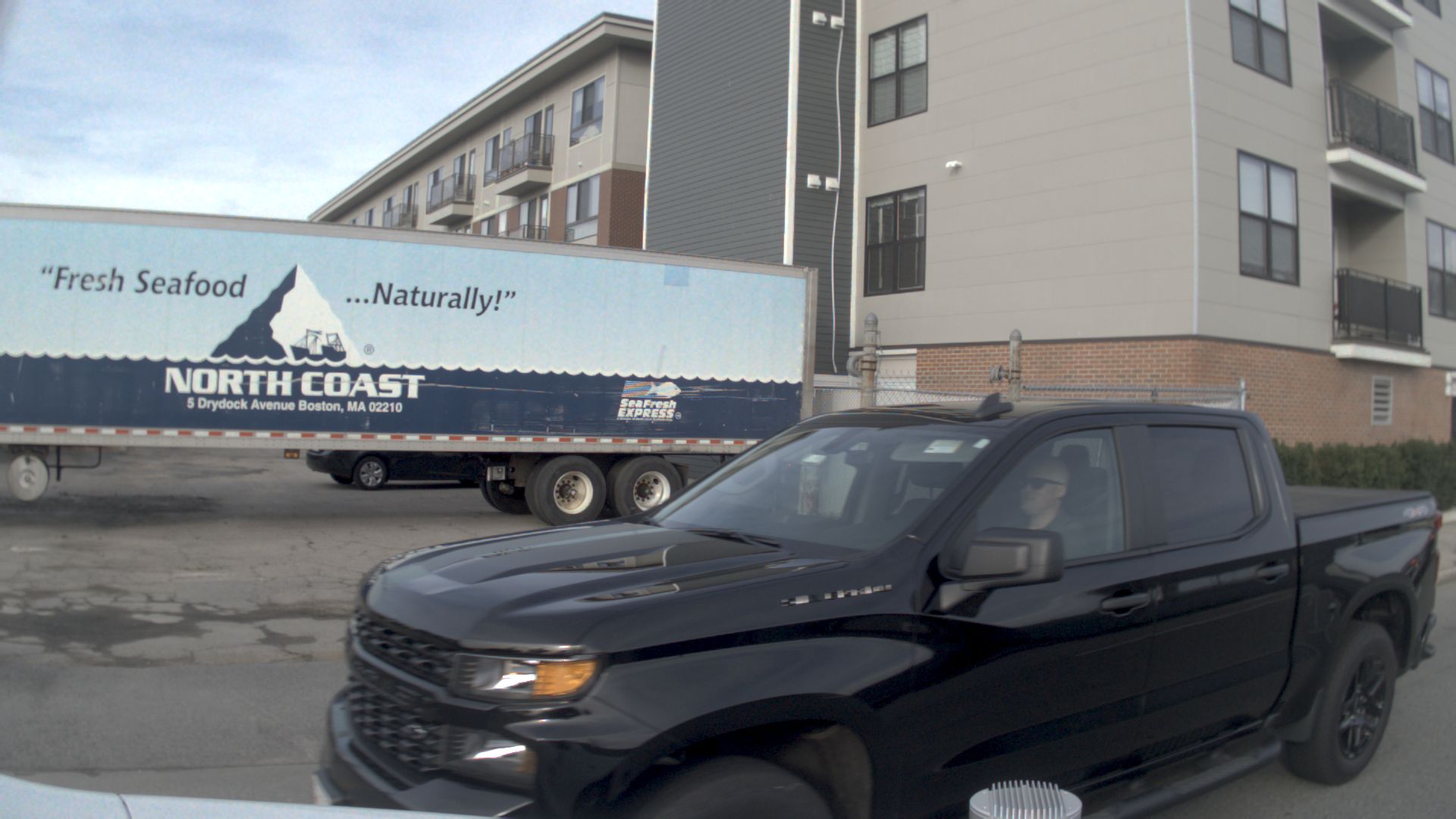}}%
  \hspace{\qkgap}%
  \raisebox{-0.5\height}{\includegraphics[width=\qkfrontw]{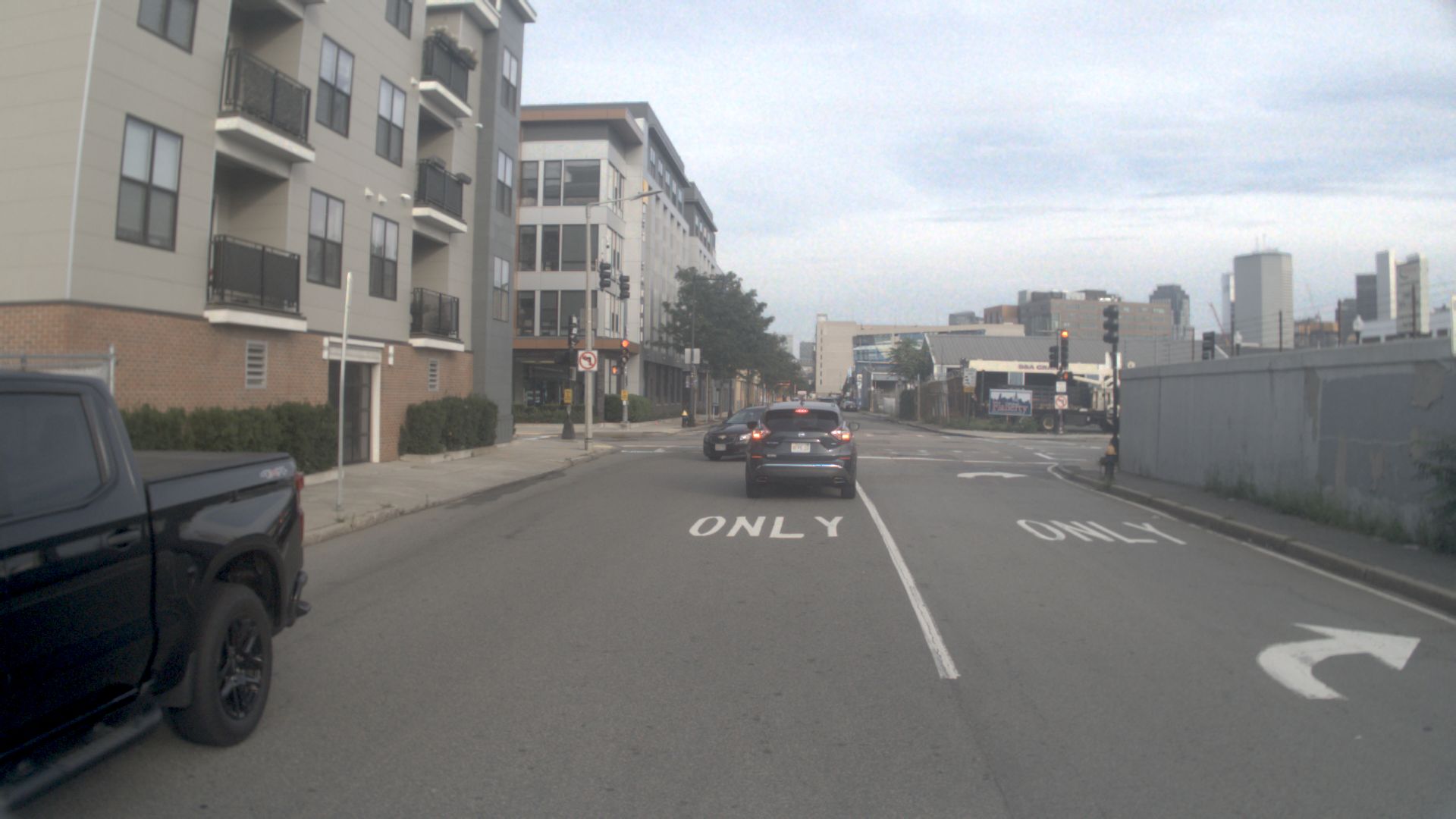}}%
  \hspace{\qkgap}%
  \raisebox{-0.5\height}{\includegraphics[width=0.5\qkfrontw]{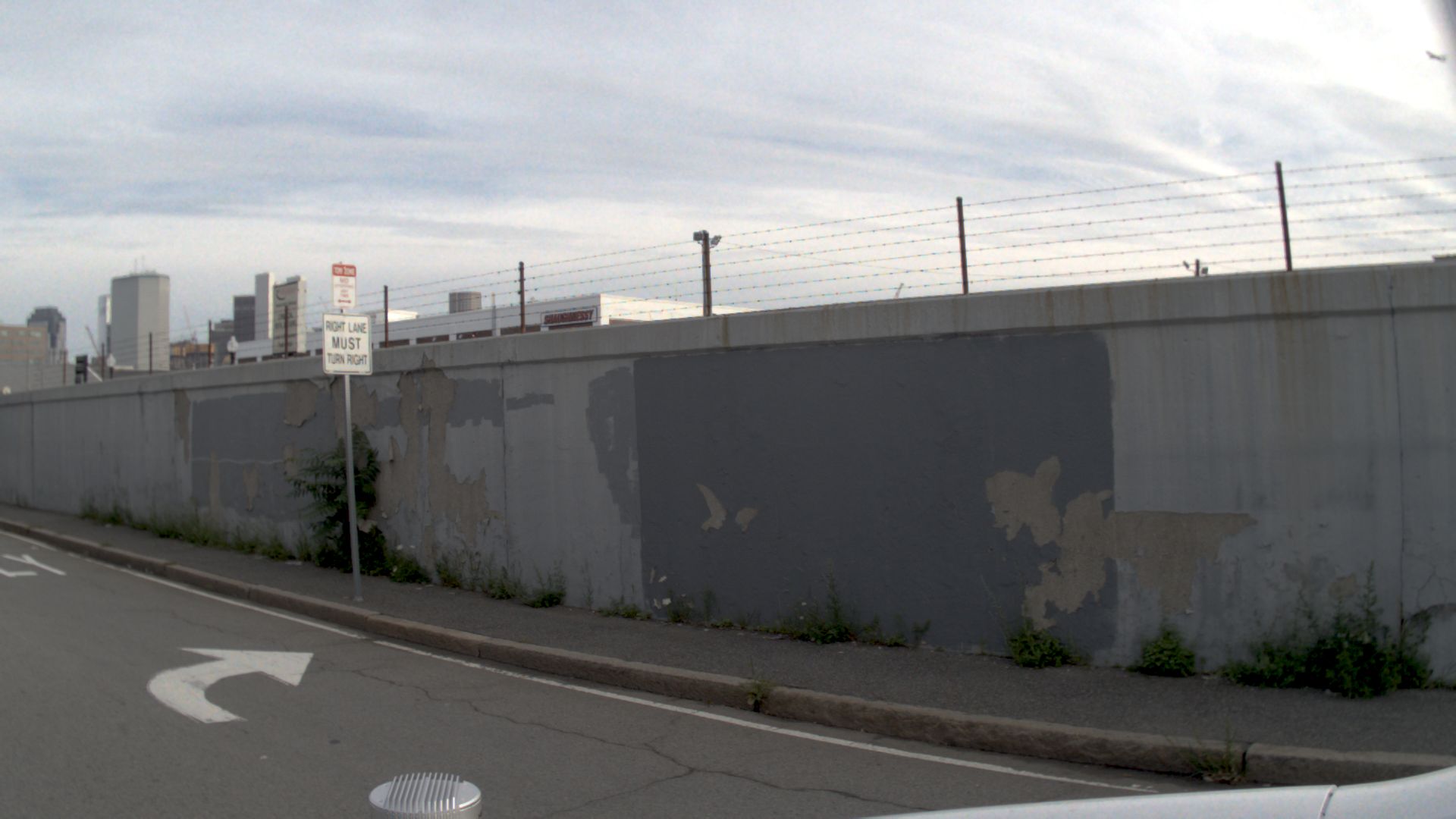}}%
}

\noindent\usebox{\qkcameras}\hspace{\qkgap}\usebox{\qkmap}\par
\vspace{2pt}
\noindent
\begin{minipage}[t]{\wd\qkcameras}
  \centering\scriptsize Cameras at $T=0$
\end{minipage}%
\hspace{\qkgap}%
\vspace{0.1cm}

\begin{minipage}[t]{\linewidth}
  \scriptsize\raggedright
  \setlength{\parskip}{0pt}
  \input{\qualknowdir/qa_1.tex}
  \input{\qualknowdir/qa_2.tex}
\end{minipage}

\caption{All KNOW question--answer pairs that our graph yields at one
held-out geographic test location in Boston, from 35 of the 40 queries that produce a valid QA pair given the scene graph.}
\label{fig:qualitative_know}
\end{figure*}

\providecommand{\qualdir}{supplementary/figures/SAY_qualitative}
\providecommand{\qualcamw}{0.3\linewidth}

\begin{figure*}[t]
\centering
\newsavebox{\qualcams}\newsavebox{\qualmap}\newsavebox{\qualreason}
\newlength{\qualgap}\setlength{\qualgap}{0.25cm}

\newcommand{\qualrow}[2]{%
  \begin{subfigure}{\linewidth}
    \centering
    \sbox{\qualcams}{\begin{minipage}[b]{\qualcamw}%
      \includegraphics[width=\linewidth]{\qualdir/#1/cam_F0.jpg}\\[1pt]%
      \includegraphics[width=\dimexpr0.5\linewidth-0.5pt\relax]{\qualdir/#1/cam_L0.jpg}\hfill%
      \includegraphics[width=\dimexpr0.5\linewidth-0.5pt\relax]{\qualdir/#1/cam_R0.jpg}%
    \end{minipage}}%
    \sbox{\qualmap}{%
      \includegraphics[height=\dimexpr\ht\qualcams+\dp\qualcams\relax]{\qualdir/#1/map.pdf}%
    }%
    \noindent
    \usebox{\qualcams}\hspace{\qualgap}%
    \usebox{\qualmap}\hspace{\qualgap}%
          \sbox{\qualreason}{%
      \begin{minipage}[b]{%
        \dimexpr\linewidth-\wd\qualcams-\wd\qualmap-2\qualgap\relax}
        \begin{tcolorbox}[
          colback=gray!8,
          colframe=gray!25,
          boxrule=0.4pt,
          arc=3pt,
          outer arc=3pt,
          boxsep=0pt,
          left=6pt,
          right=6pt,
          top=5pt,
          bottom=5pt,
          before skip=0pt,
          after skip=0pt
        ]
          \footnotesize\raggedright
          \input{\qualdir/#1/reasoning.tex}
        \end{tcolorbox}%
      \end{minipage}%
    }%
    \raisebox{%
      \dimexpr(\ht\qualmap-\dp\qualmap
        -\ht\qualreason+\dp\qualreason)/2\relax
    }{\usebox{\qualreason}}%
    \par
    \caption{#2.}
    \label{fig:qualitative-#1}
  \end{subfigure}%
}

  \begin{subfigure}{\linewidth}
    \centering
    \sbox{\qualcams}{\begin{minipage}[b]{\qualcamw}%
      \includegraphics[width=\linewidth]{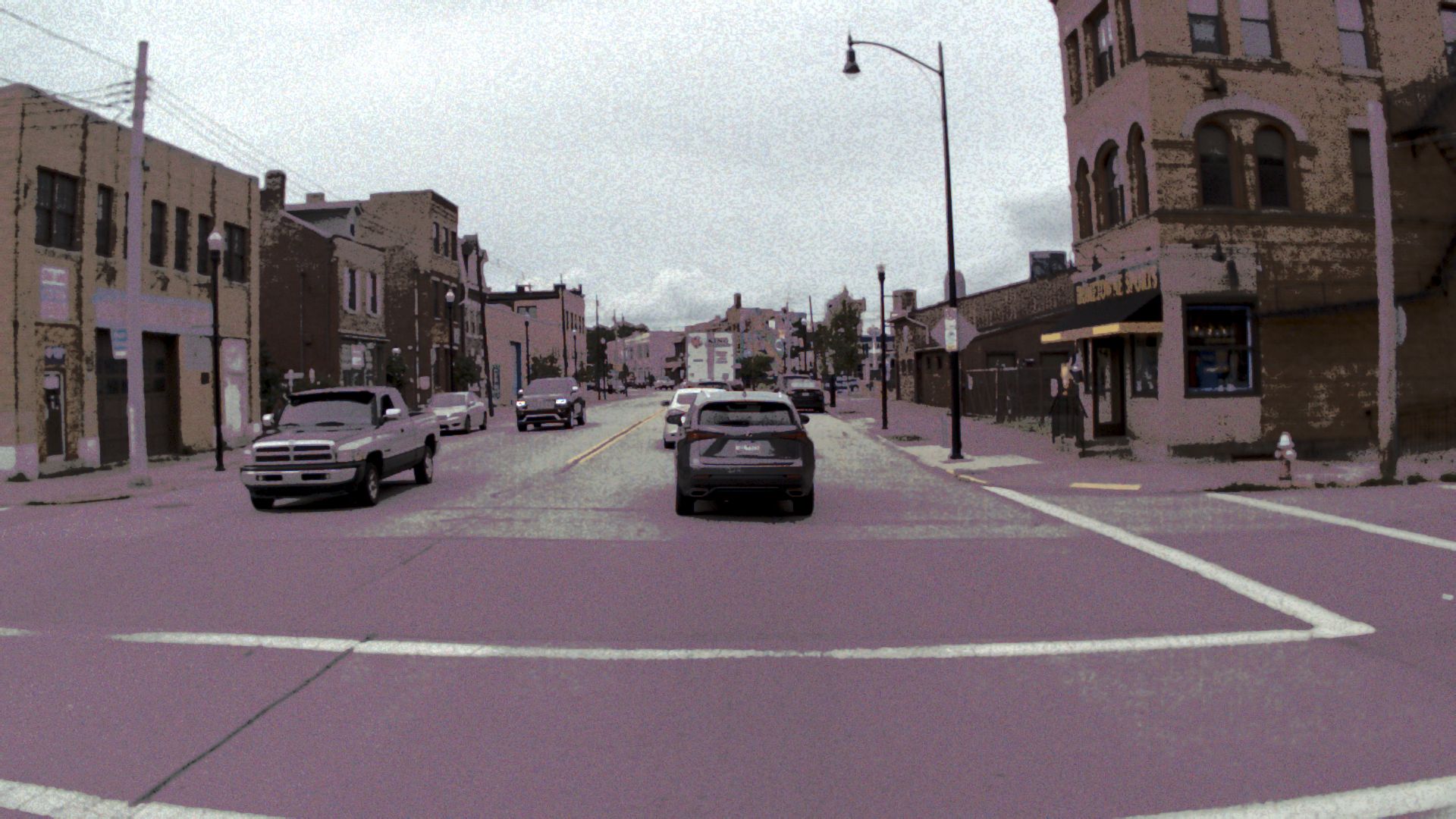}\\[1pt]%
      \includegraphics[width=\dimexpr0.5\linewidth-0.5pt\relax]{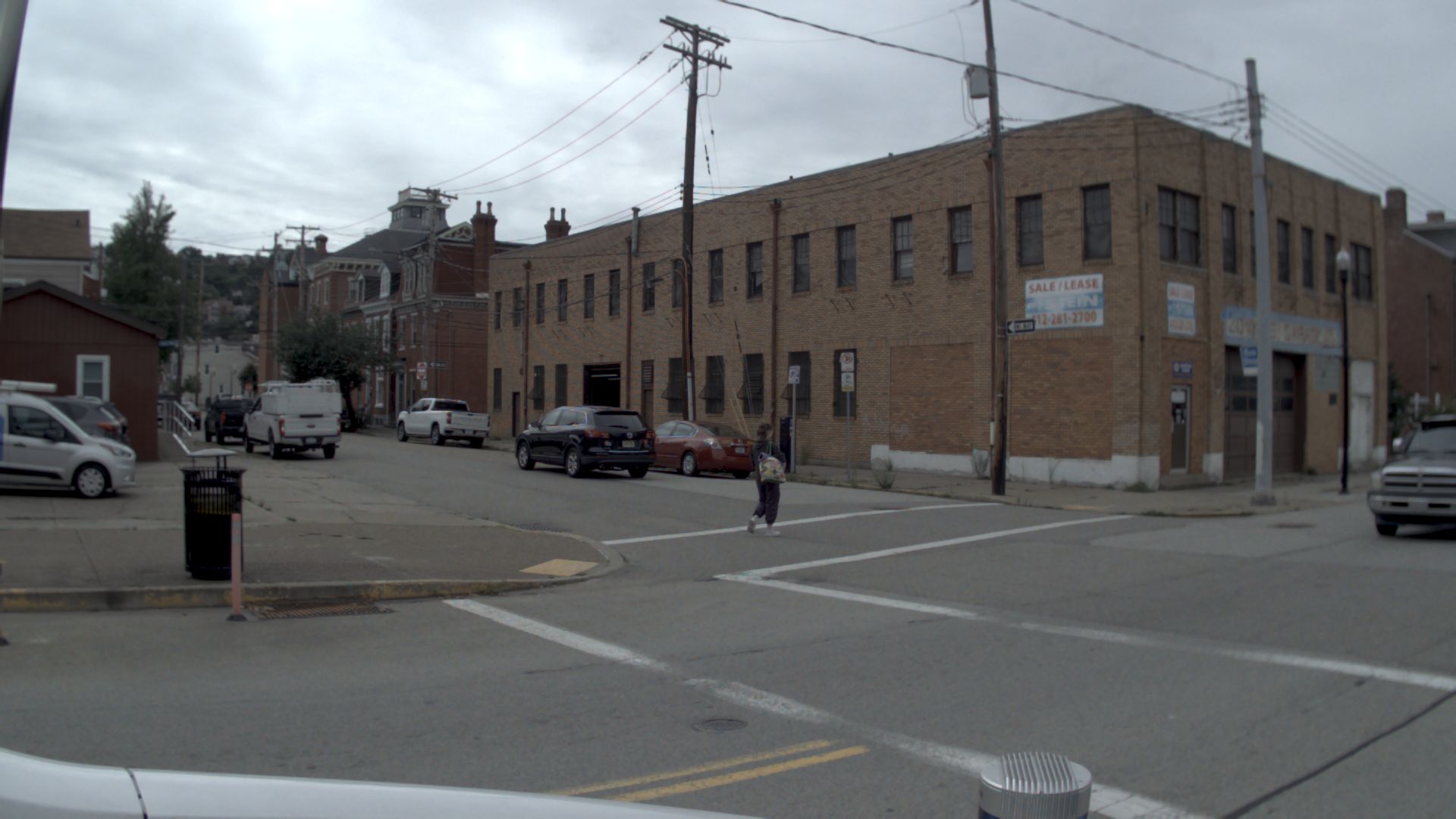}\hfill%
      \includegraphics[width=\dimexpr0.5\linewidth-0.5pt\relax]{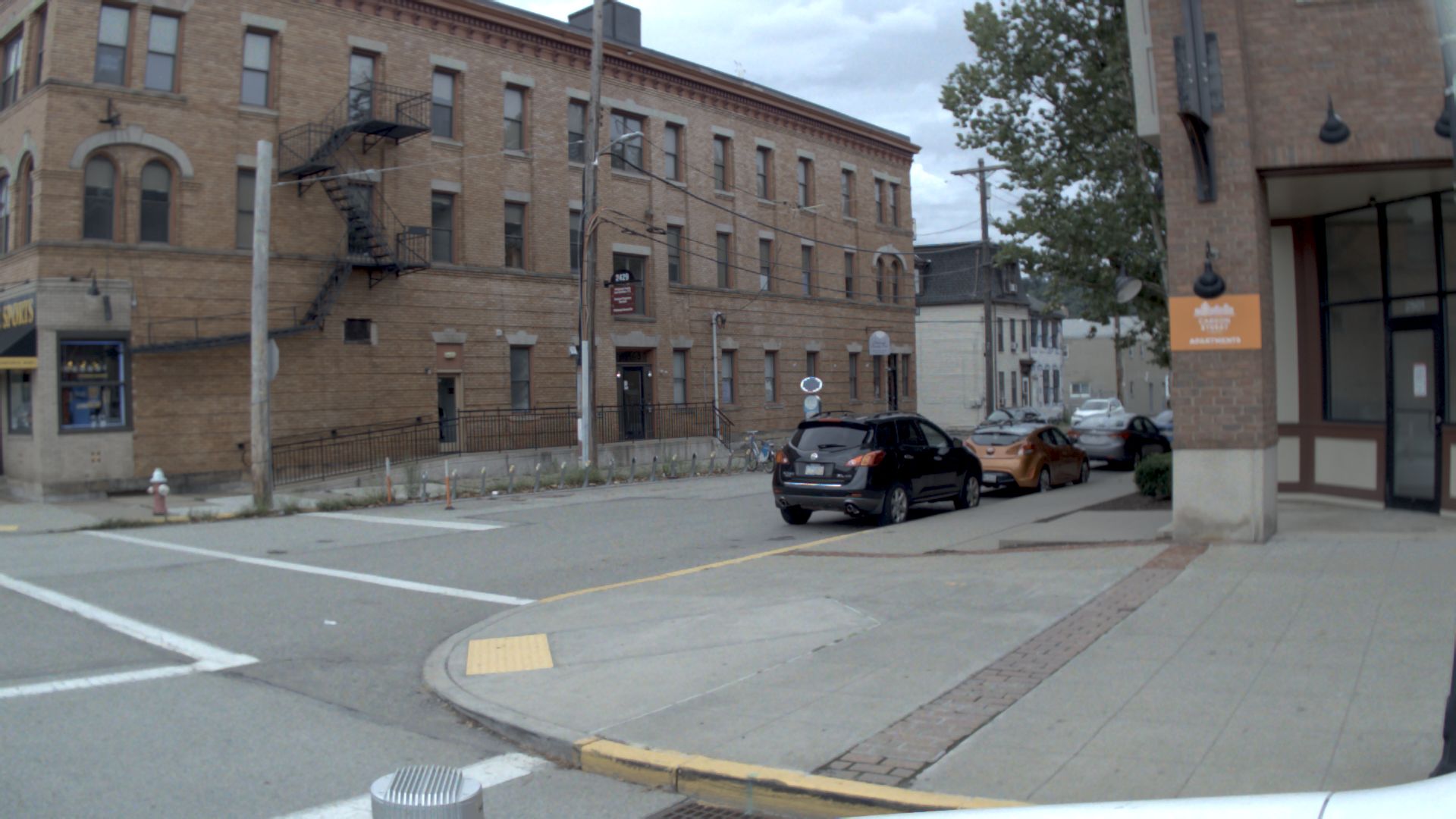}%
    \end{minipage}}%
    \sbox{\qualmap}{%
      \includegraphics[height=\dimexpr\ht\qualcams+\dp\qualcams\relax]{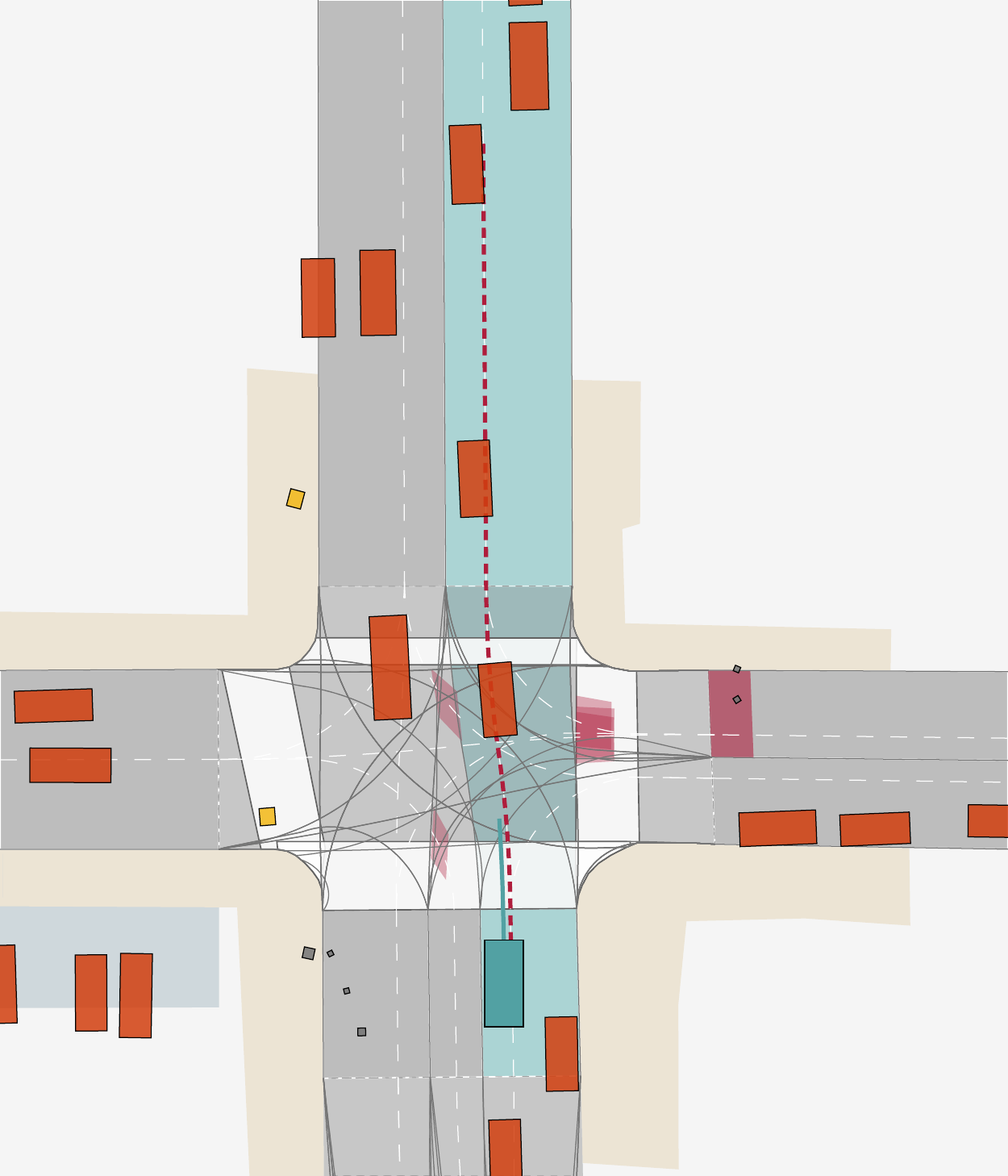}%
    }%
    \noindent
    \usebox{\qualcams}\hspace{\qualgap}%
    \usebox{\qualmap}\hspace{\qualgap}%
          \sbox{\qualreason}{%
      \begin{minipage}[b]{%
        \dimexpr\linewidth-\wd\qualcams-\wd\qualmap-2\qualgap\relax}
        \begin{tcolorbox}[
          colback=gray!8,
          colframe=gray!25,
          boxrule=0.4pt,
          arc=3pt,
          outer arc=3pt,
          boxsep=0pt,
          left=6pt,
          right=6pt,
          top=5pt,
          bottom=5pt,
          before skip=0pt,
          after skip=0pt
        ]
          \footnotesize\raggedright
          \input{\qualdir/a_pittsburgh/reasoning.tex}
        \end{tcolorbox}%
      \end{minipage}%
    }%
    \raisebox{%
      \dimexpr(\ht\qualmap-\dp\qualmap
        -\ht\qualreason+\dp\qualreason)/2\relax
    }{\usebox{\qualreason}}%
    \par
    \caption{Pittsburgh.}
    \label{fig:qualitative-a_pittsburgh}
  \end{subfigure}%
\par\vspace{\qualgap}
  \begin{subfigure}{\linewidth}
    \centering
    \sbox{\qualcams}{\begin{minipage}[b]{\qualcamw}%
      \includegraphics[width=\linewidth]{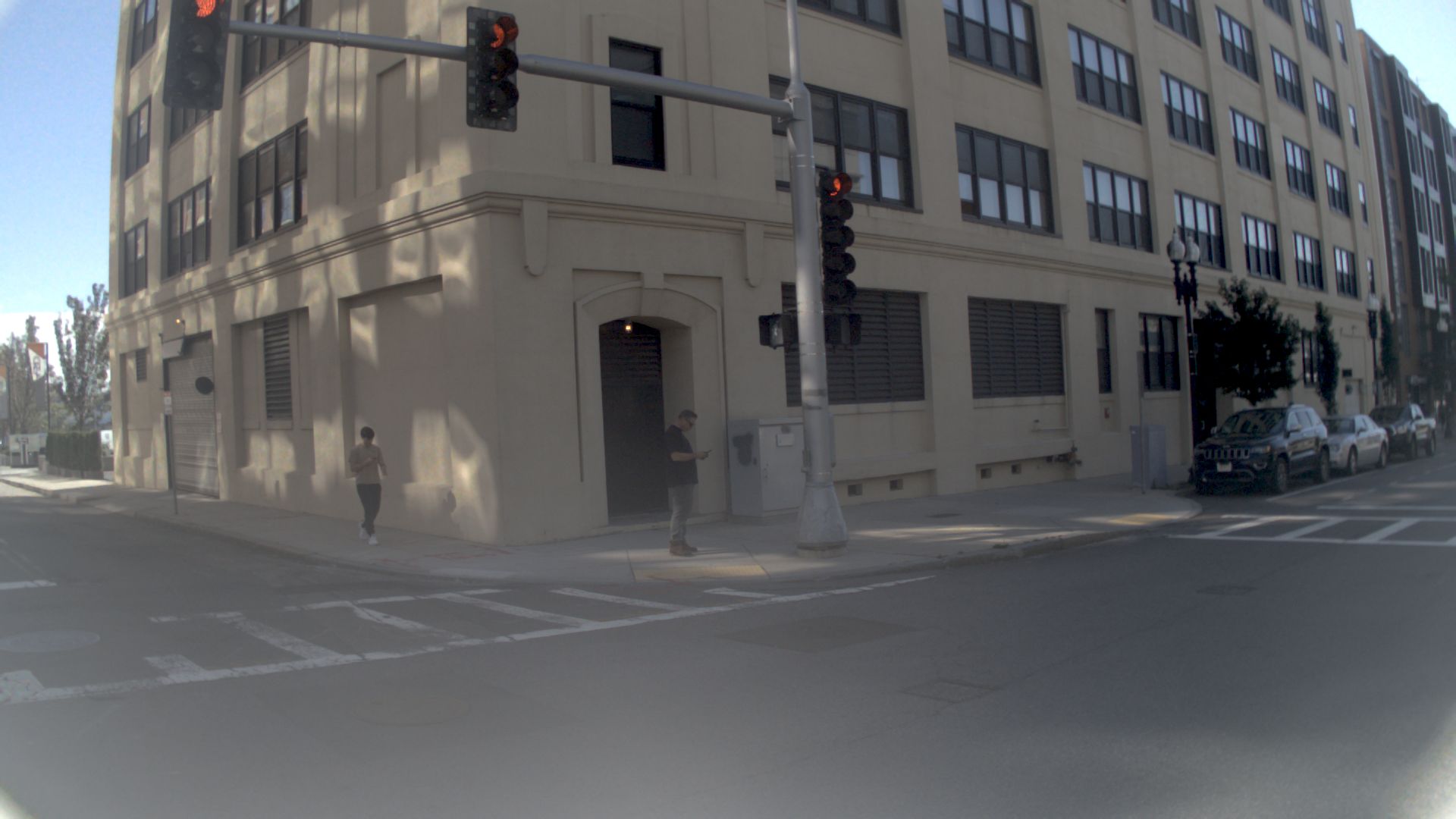}\\[1pt]%
      \includegraphics[width=\dimexpr0.5\linewidth-0.5pt\relax]{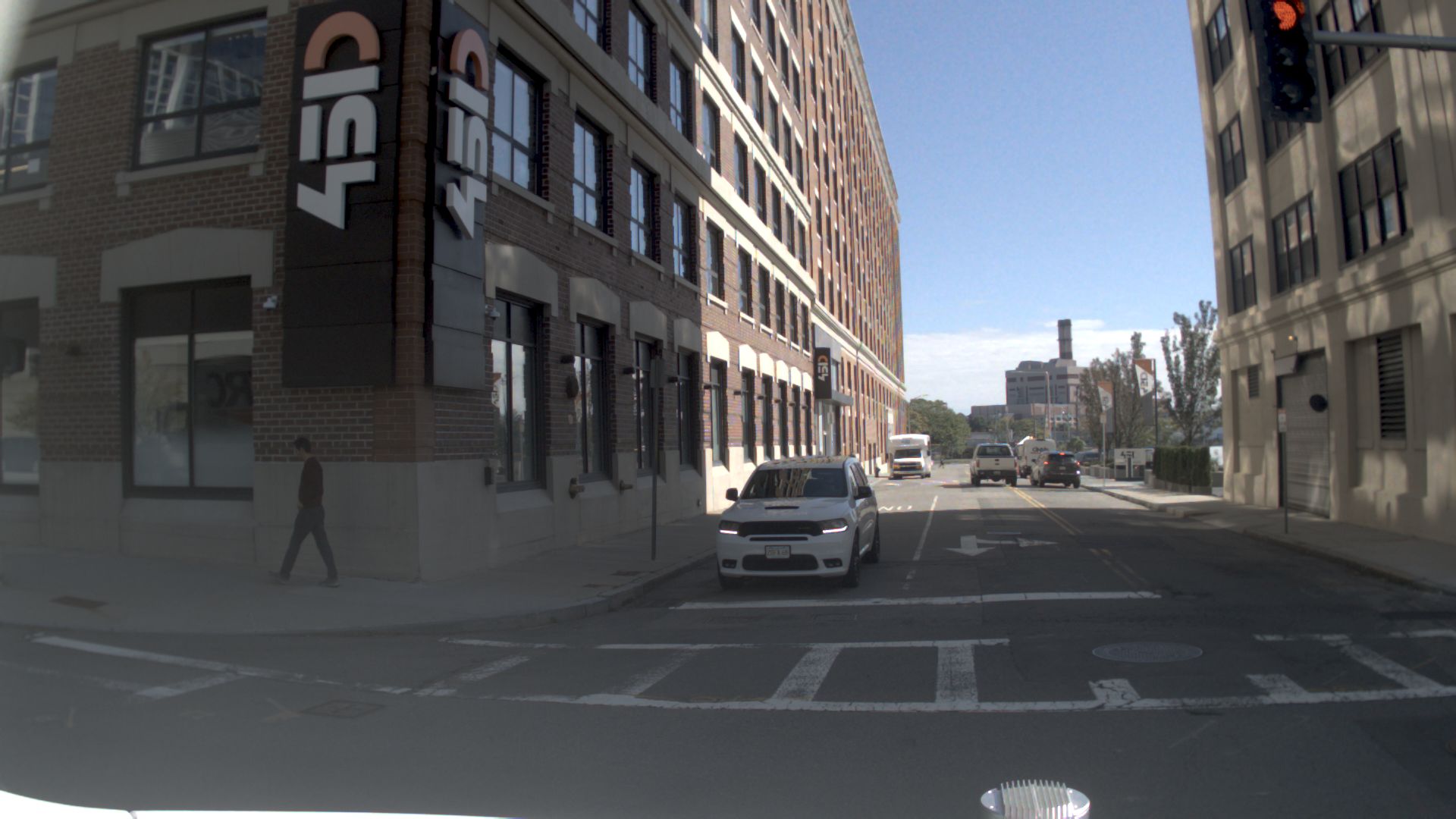}\hfill%
      \includegraphics[width=\dimexpr0.5\linewidth-0.5pt\relax]{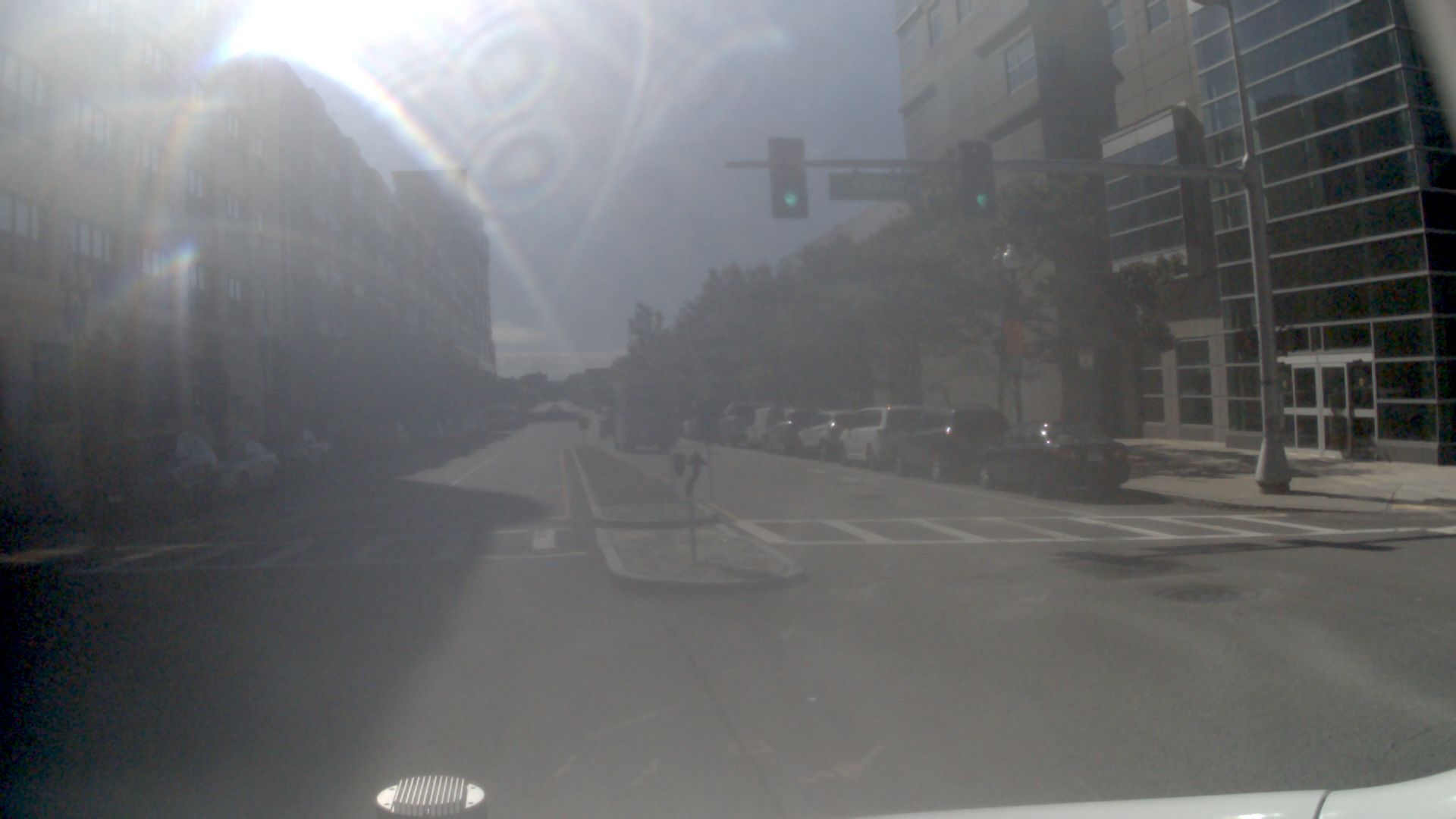}%
    \end{minipage}}%
    \sbox{\qualmap}{%
      \includegraphics[height=\dimexpr\ht\qualcams+\dp\qualcams\relax]{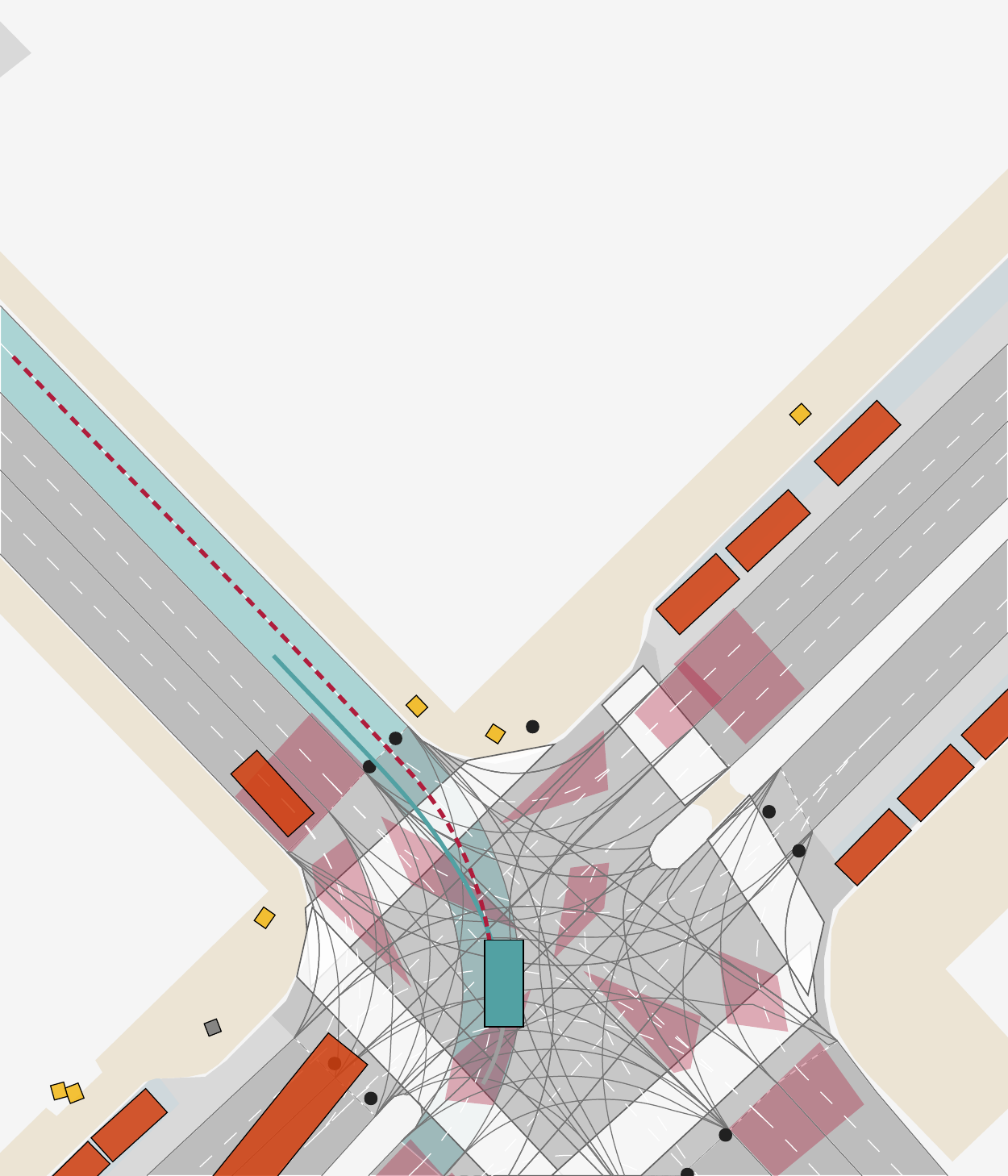}%
    }%
    \noindent
    \usebox{\qualcams}\hspace{\qualgap}%
    \usebox{\qualmap}\hspace{\qualgap}%
          \sbox{\qualreason}{%
      \begin{minipage}[b]{%
        \dimexpr\linewidth-\wd\qualcams-\wd\qualmap-2\qualgap\relax}
        \begin{tcolorbox}[
          colback=gray!8,
          colframe=gray!25,
          boxrule=0.4pt,
          arc=3pt,
          outer arc=3pt,
          boxsep=0pt,
          left=6pt,
          right=6pt,
          top=5pt,
          bottom=5pt,
          before skip=0pt,
          after skip=0pt
        ]
          \footnotesize\raggedright
          \input{\qualdir/b_boston/reasoning.tex}
        \end{tcolorbox}%
      \end{minipage}%
    }%
    \raisebox{%
      \dimexpr(\ht\qualmap-\dp\qualmap
        -\ht\qualreason+\dp\qualreason)/2\relax
    }{\usebox{\qualreason}}%
    \par
    \caption{Boston.}
    \label{fig:qualitative-b_boston}
  \end{subfigure}%
\par\vspace{\qualgap}
  \begin{subfigure}{\linewidth}
    \centering
    \sbox{\qualcams}{\begin{minipage}[b]{\qualcamw}%
      \includegraphics[width=\linewidth]{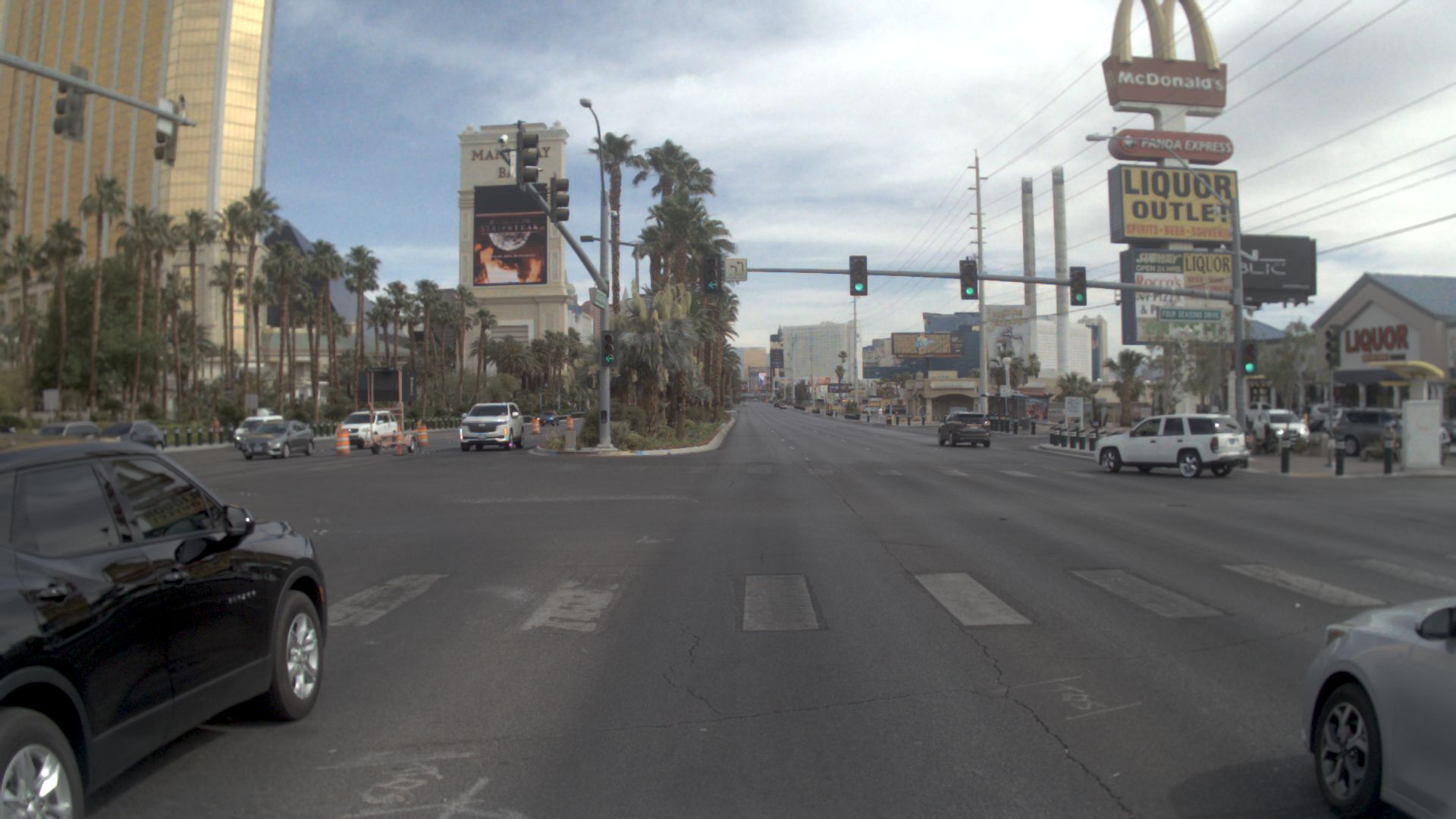}\\[1pt]%
      \includegraphics[width=\dimexpr0.5\linewidth-0.5pt\relax]{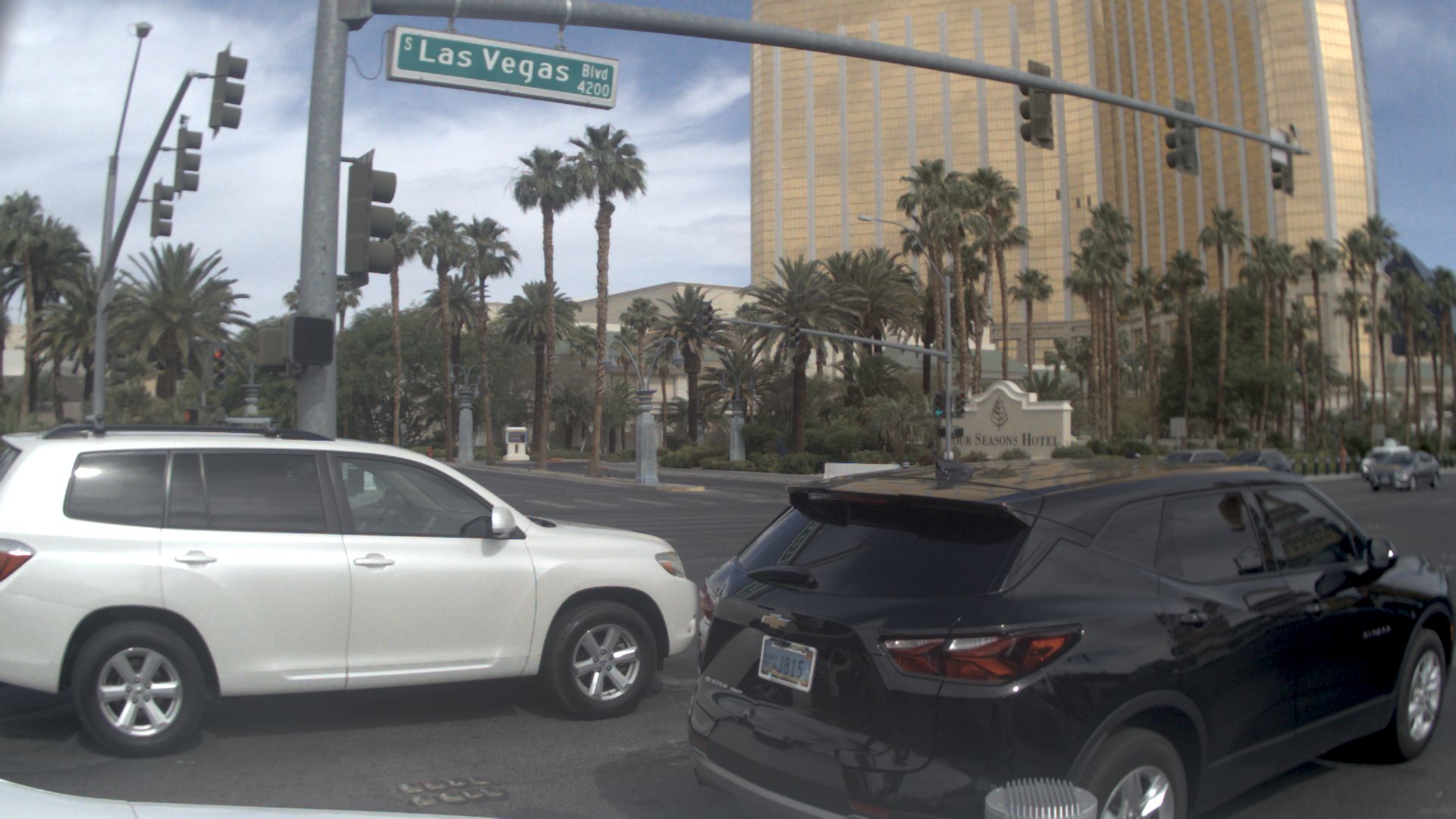}\hfill%
      \includegraphics[width=\dimexpr0.5\linewidth-0.5pt\relax]{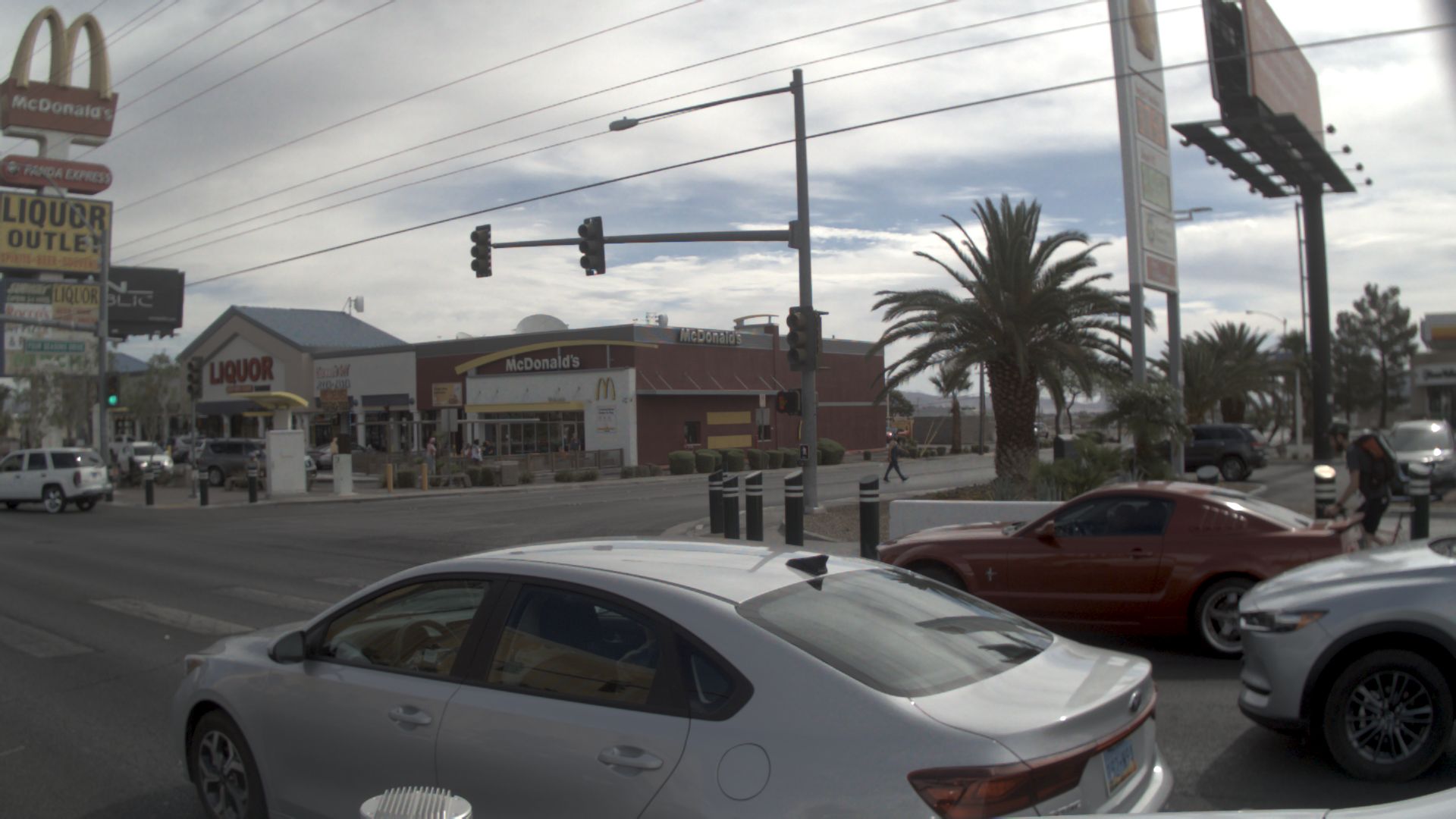}%
    \end{minipage}}%
    \sbox{\qualmap}{%
      \includegraphics[height=\dimexpr\ht\qualcams+\dp\qualcams\relax]{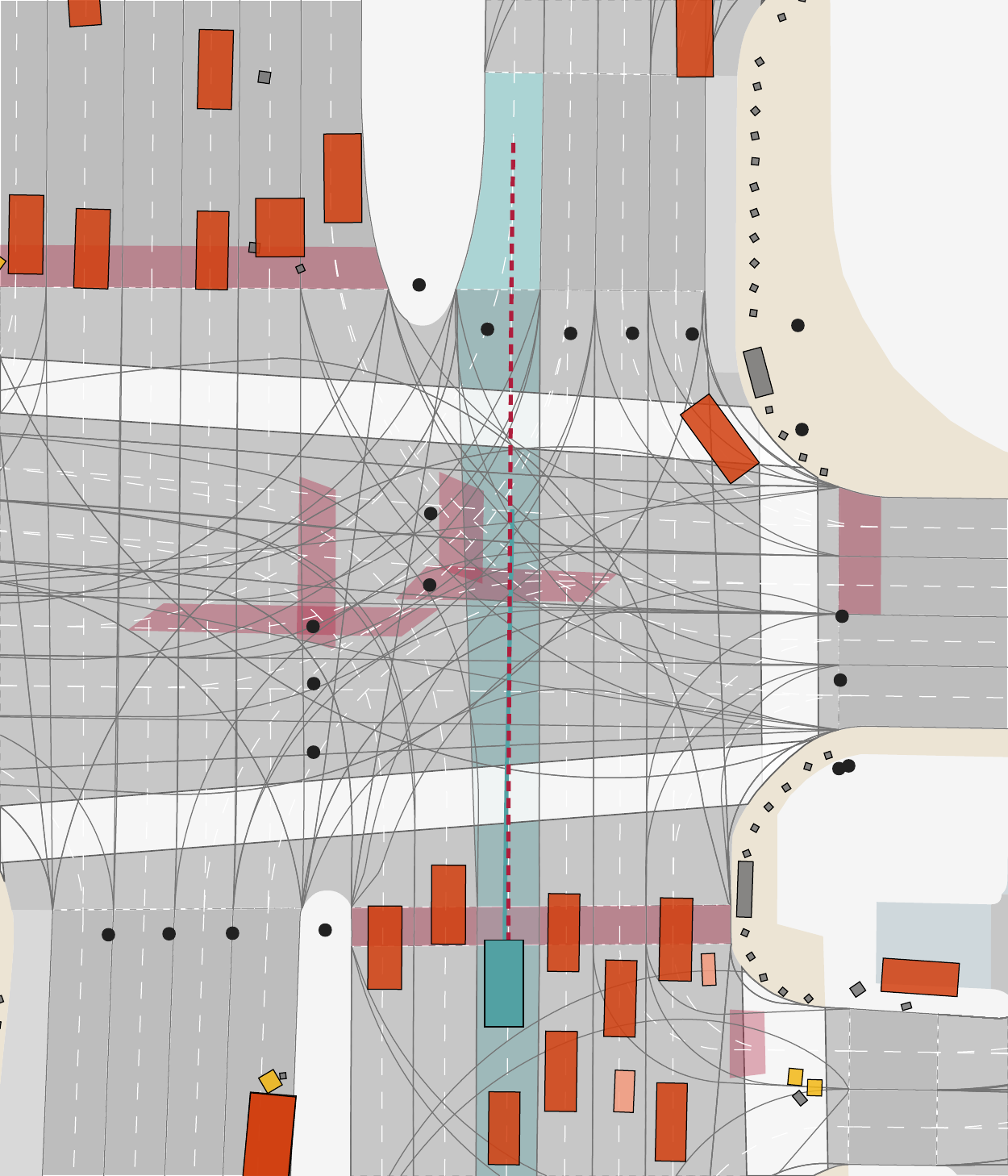}%
    }%
    \noindent
    \usebox{\qualcams}\hspace{\qualgap}%
    \usebox{\qualmap}\hspace{\qualgap}%
          \sbox{\qualreason}{%
      \begin{minipage}[b]{%
        \dimexpr\linewidth-\wd\qualcams-\wd\qualmap-2\qualgap\relax}
        \begin{tcolorbox}[
          colback=gray!8,
          colframe=gray!25,
          boxrule=0.4pt,
          arc=3pt,
          outer arc=3pt,
          boxsep=0pt,
          left=6pt,
          right=6pt,
          top=5pt,
          bottom=5pt,
          before skip=0pt,
          after skip=0pt
        ]
          \footnotesize\raggedright
          \input{\qualdir/c_lasvegas/reasoning.tex}
        \end{tcolorbox}%
      \end{minipage}%
    }%
    \raisebox{%
      \dimexpr(\ht\qualmap-\dp\qualmap
        -\ht\qualreason+\dp\qualreason)/2\relax
    }{\usebox{\qualreason}}%
    \par
    \caption{Las Vegas.}
    \label{fig:qualitative-c_lasvegas}
  \end{subfigure}%
\par\vspace{\qualgap}
  \begin{subfigure}{\linewidth}
    \centering
    \sbox{\qualcams}{\begin{minipage}[b]{\qualcamw}%
      \includegraphics[width=\linewidth]{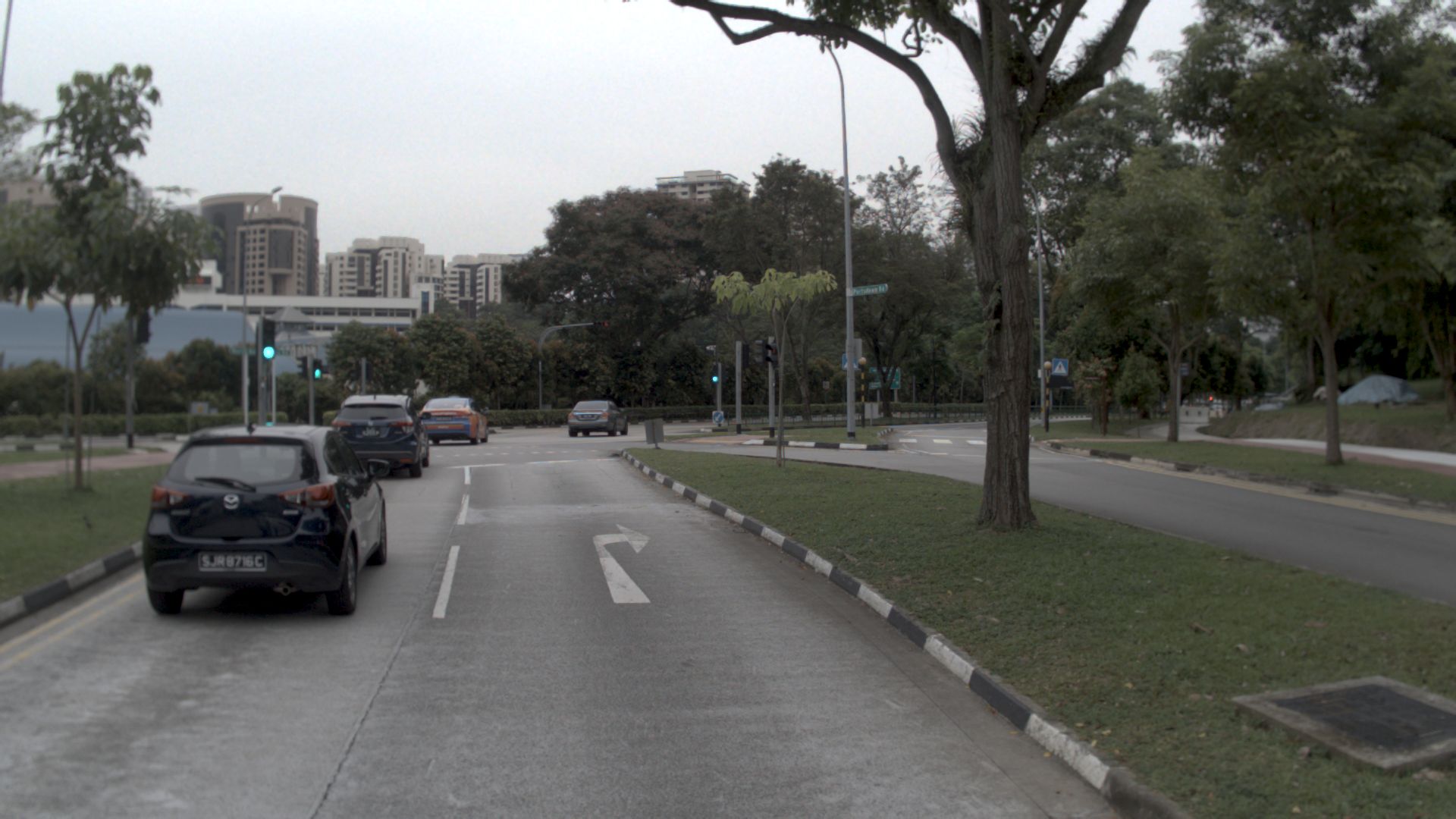}\\[1pt]%
      \includegraphics[width=\dimexpr0.5\linewidth-0.5pt\relax]{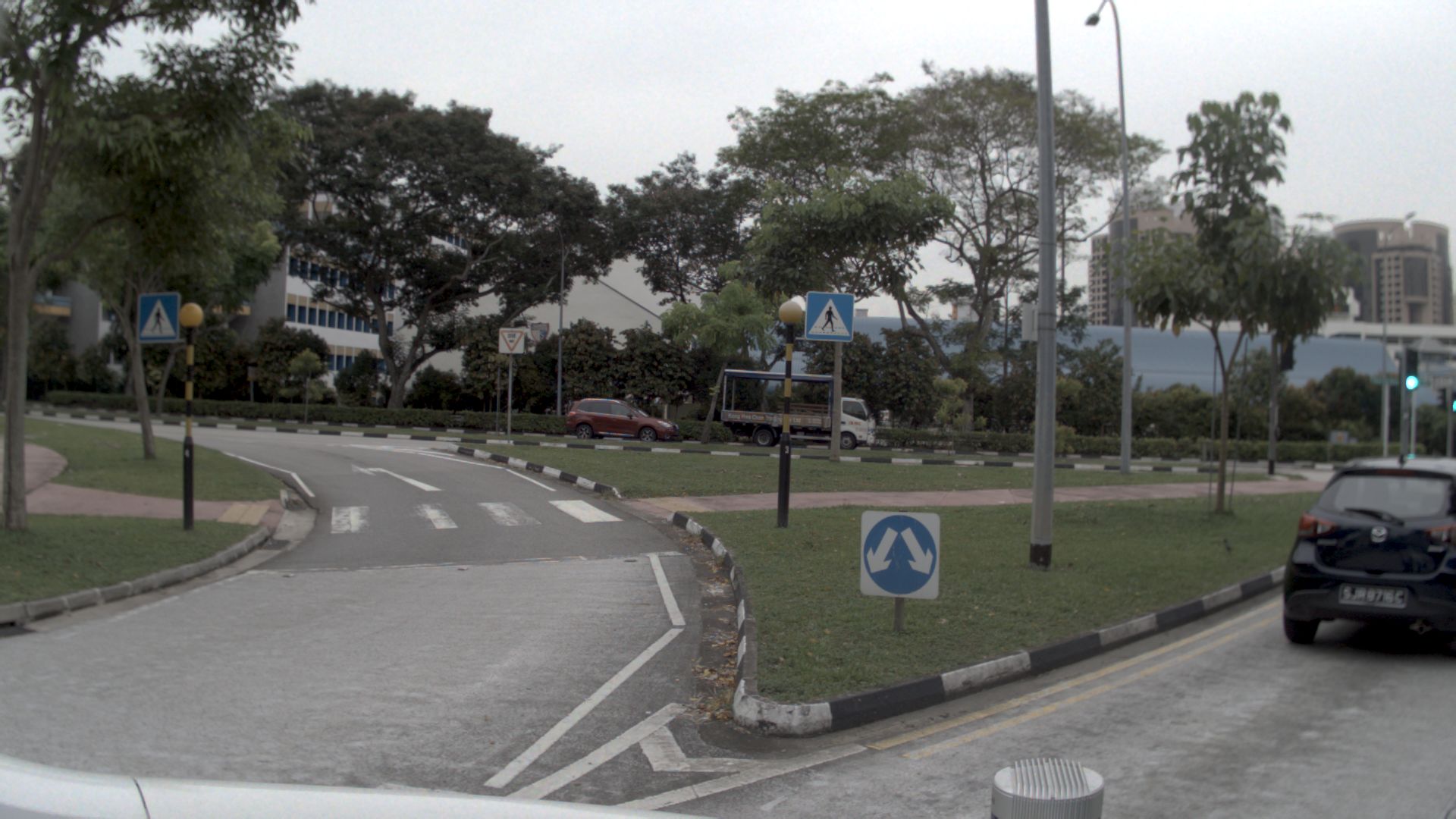}\hfill%
      \includegraphics[width=\dimexpr0.5\linewidth-0.5pt\relax]{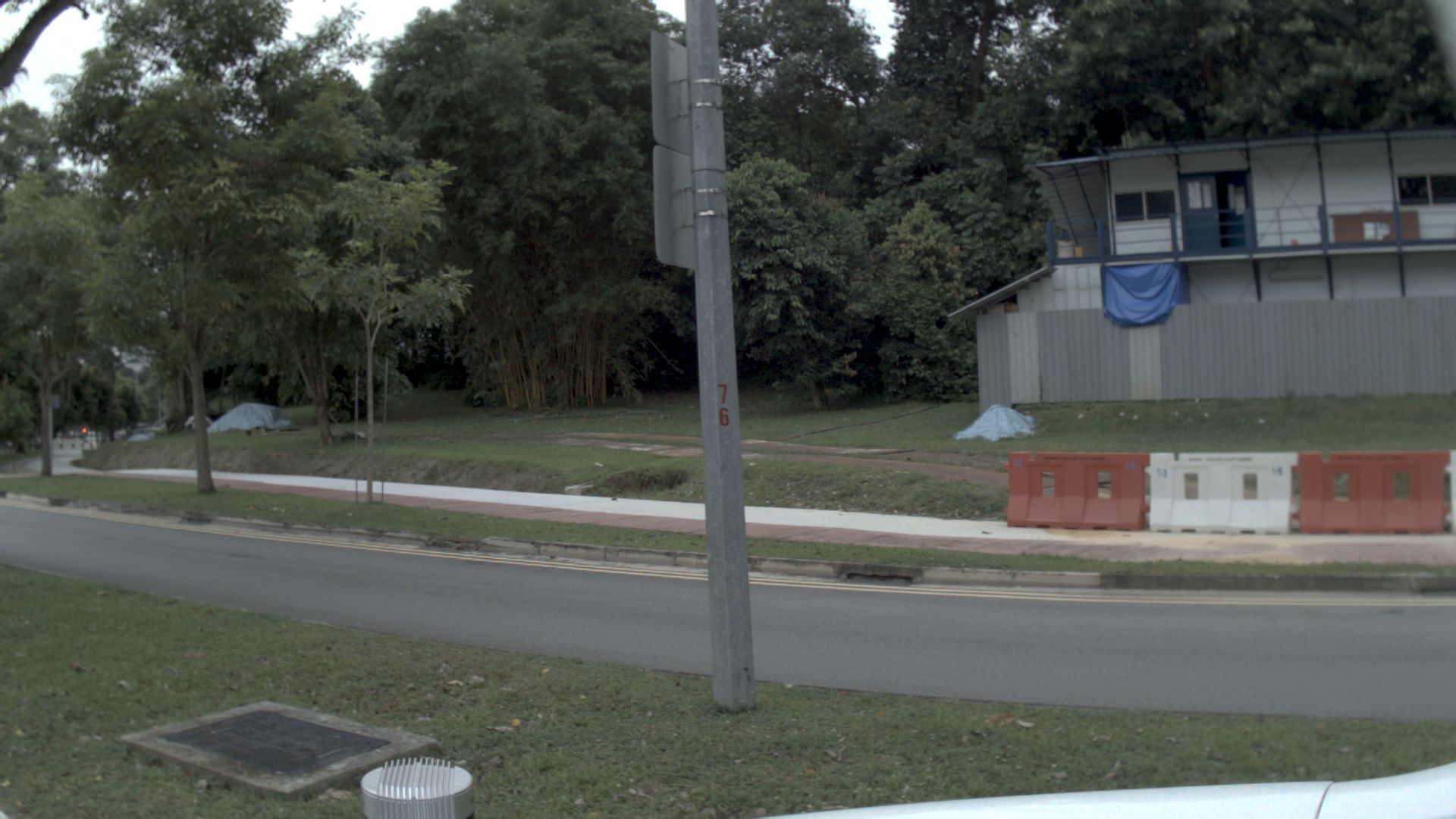}%
    \end{minipage}}%
    \sbox{\qualmap}{%
      \includegraphics[height=\dimexpr\ht\qualcams+\dp\qualcams\relax]{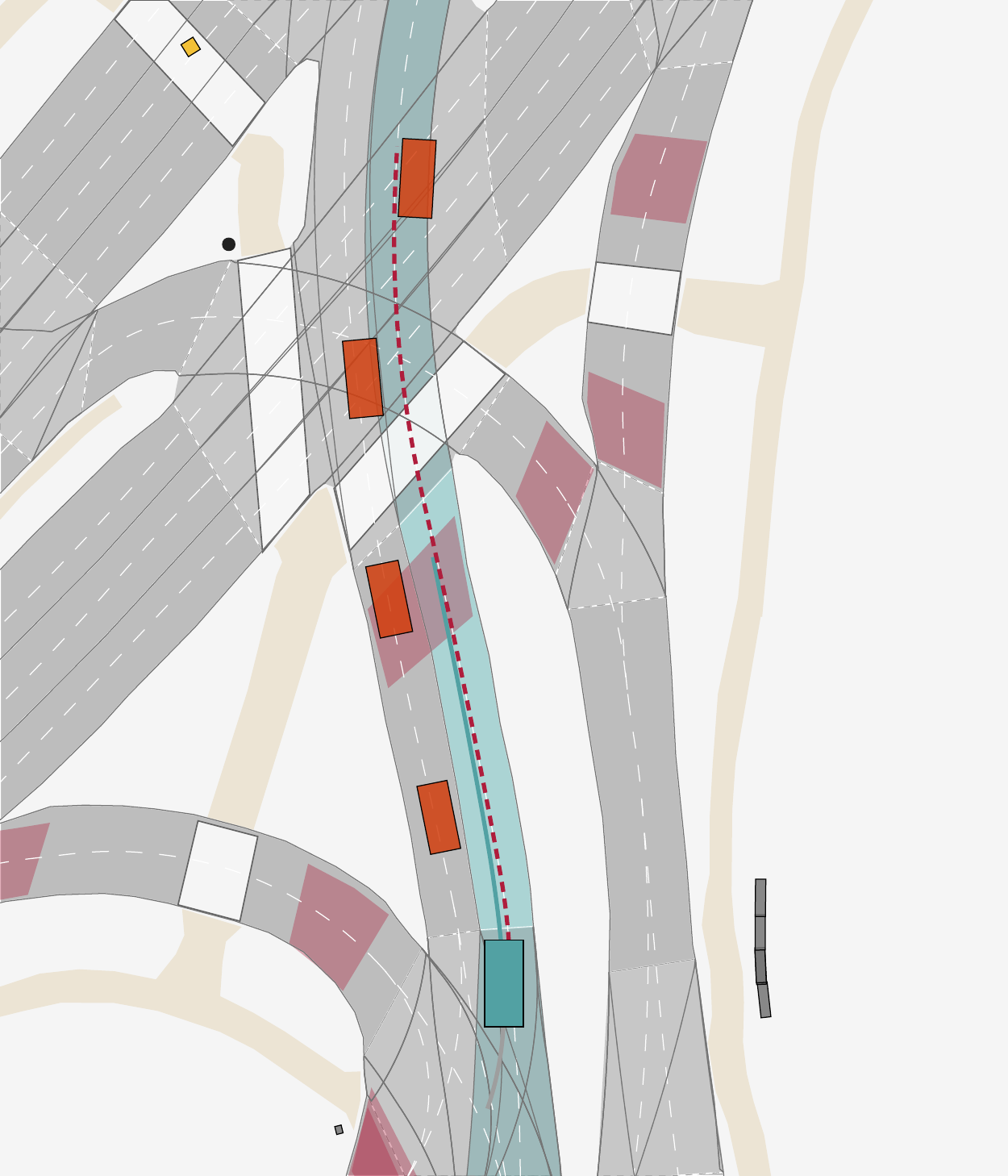}%
    }%
    \noindent
    \usebox{\qualcams}\hspace{\qualgap}%
    \usebox{\qualmap}\hspace{\qualgap}%
          \sbox{\qualreason}{%
      \begin{minipage}[b]{%
        \dimexpr\linewidth-\wd\qualcams-\wd\qualmap-2\qualgap\relax}
        \begin{tcolorbox}[
          colback=gray!8,
          colframe=gray!25,
          boxrule=0.4pt,
          arc=3pt,
          outer arc=3pt,
          boxsep=0pt,
          left=6pt,
          right=6pt,
          top=5pt,
          bottom=5pt,
          before skip=0pt,
          after skip=0pt
        ]
          \footnotesize\raggedright
          \input{\qualdir/d_singapore/reasoning.tex}
        \end{tcolorbox}%
      \end{minipage}%
    }%
    \raisebox{%
      \dimexpr(\ht\qualmap-\dp\qualmap
        -\ht\qualreason+\dp\qualreason)/2\relax
    }{\usebox{\qualreason}}%
    \par
    \caption{Singapore.}
    \label{fig:qualitative-d_singapore}
  \end{subfigure}%

\caption{Road-KSU's SAY output with cameras at $T=0$ (front; front-left and front-right below) and HD map with actor annotation input. Right: the navigation instruction and the SAY reasoning rendered from the scene graph.}
\label{fig:qualitative_say}
\end{figure*}

\subsection{Quality checks}\label{sec:quality_checks}

\begin{figure*}[t]
\newcommand{\qchead}[1]{\par\vspace{2pt}\noindent{\normalfont\sffamily\bfseries\scriptsize\color{black!70}#1}\par\vspace{1pt}}
\newcommand{\qcline}[1]{\par\noindent#1\par}
\newcommand{\qcitem}[1]{\par\hangindent=1em\hangafter=1\noindent\makebox[1em][l]{-}#1\par}
\newcommand{\qccat}[2]{\par\hangindent=2em\hangafter=1\noindent\hspace*{1em}#1: #2\par}
\newcommand{\qcgap}{\par\vspace{3pt}}
\begin{tcolorbox}[colback=black!2, colframe=black!45, boxrule=0.5pt, arc=2pt, left=5pt, right=5pt, top=3pt, bottom=4pt,
    fonttitle=\small\bfseries, coltitle=white, colbacktitle=black!55, title={KNOW check: annotator instructions}]
\footnotesize\ttfamily\raggedright\frenchspacing
\begin{minipage}[t]{0.485\linewidth}\raggedright
\qchead{Task:}
\qcline{Evidence: front-left, front and front-right cameras at T=0, earlier frames, navigation and speed.}
\qchead{Is the generated answer reasonable for the question as asked?}
\qccat{Yes}{the answer is confirmed by the scene.}
\qccat{No}{the answer is incorrect given the scene.}
\qcline
\qcline{If no, choose every error category that applies:}
\qcitem{Annotation error: dataset labels are wrong (object or traffic light missing, misclassified, misplaced or in the wrong state).}
\qcitem{Not visible: what the answer depends on cannot be seen in the displayed cameras.}
\qcitem{HD map wrong: lanes, crosswalks, stop lines or lane connections do not match the scene.}
\qcitem{Generator logic: the inputs look right, but the answer was derived wrongly.}
\qcitem{Wording: the question or answer is ambiguous, imprecise or misleading.}
\qcitem{Other: describe it in the note.}
\end{minipage}\hfill
\begin{minipage}[t]{0.485\linewidth}\raggedright
\qchead{How the questions are meant}
\qcitem{Ego is the vehicle carrying the cameras. Left, right and ahead are relative to ego's heading at T=0.}
\qcitem{Distances are measured from the driver's seat to the closest point of the object in question (e.g., intersection entrance, lead vehicle).}
\qcitem{Distances are measured along the road ego is driving, from ego's position at T=0. Nothing behind ego counts.}
\qcitem{The question only refers to what is ahead in a 50 m window. Additionally, shorter windows of 40 m are valid for pedestrians, cyclists and crosswalks.}
\qcitem{`Straight' means continuing on ego's current road, however its geometry bends.}
\qcitem{An intersection is any place where another road joins ego's road, including where ego has the right of way and only the side road must stop or yield.}
\qcitem{Questions about maneuvers ask what the road layout allows, not whether other road users or the current traffic light permit it now.}
\qcitem{Bike lanes are not lanes.}
\qcitem{Bi- and motorcyclists are not lead vehicles.}
\qcitem{Curvature and lead vehicles are always considered along ego's route.}
\end{minipage}
\end{tcolorbox}
\vspace{2pt}
\begin{tcolorbox}[colback=black!2, colframe=black!45, boxrule=0.5pt, arc=2pt, left=5pt, right=5pt, top=3pt, bottom=4pt,
    fonttitle=\small\bfseries, coltitle=white, colbacktitle=black!55, title={SAY check: annotator instructions}]
\footnotesize\ttfamily\raggedright\frenchspacing
\begin{minipage}[t]{0.485\linewidth}\vspace{0pt}\raggedright
\qchead{Task:}
\qcline{Evidence: front-left, front and front-right cameras at T=0, earlier frames, navigation and speed.}
\qcline{Map: the HD map around ego, the road users the CoT was built from and the path driven over the next 10 s.}
\qchead{Is this CoT reasonable?}
\qccat{Yes}{nothing in it is factually wrong.}
\qccat{No}{at least one statement is factually wrong. Mark each wrong statement.}
\qcgap
\qchead{Judging rule}
\qcline{A statement is wrong only if the cameras, the map, the driven path, the navigation instruction or the speed contradict it. Wording and omissions are not errors. The CoT is derived from map and tracking data that the cameras do not always show, so a statement that cannot be confirmed visually is not an error. Distances are computed from the HD map and do not need to be checked in this review.}
\end{minipage}\hfill
\begin{minipage}[t]{0.485\linewidth}\vspace{0pt}\raggedright
\qchead{Examples}
\qcitem{`The traffic light governing my route is red.' The light for my lane is green. $\rightarrow$ No: mark this statement.}
\qcitem{`I need to turn left, which I can do from my lane.' Navigation says left; no view shows the lane markings. $\rightarrow$ Yes: visual confirmation is not required.}
\qcitem{`There is a lead vehicle in my lane, 10-20 m ahead.' It looks more like 25 m. $\rightarrow$ Yes: distances are not checked, as they are derived exactly from the HD map.}
\end{minipage}
\end{tcolorbox}
\caption{Annotator instructions of the KNOW check (top) and the SAY check (bottom).}
\label{fig:qc_instructions}
\end{figure*}

Road-KSU derives the KNOW question--answer pairs and SAY reasoning traces described above automatically from the scene graph. To verify that our automatic generations are of high quality and align with human scene judgment, we audit random samples of both KNOW and SAY by hand, since the dataset scale precludes a full review. For KNOW, annotators assess whether the generated answer is reasonable for the question given the camera observations. For SAY, annotators check whether any statement in the trace contradicts the available scene evidence. The ratings were provided by three members of our research team, who took part voluntarily, knew how their ratings would be used, and agreed to their use. No personal information about the raters was collected or reported, and no one outside the team took part. Since the raters assessed data and were not themselves studied, no ethics board review was carried out.

\paragraph{KNOW check}
We draw 400 KNOW questions in two parts, each uniformly at random without replacement: 200 from all questions generated on GEOTEST and 200 from those generated on NAVTEST. Two expert annotators judge each question according to the instructions presented in \Cref{fig:qc_instructions}, top. The QAs are posed in single-choice format and comprise the generated answer and up to three other answers that the identical question receives elsewhere in the test set, drawn uniformly at random and shown in the same shuffled order to both annotators. The two annotators agree on 384 of the 400 questions (96.0\%, Gwet's AC1 0.96). A third annotator judges the remaining 16, and the majority of the three judgments is final. After this resolution, 387 of the 400 generated answers are judged reasonable. The answers derived from the scene graph thus hold up against the camera images for all but 13 of the sampled questions (3.25\%), with annotation errors in the HD maps and visibility flagged as the most common issues.

\paragraph{SAY check}
We draw 400 decision points uniformly at random without replacement from the union of NAVTRAIN and GEOTRAIN. Two annotators independently judged whether each of the randomly sampled reasoning traces contained a statement contradicted by the cameras, HD map, driven path, or navigation instruction, following the guidelines in \Cref{fig:qc_instructions}, bottom. Note that since SAY supervision is derived from privileged HD-map and actor information that the cameras do not always show, the raters were advised to count a statement as wrong only if this evidence contradicts it, not if the camera images merely fail to confirm it. Wording and distances were not judged, since they are calculated directly from the ground truth maps. The raters agreed on 94.5\% of the complete CoT verdicts (Gwet's AC1 = 0.94). A third annotator resolved the 22 disagreements, leaving 369 of the 400 CoTs (92.3\%) without a single wrong statement. Notably, since one reasoning trace contains several clauses, only 36 of 2,376 rated clauses (1.5\%) were wrong, mostly traffic-light states, permitted lane movements and intersection control stemming from incorrect HD map annotations and light states.

Taken together, both checks confirm that deriving KNOW and SAY supervision from the scene graph with Road-KSU yields reliable annotations, with errors confined to a small fraction of answers and to individual clauses.

\section{Test set distributions}
\label{sec:nat_bal_stats}

\begin{figure*}[t]
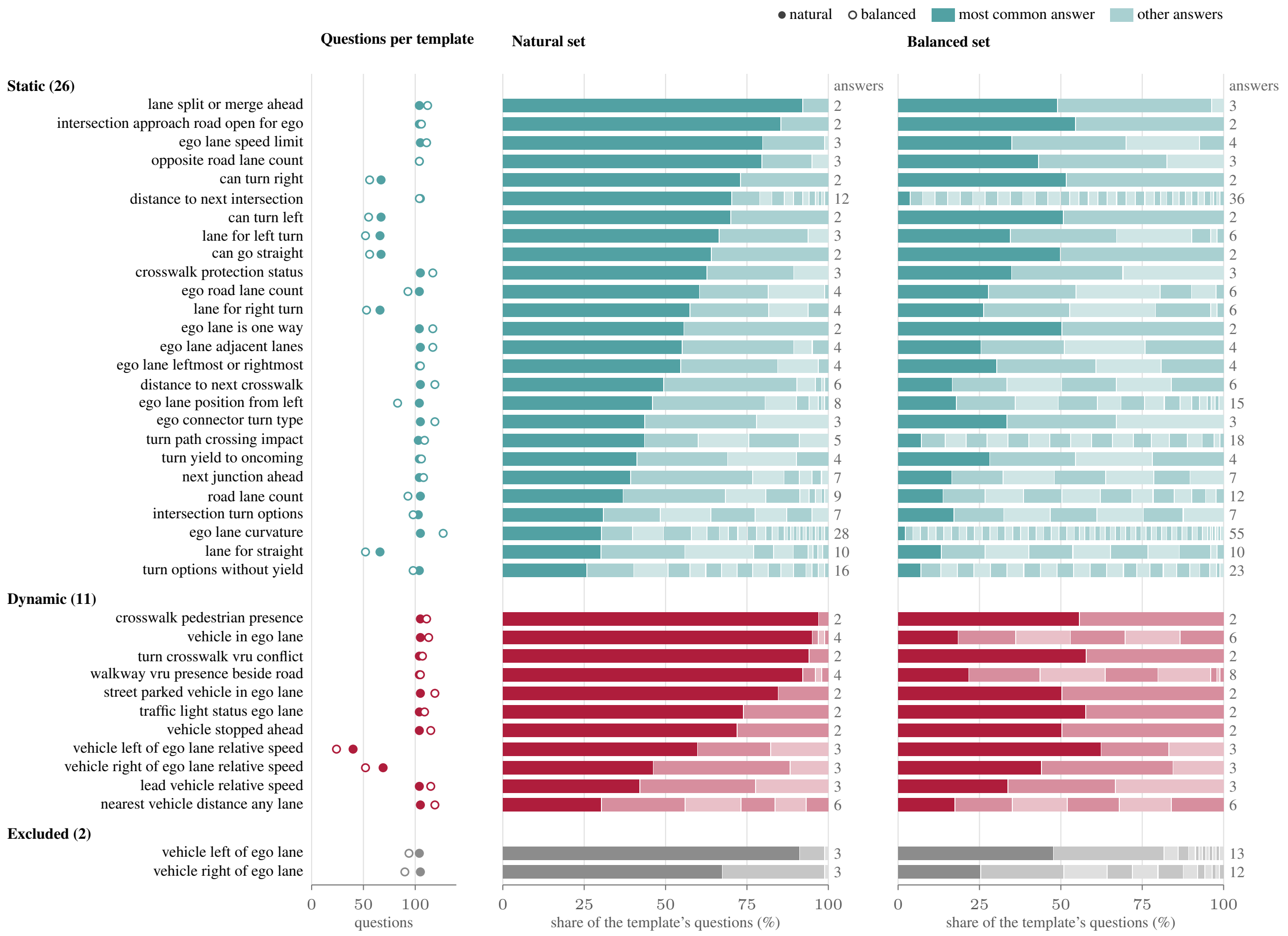

\centering
\def\qssplit{geo}\providecommand{\questionsetsdir}{supplementary/figures/question_stats}%
\colorlet{qsstatic}{_teal}%
\colorlet{qsdynamic}{_red}%
\colorlet{qsexcluded}{black!45}%
\providecommand{\qsfont}{\fontsize{6}{7}\selectfont}%
\pgfplotsset{
    qs axis/.style={scale only axis, y=7.2pt, y dir=reverse, ymin=-0.7, ymax=\qsymax, clip=false,
        axis y line=none, axis x line*=bottom, axis line style={draw=black!55, line width=0.4pt},
        tick align=outside, major tick length=2pt, xtick style={draw=black!55, line width=0.4pt},
        ticklabel style={font=\qsfont, text=black!65}, xlabel style={font=\qsfont, text=black!65, yshift=4pt},
        xmajorgrids, major grid style={draw=black!12, line width=0.4pt},
        title style={font=\qsfont\bfseries, at={(0,1)}, anchor=south west, yshift=1pt},
        unbounded coords=jump, every axis plot/.append style={forget plot}},
    qs rows/.style={restrict expr to domain={\thisrow{heading}}{#1:#1}, unbounded coords=discard},
    qs text/.style={draw=none, mark=none, nodes near coords, point meta=explicit symbolic, qs rows=0,
        every node near coord/.append style={font=\qsfont, inner sep=0pt, #1}},
    qs dot/.style={only marks, mark=*, mark size=1.45pt, mark options={line width=0.6pt}},
}%
\tikzset{
    qs seg 0/.style={fill=qsstatic},   qs seg 1/.style={fill=qsstatic!50},   qs seg 2/.style={fill=qsstatic!28},
    qs seg 3/.style={fill=qsdynamic},  qs seg 4/.style={fill=qsdynamic!50},  qs seg 5/.style={fill=qsdynamic!28},
    qs seg 6/.style={fill=qsexcluded}, qs seg 7/.style={fill=qsexcluded!50}, qs seg 8/.style={fill=qsexcluded!28},
}%
\def\qsseg#1#2#3#4#5{\fill[qs seg #1] (axis cs:#2,#4) rectangle (axis cs:#3,#5);}%
\pgfplotstableread[col sep=comma]{\questionsetsdir/\qssplit_templates.csv}\qstemplates%
\pgfplotstablegetrowsof{\qstemplates}\pgfmathtruncatemacro\qslast{\pgfplotsretval-1}%
\pgfplotstablegetelem{\qslast}{y}\of\qstemplates\pgfmathsetmacro\qsymax{\pgfplotsretval+0.7}%
\begin{tikzpicture}[font=\qsfont]
\begin{axis}[qs axis, name=count, width=56pt, xmin=0, enlarge x limits={upper, value=0.1}, xlabel={questions},
             max space between ticks=40, try min ticks=3,
             title={Questions per template}]
    \addplot [qs text={anchor=east, xshift=-3pt, text=black}] table [x expr=0, y=y, meta=label] {\qstemplates};
    \addplot [draw=none, mark=none, nodes near coords, point meta=explicit symbolic, qs rows=1,
              every node near coord/.append style={font=\qsfont\bfseries, anchor=west, inner sep=0pt, xshift=-118pt}]
        table [x expr=0, y=y, meta=label] {\qstemplates};
    \pgfplotsinvokeforeach{0,1,2}{
        \addplot [qs dot, color=\ifcase#1 qsstatic\or qsdynamic\else qsexcluded\fi, unbounded coords=discard,
                  restrict expr to domain={\thisrow{group}}{#1:#1}] table [x=n_natural, y=y] {\qstemplates};
        \addplot [qs dot, draw=\ifcase#1 qsstatic\or qsdynamic\else qsexcluded\fi, fill=white, unbounded coords=discard,
                  restrict expr to domain={\thisrow{group}}{#1:#1}] table [x=n_balanced, y=y] {\qstemplates};
    }
\end{axis}
\begin{axis}[qs axis, name=natural, at={(count.south east)}, anchor=south west, xshift=18pt, width=126pt,
             xmin=0, xmax=100, xtick={0,25,...,100}, xlabel={share of the template's questions (\%)},
             title={Natural set}]
    \input{\questionsetsdir/\qssplit_natural}
    \addplot [qs text={anchor=west, xshift=2pt, text=black!60}] table [x expr=100, y=y, meta=answers_natural] {\qstemplates};
    \node [anchor=west, xshift=2pt, inner sep=0pt, text=black!60] at (axis cs:100,0) {answers};
\end{axis}
\begin{axis}[qs axis, name=balanced, at={(natural.south east)}, anchor=south west, xshift=27pt, width=126pt,
             xmin=0, xmax=100, xtick={0,25,...,100}, xlabel={share of the template's questions (\%)},
             title={Balanced set}]
    \input{\questionsetsdir/\qssplit_balanced}
    \addplot [qs text={anchor=west, xshift=2pt, text=black!60}] table [x expr=100, y=y, meta=answers_balanced] {\qstemplates};
    \node [anchor=west, xshift=2pt, inner sep=0pt, text=black!60] at (axis cs:100,0) {answers};
\end{axis}
\node [anchor=south east, inner sep=0pt, yshift=20pt] at (balanced.north east) {%
    \tikz[baseline=-0.6ex]\fill[black!75] (0,0) circle (1.45pt);~natural\quad
    \tikz[baseline=-0.6ex]\draw[black!75, line width=0.6pt, fill=white] (0,0) circle (1.45pt);~balanced\quad
    \tikz[baseline=-0.6ex]\fill[qsstatic] (0,-2.6pt) rectangle (9pt,2.6pt);~most common answer\quad
    \tikz[baseline=-0.6ex]\fill[qsstatic!50] (0,-2.6pt) rectangle (9pt,2.6pt);~other answers};
\end{tikzpicture}%
\caption{KNOW test questions on GEOTEST. Rows: the 39 templates, grouped into static (road layout and rules), dynamic (actors and traffic lights) and excluded, within a group the most skewed natural template first.
Left: questions per template in the natural (filled) and the balanced (hollow) set. Middle and right: each template's answers in the natural and the balanced set, one segment per answer from the most to the least common, the most common in full color. Right of each bar: the number of distinct answers.}
\label{fig:supp_question_sets_navtest}
\end{figure*}

Each KNOW test set asks one question per decision point, each about one fact the graph engine states (one template per question). We build two sets on the same decision points. Of the 40 queries, we exclude the bicycle-ahead query (\cf \Cref{tab:know_queries}, K31) from evaluation, as it has very few positive instances on NAVTEST or GEOTEST, leaving 39 templates; the two adjacent-lane lead-vehicle templates (\Cref{tab:know_queries}, K34–35) are additionally excluded from static and dynamic accuracy, since their answers partly state the lane layout. The \emph{natural} set spreads the 39 templates evenly over the available decision points. In a seeded random order, each point takes the template, among those its graph states, used least so far, while answers follow along as they occur. 
The \emph{balanced} set uses the same decision points as the natural set and takes its per-template question counts as target quotas. Subject to these quotas, the assignment seeks to distribute answers within each template as evenly as availability permits. Because each decision point supports only a subset of template--answer pairs and contributes exactly one question,
the greedy assignment does not enforce the target quotas exactly. \Cref{fig:supp_question_sets_navtest} shows both sets on GEOTEST.

When scoring models on these sets, accuracy is computed separately for each template and then
averaged with equal weight across templates. Consequently,
differences in template counts do not change the templates'
weights in the reported score, although the selected questions
within each template can differ. For multiple-choice scoring, the options are the reference answer and up to three distractors sampled from other answers among GEOTRAIN's training questions, in random order (two to four options per question, so chance averages 32.5\%). The blind prior in \Cref{tab:vqa_investigation} sees neither images nor ego motion: among a question's options, it picks the answer that training questions of the same template and with the same navigation instruction give most often (template only, where the instruction never occurs in training), and is therefore right whenever the reference is the common answer, as it often is on the natural set and rarely on the balanced one. Without the instruction, it reaches 49.8\% on the natural and 31.0\% on the balanced set. 

\section{Details on partitioning}
\label{sec:partitioning_details}

Our geographic split assigns complete intersection-centered road regions to the held-out test set and removes a surrounding buffer from training. All repeated traversals of a location receive the same assignment, irrespective of the original log split. We construct the regions separately in each of the four nuPlan cities, using recorded ego positions and HD-map intersection annotations. The geographic assignment is fixed before selecting the NAVSIM \cite{navsim} decision points used in our experiments.

\paragraph{Source trajectories and intersection anchors}
The geographic index is built from 1,457 camera-supported nuPlan logs. For each log, we sort its ego poses by timestamp and subsample them to 5\,Hz. Candidate anchors are map intersections with control type \texttt{default}, \texttt{traffic light}, or \texttt{stop sign}. An intersection is eligible when at least one log has a run of three or more consecutive sampled poses within 30\,m of its polygon, with at least ten preceding and more than thirty following poses in that log (approximately 2\,s and 6\,s). The source cells of an eligible intersection are all occupied cells containing a pose within its 30\,m approach area. We merge anchors transitively whenever these source-cell sets overlap, so nearby junctions with shared approaches form one anchor group.

\paragraph{Road catchments}
We then discretize each city's trajectories into axis-aligned $10\times10$\,m cells, retaining only cells occupied by a sampled ego pose. An undirected graph connects two distinct cells when consecutive poses from the same log move between them, provided the time difference is positive and at most 1\,s and the displacement is at most 30\,m. An edge is weighted by the shortest observed pose-to-pose displacement for that cell pair. Thus proximity alone does not connect neighboring cells: connectivity must have been observed in a recorded drive.
A multi-source Dijkstra search assigns each occupied cell to its nearest anchor group on this graph, with all source cells initialized at zero distance. Equal-distance ties are resolved by the groups' sorted map intersection identifiers. The resulting \emph{catchment} comprises an anchor group and the observed road cells assigned to it. Notably, a catchment can contain several intersections and is selected as a whole.

\paragraph{Buffers and region selection}
For each catchment $C_i$, we define its protected footprint as
\begin{equation}
 P_i=\{c:\min_{c'\in C_i}d_{\mathrm{cell}}(c,c')\leq100\,\mathrm{m}\},
\end{equation}
where $c$ ranges over occupied cells and $d_{\mathrm{cell}}$ is the minimum Euclidean distance between the two cell squares. Using square-to-square rather than center-to-center distance conservatively protects every cell that could contain a point within 100\,m of a test cell. Two catchments cannot both be selected if any pair of their cells is within 200\,m under the same distance rule.
We select test catchments independently per city. If $z_i$ indicates selection, the coverage and region-count constraints are
\begin{equation}
\begin{aligned}
 \lceil0.15N\rceil &\leq \sum_i |C_i|z_i\leq\lfloor0.20N\rfloor,\\
 \sum_i z_i&\geq2,
\end{aligned}
\end{equation}
where $N$ is the city's number of occupied cells. We also require at least one selected catchment containing a traffic-light intersection and at least one containing a stop-sign intersection whenever these types occur among the eligible catchments.

Among feasible selections, we minimize, in strict priority order: recorded driving distance in the protected footprints (test plus buffer), their pose count, driving distance in the test catchments, protected-cell count, and selected-catchment count. Each objective is solved to optimality and fixed before optimizing the next. The procedure uses neither model predictions nor evaluation scores.

\begin{table*}[t]
\centering
\caption{Frozen geographic partition before NAVSIM decision-point filtering. Test-cell share is relative to all occupied $10\times10$\,m cells in the city. Distances sum recorded travel across repeated traversals. Clearance is the minimum Euclidean distance between any sampled training and test ego positions in the same city.}
\label{tab:geographic_partition_statistics}
\small
\begin{tabular*}{\textwidth}{@{\extracolsep{\fill}}lrrrrrrrr@{}}
\toprule
City & Logs & Test & Test & Test cells & Train & Test & Buffer & Clearance \\
 & & regions & junctions & (\%) & (km) & (km) & (km) & (m) \\
\midrule
Singapore & 110 & 3 & 29 & 15.1 & 86.0 & 12.1 & 14.9 & 102.3 \\
Boston & 317 & 5 & 29 & 15.5 & 227.6 & 18.2 & 19.5 & 106.9 \\
Las Vegas & 868 & 6 & 46 & 15.1 & 1,082.1 & 8.9 & 4.7 & 103.5 \\
Pittsburgh & 162 & 4 & 9 & 15.7 & 140.9 & 21.8 & 27.0 & 104.1 \\
\midrule
Total & 1,457 & 18 & 113 & --- & 1,536.6 & 61.0 & 66.1 & --- \\
\bottomrule
\end{tabular*}
\end{table*}

\paragraph{Assignments and decision-point selection}
Every sampled pose in a selected catchment is assigned to test. Poses in the union of protected footprints but outside the test catchments are discarded as buffer; all remaining poses are assigned to training. We split each log into contiguous intervals of one assignment. A single log can therefore contribute to both training and test, as long as a suitably large buffer lies between them. 

For the geographic experiments, we pool decision points from NAVSIM's NAVTRAIN and NAVTEST filters, then apply the geographic intervals. The RoadVLA exporter accepts a sample only when its resolved image and ego-motion history and its future trajectory target fit within a single interval of the requested side. With the training settings used here, this covers 1.5\,s of input history and 5\,s of future waypoints. Buffer intervals contribute no samples.

\paragraph{Realized split and verification}
\Cref{tab:geographic_partition_statistics} summarizes the frozen geographic index before NAVSIM filtering and sample-window checks. The 18 selected catchments contain 113 map intersection anchors and cover 15.1--15.7\% of occupied cells in each city. We verify that every pose has exactly one assignment, that the per-city cell budget and selection constraints hold, and that the closest training--test pose pair is strictly more than 100\,m apart in every city. 

\paragraph{Comparison with NAVTEST}
NAVSIM's original split separates training and test samples while allowing repeated visits to the same roads across the two sets. Our GEO split instead pools its NAVTRAIN and NAVTEST decision points and reassigns them by location. Thus the geographic test set, GEOTEST, is not a subset of NAVTEST: it contains decision points from either original partition whose locations fall in the selected regions. The reference models train on NAVTRAIN and are evaluated on NAVTEST; their geographically held-out counterparts train on the geographic training set and are evaluated on the geographic test set.

\Cref{tab:partition_navtest_comparison} compares the sets: The training-set sizes are similar (103k versus 106k decision points), while GEOTEST is smaller than NAVTEST (3.7k versus 12.1k), and differs in city composition: Singapore and Pittsburgh account for 69.4\% of GEOTEST against 35.8\% of NAVTEST, and Las Vegas for 7.4\% against 33.4\%.

\paragraph{Why the test sets cannot be matched}
This difference is not a free design choice; it follows from geographic disjointness on a dataset that drives the same roads repeatedly. NAVTEST is drawn from densely driven roads: the place (10\,m cell and heading bin) of its median decision point is driven by 9 distinct NAVSIM drives, 30 in Las Vegas. A geographically disjoint test set that matches NAVTEST therefore does not exist with a usable training set. Conversely to \Cref{tab:train_test_overlap}, which measures how close each test decision point lies to training, we ask how close each training decision point lies to NAVTEST: 95.1\% of NAVTRAIN's points lie within 100\,m of a NAVTEST point (85.8--99.0\% per city), so holding out NAVTEST's own locations with our 100\,m buffer would leave about 5\% of the training data. Any held-out set that keeps most of the training data must therefore come from less-driven roads, which is what our selection does by minimizing the driving it removes. This hits Las Vegas hardest, since its densely driven core cannot be held out without removing most of its training data.

\paragraph{Controlling for the difference}
We address the remaining mismatch in four ways. First, we report the performance gap between the test sets (\cf \Cref{sec:geo_drop_main}) under three weightings: raw, per city averaged equally, and per city reweighted to NAVTEST's city mix. Second, the human agent, scored on both test sets, drops by only 0.4 points on straight driving and 1.0 on turns, so the scorer does not judge the held-out scenes more harshly. Third, the additional failures on GEOTEST are geometric (DAC), while collisions, traffic lights and USE's rule terms fail at comparable rates on both sets (\Cref{fig:failures}); a shift toward generally harder scenes would raise those as well. Fourth, our central findings do not rest on comparing the two test sets. The dose-response analysis (\Cref{fig:dose_response}) compares NAVTEST decision points with each other, within each city, by how often NAVTRAIN drove their place. Restricting NAVTEST to familiar places widens the retraining gap monotonically (Appendix \cref{sec:geo_drop_addition}). The place-scaling experiment (\Cref{fig:place_scaling}) varies only the training set at a fixed test set and finds static KNOW accuracy rising with repeated drives only at the driven places. These comparisons do not match every road configuration or traffic situation between the two test sets, but each points the same way.

\begin{table}[t]
\centering
\caption{Comparison of NAVSIM's original partition (train refers to NAVTRAIN) and our GEO split.}
\label{tab:partition_navtest_comparison}
\small
\begin{tabular*}{\columnwidth}{@{\extracolsep{\fill}}lrr@{}}
\toprule
 & NAVSIM & GEO split \\
\midrule
Training decision points & 103.3k & 106.3k \\
Test decision points & 12.1k & 3.7k \\
Test logs & 136 & 162 \\
Median distance to train (m) & 0.4 & 185.5 \\
\midrule
\multicolumn{3}{@{}l}{\emph{Share of test decision points (\%)}} \\
Singapore & 15.8 & 38.0 \\
Boston & 30.7 & 23.2 \\
Las Vegas & 33.4 & 7.4 \\
Pittsburgh & 20.1 & 31.4 \\
\bottomrule
\end{tabular*}
\end{table}

\begin{figure*}[h!]
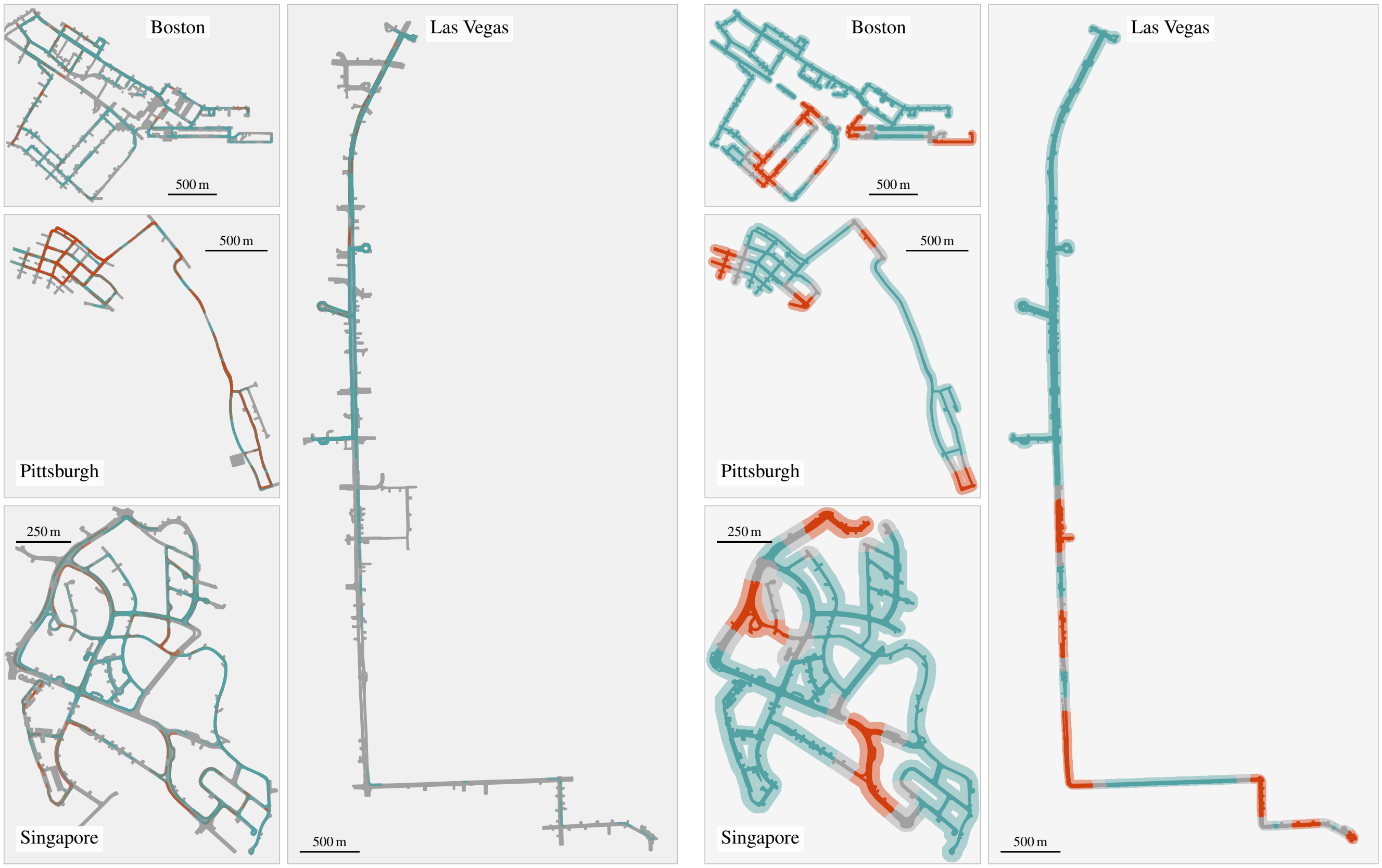

    \centering
    \begin{minipage}[t]{0.49\textwidth}
        \centering
        \def\citiesprefix{leaky}
        \input{supplementary/figures/city_splits}
    \end{minipage}%
    \hfill
    \begin{minipage}[t]{0.49\textwidth}
        \centering
        \def\citiesprefix{geo}
        \input{supplementary/figures/city_splits}
    \end{minipage}
    \caption{Left: NAVSIM's \textcolor{_teal}{\textbf{training}} and \textcolor{_orange}{\textbf{test}} drives share the same location. Right: Our GEO split repartitioning cleanly separates \textcolor{_teal}{\textbf{training}} from \textcolor{_orange}{\textbf{test}} zones, with a \textcolor{gray}{\textbf{buffer}} in between.}
    \label{fig:leaky-splits}
\end{figure*}
\section{Implementation details of RoadVLA}
\label{sec:training_details}
RoadVLA-0, RoadVLA-SAY and RoadVLA-KNOW finetune Qwen3-VL-4B-Instruct~\cite{Qwen3-VL} with one shared recipe and differ only in their target text. Each variant is trained twice from the same base weights. The first run uses the GEOTRAIN of our geographically held-out split (106k decision points) and is evaluated on GEOTEST. The second uses NAVTRAIN (103k decision points) and is evaluated on NAVTEST as the exposed reference. Both training sets are exported with the same code and prompt format.

\paragraph{Inputs}
Each sample contains 12 images from the front-left, front and front-right cameras at $T={-}1.5$, ${-}1.0$, ${-}0.5$ and $0$\,s. Images are resized from $1920\times1080$ to $416\times224$ pixels. The text prompt (\Cref{fig:roadvla_prompt}) adds the ego motion over the preceding 1.5\,s at 10\,Hz and a navigation instruction, following the configuration of \cite{qwendrive}. The ego motion is given as 16 states of position, yaw and speed in the ego frame at $T=0$, with positions in integer decimeters, yaw in radians relative to the current heading, and speed in m/s. The navigation instruction is \emph{Go straight} (67\% of the GEOTRAIN), \emph{Turn right} (17\%), \emph{Turn left} (16\%) or \emph{Make a u-turn} (${<}0.1\%$).

\paragraph{Navigation instruction}
Our instruction names the movement ego makes at the next intersection along its recorded drive, read from the lane connector it traverses there: \emph{Turn left}, \emph{Turn right}, \emph{Go straight} or \emph{Make a u-turn}. The intersection is searched up to 200\,m ahead. Inside an intersection, the instruction names the current movement until ego is leaving it, and without an intersection within 200\,m it is \emph{Go straight}. 

We derive our own instruction for two reasons. First, it is bound to the road layout: it names ego's movement at the next intersection, read from the lane connector ego traverses there, so it refers to the same intersection and movements that the KNOW queries (\Cref{tab:know_queries}) and SAY clauses (\Cref{sec:clauses_templates_say}) describe. NAVSIM's native driving command (left, straight, right or unknown) instead follows the direction of the route ahead without reference to an intersection, and also covers lane changes and sharp curves~\cite{navsim}. Second, an instruction bound to intersections must not reveal where they are, since several KNOW queries about intersection proximity and SAY's intersection-ahead clause would otherwise leak their answers into the prompt. Our navigation instruction is derived from an along-path look-up of 200\,m and is present at every decision point, so it reveals at most that an intersection lies within 200\,m, not how near it is. 

We verify that our choice does not change the action-space results. We retrained RoadVLA-0 on both training sets with NAVSIM's command in place of ours, worded the same way and otherwise unchanged, and evaluated it with NAVSIM's command. On NAVTEST, both models perform the same: NAVSIM's command minus ours gives $-0.1$ PDMS [$-0.5$, $+0.4$] and $+0.002$\,m ADE [$-0.006$, $+0.011$], with paired 95\% bootstrap intervals over logs. 

\paragraph{Targets}
All variants predict the next 5\,s as 50 waypoints at 10\,Hz, written after the prefix \texttt{TRAJECTORY:} as space-separated pairs \texttt{x,y}. Coordinates are integers in decimeters, in the ego frame with $x$ forward and $y$ left. RoadVLA-SAY first writes the SAY reasoning of the decision point (see Appendix \cref{sec:clauses_templates_say}) after \texttt{REASONING:}, then the trajectory. RoadVLA-KNOW is trained on two-turn conversations. The first turn is the RoadVLA-0 sample unchanged, and the second asks one KNOW question about the same decision point, answered after \texttt{ANSWER:}, so that the trajectory prediction does not get any privileged information from the KNOW question.
Each decision point is assigned its KNOW question once, before training, and keeps it in every epoch. At test time, the second turn follows the model's own predicted trajectory. The loss is the next-token cross-entropy on the assistant turns only, so it covers both turns for RoadVLA-KNOW. Prompt and image tokens are masked.

\begin{figure*}[t]
\newcommand{\rvph}[1]{{\normalfont\color{black!55}$\langle$#1$\rangle$}}
\newcommand{\rvline}[1]{\par\hangindent=1.5em\hangafter=1\noindent#1\par}
\newcommand{\rvgap}{\par\vspace{3pt}}
\newcommand{\rvrole}[1]{\par\vspace{2pt}\noindent{\normalfont\sffamily\bfseries\scriptsize\color{black!70}#1}\par\vspace{1pt}}
\newcommand{\rvimg}{\textless{}image\textgreater{}}
\begin{tcolorbox}[colback=black!2, colframe=black!45, boxrule=0.5pt, arc=2pt, left=5pt, right=5pt, top=3pt, bottom=4pt,
    fonttitle=\small\bfseries, coltitle=white, colbacktitle=black!55, title={RoadVLA prompt and targets}]
\footnotesize\ttfamily\raggedright
\rvrole{User (all variants)}
\rvline{T-1.5s: \rvimg\rvimg\rvimg}
\rvline{T-1.0s: \rvimg\rvimg\rvimg}
\rvline{T-0.5s: \rvimg\rvimg\rvimg}
\rvline{T=0: \rvimg\rvimg\rvimg}
\rvgap
\rvline{You are given images for the past 4 frames (1.5s at 2 Hz). 3 images per frame are ordered LEFT-CENTER-RIGHT.}
\rvline{Positions are integer tenths of a meter in ego frame at T=0: x forward, y left, (0,0) is ego at T=0.}
\rvline{Past ego-motion for the 1.5s before T=0 at 10 Hz (step: index from T=0, x/y: tenths of a meter, yaw: radians relative to T=0 heading, speed: m/s):}
\rvline{step,x,y,yaw,speed;-15,\rvph{$x$},\rvph{$y$},\rvph{yaw},\rvph{speed};-14,\ldots;-1,\ldots;0,0,0,0,\rvph{speed}}
\rvgap
\rvline{Nav Instr: \rvph{Go straight. $|$ Turn left. $|$ Turn right. $|$ Make a u-turn.}}
\par\vspace{4pt}{\color{black!30}\hrule height 0.4pt}\vspace{2pt}
\noindent\begin{minipage}[t]{0.31\linewidth}\raggedright
\rvrole{RoadVLA-0 and KNOW turn 1 (user, cont.)}
\rvline{Task: predict trajectory for the next 5.0s as 50 waypoints at 0.1s in tenths of a meter. Format:}
\rvline{TRAJECTORY: x,y x,y ...}
\rvrole{Assistant}
\rvline{TRAJECTORY: \rvph{$x_1$},\rvph{$y_1$} \rvph{$x_2$},\rvph{$y_2$} \ldots{} \rvph{$x_{50}$},\rvph{$y_{50}$}}
\end{minipage}\hfill
\begin{minipage}[t]{0.31\linewidth}\raggedright
\rvrole{RoadVLA-SAY (user, cont.)}
\rvline{Task: first reason about the scene, then predict trajectory for the next 5.0s as 50 waypoints at 0.1s in tenths of a meter. Format:}
\rvline{REASONING: \textless{}your reasoning\textgreater{}}
\rvline{TRAJECTORY: x,y x,y ...}
\rvrole{Assistant}
\rvline{REASONING: There is a lead vehicle in my lane, 10-20 m ahead, moving faster than me. The road runs straight through this intersection. \ldots}
\rvline{TRAJECTORY: \rvph{$x_1$},\rvph{$y_1$} \ldots{} \rvph{$x_{50}$},\rvph{$y_{50}$}}
\end{minipage}\hfill
\begin{minipage}[t]{0.31\linewidth}\raggedright
\rvrole{RoadVLA-KNOW turn 2 (user)}
\rvline{Question: Is there a lead vehicle ahead of ego in its own lane, and how far away is it?}
\rvgap
\rvline{Task: answer the question. Format:}
\rvline{ANSWER: \textless{}your answer\textgreater{}}
\rvrole{Assistant}
\rvline{ANSWER: yes, 10-20 m ahead}
\end{minipage}
\end{tcolorbox}
\caption{Prompt and target text of the RoadVLA variants. The user turn above the line is shared by all variants; each variant appends its own task line. The reasoning, question and answer shown are from one decision point of GEOTRAIN, with the reasoning shortened.}
\label{fig:roadvla_prompt}
\end{figure*}

\begin{table}[t]
\centering
\caption{Shared finetuning recipe of the RoadVLA variants.}
\label{tab:roadvla_hparams}
\small
\begin{tabular}{@{}ll@{}}
\toprule
Parameter & Value \\
\midrule
Base model & Qwen3-VL-4B-Instruct \cite{Qwen3-VL} \\
Trained & LLM, vision--language merger \\
Frozen & vision encoder \\
Image size & $416\times224$ pixels \\
Optimizer & AdamW, $\beta_1{=}0.9$, $\beta_2{=}0.999$, $\epsilon{=}10^{-8}$ \\
Weight decay & 0.01 \\
Learning rate & $6\times10^{-5}$ \\
Schedule & 3\% linear warm-up, cosine to 0 \\
Global batch & 64 (8 per GPU, 8 GPUs) \\
Epochs & 5 \\
Steps & 8,295 geographic, 8,055 NAVTRAIN \\
Gradient clipping & global norm 1.0 \\
Precision & bf16, fp32 optimizer states \\
Max.\ sequence length & 16,384 tokens \\
Seed & 2025 \\
Evaluated checkpoint & final \\
\bottomrule
\end{tabular}
\end{table}

\paragraph{Optimization}
\Cref{tab:roadvla_hparams} lists the settings. We train the language model and the vision--language merger and keep the vision encoder frozen. Training runs for 5 epochs with a global batch of 64 (8 per GPU on 8 GPUs, without gradient accumulation). This gives around 8k steps on both GEOTRAIN and NAVTRAIN.

\paragraph{Inference}
We decode greedily with vLLM (temperature 0, at most 2,048 new tokens) at the training image resolution. For PDMS and EPDMS, we convert the first 4\,s of the predicted waypoints to NAVSIM's eight poses at 0.5\,s intervals. Each pose's heading is the direction of the predicted path over $\pm1$\,s around it. Where the path moves slower than 1\,m/s over that window, the last heading is kept, since the direction of a nearly stationary path is dominated by rounding.

\paragraph{Compute}
Each run trains on a single node with eight NVIDIA A100 GPUs (40\,GB). We use bf16 mixed precision with fp32 master weights and optimizer states, sharded with DeepSpeed ZeRO stage 2. Gradient checkpointing and a fused linear cross-entropy kernel (Liger) keep the batch within memory. A five-epoch run takes 9.5 to 10\,h.

\section{Robustness of training exposure study}
\label{sec:dose_response_details}

The findings in \Cref{fig:dose_response} rest on our  choices: how a NAVTEST decision point's neighborhood is drawn according to radius and heading agreement (training samples within 5\,m of its ego pose and within $\pm$45$^\circ$ of its heading), how the number of training drives through that neighborhood is binned (0, 1--3, 4--7 and 8 or more), and how a turn is told from straight driving (a net heading change of at least 15$^\circ$ over the recorded 4\,s future, measured between the first and the
last quarter of the path). Decision points whose recorded path is shorter than 5\,m (8.7\% of NAVTEST) count as neither turn nor straight driving. We rerun the analysis with one choice changed at a time and everything else at its default: the same 13 baselines trained on NAVSIM's training logs (85k decision points), and the same comparison inside each city against the lowest bin.

\paragraph{Which cities enter}
A model's gain in a bin is its mean PDMS there minus its mean in the lowest bin, taken inside each city and averaged over the cities, so that no gain compares one city with another. A city enters only if it has at least ten NAVTEST decision points in every bin, for turns and for straight driving alike; the same cities then enter every bin and both curves. \Cref{tab:supp_dose_cities} shows why only Boston and Pittsburgh qualify under the defaults. Las Vegas has no never-seen turn at all: NAVSIM's training logs drive nearly all of its NAVTEST neighborhoods, so it offers no reference to compare against. Singapore is the opposite: none of its neighborhoods is driven eight times or more, so it has no top bin. Comparing the remaining samples across cities would confound familiarity with the city itself, since Las Vegas is always familiar and Singapore never heavily so. 

\begin{table}[t]
\centering
\caption{NAVTEST samples per city, maneuver and exposure bin under the default choices in \Cref{sec:dose_response}; the decision points whose recorded path is shorter than 5\,m count as neither turn nor straight driving (8.7\% of NAVTEST). A city enters only with at least ten samples in every cell; the empty cells (bold) leave out Las Vegas and Singapore.}
\label{tab:supp_dose_cities}
\small
\begin{tabular}{@{}ll cccc@{}}
\toprule
 & & \multicolumn{4}{c}{Train drives at test neighborhood} \\
\cmidrule(l){3-6}
City & Group & 0 & 1--3 & 4--7 & 8+ \\
\midrule
\multicolumn{6}{l}{\textit{included cities}} \\
\qquad Boston & turn & 116 & 182 & 195 & 208 \\
 & straight & 341 & 954 & 746 & 556 \\
\qquad Pittsburgh & turn & 64 & 271 & 170 & 271 \\
 & straight & 197 & 536 & 504 & 281 \\
\midrule
\multicolumn{6}{l}{\textit{excluded cities}} \\
\qquad Las Vegas & turn & \textbf{0} & 1 & 3 & 1{,}091 \\
 & straight & 30 & 107 & 276 & 2{,}102 \\
\qquad Singapore & turn & 214 & 602 & 53 & \textbf{0} \\
 & straight & 402 & 584 & 36 & \textbf{0} \\
\bottomrule
\end{tabular}
\end{table}

\paragraph{Results}
\Cref{tab:supp_dose_robustness} reports, for every setting, the cities that enter, the turning samples in the lowest and the top bin, the baselines' median gain on turns in each bin above the lowest, how many of them gain in the top bin and the smallest such gain, the human agent's top-bin gain, and the baselines' median top-bin gain on straight driving. This shows that the findings of \Cref{fig:dose_response} do not depend on any of the choices. In every setting all 13 baselines
score higher on turns at the most-driven neighborhoods than in the lowest bin, and the median gain grows with the number of drives wherever there are more than two bins.
The human agent, scored the same way, gains at most 2.9 in any setting but the 20\,m radius (4.2), and on straight driving the median baseline gains at most 2.5, except with a 1\,m radius (5.2) and turn thresholds of 45$^\circ$ and 60$^\circ$ (3.5 and 3.7), where the straight group also holds bends of up to that angle. Finer bins at the top run out of samples: with a top bin of 16 or more drives, Boston has no turning sample there and Pittsburgh eight, so no city qualifies and the setting cannot be evaluated. Finally, counting the training samples' supervised 4\,s future waypoints as well as their decision points when drawing a neighborhood (765{,}981 instead of 85{,}109 points) leaves the finding intact: on turns, the median baseline gains 4.1, 11.0 and 10.8 points in the three bins, all 13 baselines gain in the top bin (smallest 6.0).

\begin{table*}[t]
\centering
\caption{The gap between high- and low-exposure neighborhoods on NAVTEST under other parameter choices, one changed at a time. Gains are in
PDMS points against the lowest bin, inside
each city and averaged over the cities that meet the sample rule (\emph{Cities}: B Boston, P Pittsburgh, S Singapore).
\emph{Gain $>$0}, \emph{Smallest}: how many baselines
gain in the top bin, and the smallest top-bin gain. \emph{Human}: the NAVSIM human agent's top-bin gain.
\emph{Straight}: the baselines' median top-bin gain on straight driving.}
\label{tab:supp_dose_robustness}
\small
\setlength{\tabcolsep}{4pt}
\begin{tabular}{@{}ll c c c c c c c@{}}
\toprule
 & & & \multicolumn{5}{c}{Turns} & Straight \\
\cmidrule(lr){4-8}\cmidrule(l){9-9}
Parameter & Setting & Cities & Samples (lowest / top) & Median gain by bin & Gain $>$0  & Smallest & Human & Median \\
\midrule
\multicolumn{2}{@{}l}{Defaults (5\,m, $\pm$45$^\circ$, 0,1--3,4--7,8+, 15$^\circ$)} & BP & 180 / 479 & 4.4 / 10.3 / 13.0 & 13/13 & 7.5 & 1.8 & 1.2 \\
\midrule
Bins & 0,1--2,3--5,6+ & BP & 180 / 621 & 1.6 / 9.3 / 12.2 & 13/13 & 7.6 & 1.8 & 0.2 \\
 & 0,1--4,5--9,10+ & BP & 180 / 337 & 4.8 / 11.6 / 13.3 & 13/13 & 8.1 & 2.1 & 0.9 \\
 & 0,1--5,6+ & BP & 180 / 621 & 5.7 / 12.2 & 13/13 & 7.6 & 1.8 & 0.2 \\
 & 0,1+ & BPS & 394 / 1{,}952 & 8.6 & 13/13 & 3.5 & 1.5 & 1.0 \\
 & 0--3,4--7,8+ & BP & 633 / 479 & 7.2 / 9.6 & 13/13 & 5.8 & 0.3 & 1.0 \\
 & 1--3,4--7,8+ & BP & 453 / 479 & 6.4 / 8.0 & 13/13 & 4.4 & $-$0.3 & 0.9 \\
 & 0--3,4+ & BPS & 1{,}449 / 897 & 4.1 & 13/13 & 2.8 & 0.8 & 1.2 \\
 & 0,1--7,8--15,16+ & \multicolumn{7}{l}{\emph{no city meets the sample rule}} \\
\midrule
Radius & 1\,m & BP & 439 / 151 & 5.6 / 10.5 / 11.6 & 13/13 & 6.5 & 0.2 & 5.2 \\
 & 2.5\,m & BP & 242 / 350 & 3.6 / 10.0 / 11.2 & 13/13 & 6.5 & 0.7 & 1.0 \\
 & 7.5\,m & BP & 160 / 541 & 3.5 / 9.2 / 10.7 & 13/13 & 7.3 & 2.2 & 0.8 \\
 & 10\,m & BP & 145 / 567 & 2.4 / 9.1 / 10.3 & 13/13 & 7.3 & 2.9 & 0.8 \\
 & 20\,m & BP & 104 / 651 & 7.4 / 12.9 / 15.7 & 13/13 & 1.4 & 4.2 & 1.1 \\
\midrule
Heading cone & $\pm$10$^\circ$ & BP & 227 / 406 & 3.3 / 9.9 / 12.1 & 13/13 & 7.2 & 0.8 & 1.4 \\
 & $\pm$22.5$^\circ$ & BP & 196 / 462 & 3.1 / 9.4 / 11.9 & 13/13 & 6.3 & 1.5 & 1.3 \\
 & $\pm$90$^\circ$ & BP & 172 / 484 & 4.0 / 10.4 / 12.8 & 13/13 & 7.4 & 1.9 & 1.2 \\
 & none & BP & 95 / 616 & 3.6 / 8.2 / 11.0 & 13/13 & 8.6 & 1.5 & 0.7 \\
\midrule
Turn & threshold 10$^\circ$ & BP & 230 / 597 & 3.1 / 8.1 / 11.0 & 13/13 & 6.5 & 1.4 & 1.3 \\
 & threshold 20$^\circ$ & BP & 157 / 431 & 4.8 / 11.3 / 13.7 & 13/13 & 7.4 & 1.7 & 1.6 \\
 & threshold 30$^\circ$ & BP & 125 / 369 & 5.6 / 9.9 / 12.9 & 13/13 & 6.6 & 1.6 & 2.5 \\
 & threshold 45$^\circ$ & BP & 87 / 282 & 3.4 / 7.2 / 11.7 & 13/13 & 4.2 & 1.9 & 3.5 \\
 & threshold 60$^\circ$ & BP & 58 / 210 & 4.3 / 8.9 / 15.2 & 13/13 & 6.2 & 1.6 & 3.7 \\
 & chords over 15\% & BP & 181 / 478 & 5.0 / 10.9 / 12.4 & 13/13 & 7.3 & 1.7 & 1.2 \\
 & chords over 35\% & BP & 183 / 480 & 4.5 / 10.7 / 12.9 & 13/13 & 7.8 & 1.8 & 1.2 \\
\bottomrule
\end{tabular}
\end{table*}

\section{Additional results for geographic retraining}
\label{sec:geo_drop_addition}

The retraining table, \Cref{tab:retrain_leakage} in \Cref{sec:geo_drop_main}, compares NAVTRAIN-trained models on NAVTEST with their GEOTRAIN-trained counterparts on GEOTEST. Since the external baselines \cite{chitta2023transfuser, kirby2026drivor, li2026recogdrive} train on the train logs of NAVTRAIN only (85k), we follow their original recipes and match their GEOTRAIN counterparts with an equally large subset of GEOTRAIN. RoadVLA trains on all of NAVTRAIN (103k) and all of GEOTRAIN (106k). 

\paragraph{Stratifying by training exposure} NAVTEST also contains places that the exposed models' training split drove rarely or never (as discussed in \Cref{sec:dose_response} and Appendix \cref{sec:dose_response_details}, the zero-exposure bin).  In this ablation, we therefore restrict the exposed side to NAVTEST frames with a training exposure of at least two, four or six drives, with exposure defined as in the dose response (\Cref{sec:dose_response}), \ie, the number of distinct training drives with a decision point within 5\,m of the frame's ego pose and within $\pm$45$^\circ$ of its heading. Drives are counted over each model's training data: NAVTRAIN's training logs for the external baselines, all of NAVTRAIN for RoadVLA and the human agent. The held-out side, the scores and the three city weightings stay as in the main table; all four cities have at least ten scored samples per city on both sides at every threshold, except Singapore for the external baselines at six drives or more, where the city weightings use the other three.
\Cref{tab:supp_retrain_familiar} reports the mean drop over the three weightings. The higher the training exposure, the larger the drop on turns, for every model: from 8.2--11.1 points of EPDMS+USE on all of NAVTEST to 10.7--16.5 at an exposure of six drives or more. On straight driving it grows less, from 2.0--5.2 to 3.7--6.7. The human agent's drop stays small, so familiar roads are not simply easier to drive. The gap between exposed and held-out models is largest where the exposed models had seen the road most often.

\begin{table}[t]
\centering
\caption{Drop from the models trained on seen roads on NAVTEST to the models trained on unseen roads and evaluated on GEOTEST (differences in EPDMS+USE; human agent: differences in PDMS). The exposed side is restricted to NAVTEST frames with a training exposure (distinct training drives within 5\,m and $\pm$45$^\circ$, as in \Cref{fig:dose_response}) of at least two, four or six drives (\emph{All}: every frame, as in \Cref{tab:retrain_leakage}). Mean over raw, equal-city and NAVTEST-city weighting.}
\label{tab:supp_retrain_familiar}
\small
\setlength{\tabcolsep}{2.5pt}
\begin{tabular}{@{}l cccc cccc@{}}
\toprule
 & \multicolumn{4}{c}{Straight} & \multicolumn{4}{c}{Turn} \\
\cmidrule(lr){2-5}\cmidrule(l){6-9}
Training exposure & All & $\geq$2 & $\geq$4 & $\geq$6 & All & $\geq$2 & $\geq$4 & $\geq$6 \\
\midrule
Latent TransFuser \cite{chitta2023transfuser} & 3.8 & 4.5 & 4.7 & 5.5 & 10.6 & 12.5 & 14.3 & 14.6 \\
ReCogDrive IL \cite{li2026recogdrive} & 4.5 & 4.7 & 4.1 & 5.7 & 8.5 & 10.7 & 13.2 & 15.8 \\
DrivoR \cite{kirby2026drivor} & 2.0 & 2.4 & 3.3 & 3.7 & 8.2 & 9.6 & 10.3 & 10.7 \\
RoadVLA-0 & 4.2 & 4.7 & 5.4 & 6.1 & 10.6 & 11.9 & 13.6 & 16.0 \\
\midrule
RoadVLA-SAY & 4.4 & 4.8 & 5.4 & 5.8 & 11.1 & 12.6 & 14.2 & 16.5 \\
RoadVLA-KNOW & 5.2 & 5.8 & 6.3 & 6.7 & 9.0 & 10.4 & 12.3 & 14.8 \\
\midrule
Human agent & 0.4 & 0.5 & 0.7 & 0.7 & 1.0 & 1.2 & 1.8 & 2.2 \\
\bottomrule
\end{tabular}
\end{table}

\paragraph{Longer prediction horizon}
Our models predict 5\,s of trajectory, but NAVSIM's native PDMS evaluation, and with it the retraining table (\Cref{tab:retrain_leakage}), only scores the first 4\,s. We rescore the same predictions of RoadVLA-0, -SAY and -KNOW over all 5\,s with the same
EPDMS+USE rule, on the same decision points to test whether the turn gap widens. For comparability, straight and turning samples are split by the recorded 4\,s path as before.
\Cref{tab:supp_retrain_5s} reports the mean EPDMS+USE drop over the three weightings. The drop grows at 5\,s for every model: on turns from 9.0--11.1 to 15.9--18.1 points, on straight driving from 4.2--5.2 to 6.8--8.1. Both sides score lower over the longer horizon, the held-out side more (RoadVLA-0 on turns: from 78.3 to 71.1 on NAVTEST, from 68.6 to 55.1 on the held-out zones). The extra drop comes from EPDMS, not from USE, with which 98--99\% of samples comply at either
horizon. On the held-out zones, drivable-area compliance falls by 7.1--7.6 points, while on NAVTEST it falls by 2.7--3.1. The last second leads further along the road ahead, and there GEOTRAIN models leave the drivable area far more often than NAVTRAIN ones.

\begin{table}[t]
\centering
\caption{Drop from the NAVTRAIN model on NAVTEST to the GEOTRAIN model on the held-out GEOTEST zones (difference in EPDMS+USE), with the
same predictions scored over NAVSIM's 4\,s horizon (as in \Cref{tab:retrain_leakage}) and over all 5\,s our models predict.
Mean over raw, equal-city and NAVTEST-city weighting.}
\label{tab:supp_retrain_5s}
\small
\setlength{\tabcolsep}{5pt}
\begin{tabular}{@{}l cc cc@{}}
\toprule
 & \multicolumn{2}{c}{Straight} & \multicolumn{2}{c}{Turn} \\
\cmidrule(lr){2-3}\cmidrule(l){4-5}
Horizon & 4\,s & 5\,s & 4\,s & 5\,s \\
\midrule
RoadVLA-0 & 4.2 & 6.8 & 10.6 & 16.7 \\
RoadVLA-SAY & 4.4 & 7.2 & 11.1 & 18.1 \\
RoadVLA-KNOW & 5.2 & 8.1 & 9.0 & 15.9 \\
\bottomrule
\end{tabular}
\end{table}

\paragraph{Counting every failed term}
\Cref{fig:failures} attributes each failing sample to the first term at zero, in the order DAC, DDC, NC, TLC, USE, so a sample that both leaves the drivable area and collides counts only as a DAC failure. If this order hid additional rule failures on unseen roads behind DAC, counting every term at zero would reveal them. On turns, the share of held-out samples failing NC exceeds NAVTEST by 0.4--1.5 points across the six models (0.3--0.9 as first failure), TLC by $-$0.2 to $-$0.1 and USE by $-$0.1 to 0.5, against 5.6--9.6 points for DAC, which comes first and is therefore identical under both counts. DDC, the second geometric term, exceeds NAVTEST by 0.7–2.2 points (0.2–1.3 as first failure). On straight driving, no term other than DAC differs by more than 0.8 points (DAC: 1.9--4.9). Interestingly, DrivoR \cite{kirby2026drivor} fails differently from the other models: on NAVTEST it leaves the drivable area on only 1.5\% of turns (10.4--14.2\% for the others), but fails driving direction (1.8\% \vs 0.4--0.6\%) and USE (2.1\% \vs 0.8--0.9\%) more often, running stop signs far more often than other policies. Its additional failures on unseen roads are nevertheless geometric as well, with DAC rising by 7.5 points on turns.

\section{External VQA baselines and human study}
\label{sec:external_baselines_prompt}

\paragraph{External VQA inference}
We evaluate external VQA baselines in \Cref{tab:vqa_investigation} using a shared multiple-choice prompt. Each model receives the camera views, ego-motion history and navigation context shown in \Cref{fig:roadvla_prompt}. The prompt also includes the question-interpretation instructions under ``How the questions are meant'' in \Cref{fig:qc_instructions}, along with a single question and its answer options. We request a single option letter with the instruction \texttt{Task: answer the question with the letter of the correct option only}, and generate responses without constraining decoding to the option letters.
We then score external VQA responses with a shared deterministic parser. After removing completed thinking blocks and normalizing formatting, the parser extracts a single choice from an explicit option letter or uniquely matching option text. Extraction uses the question and displayed options without access to the reference answer. Responses that yield no valid choice or trigger the parser's conflict checks count as incorrect and remain in the denominator (less than 1.2\% for all evaluated baselines). For natural and balanced set, we average accuracy over all 39 query templates.

\paragraph{Human KNOW study}
We measure human accuracy on KNOW VQA with two raters on 500 QA pairs, 250 drawn from the natural and 250 from the balanced set (see Appendix \cref{sec:nat_bal_stats}), covering all 39 templates. The two sets are interleaved in one fixed order, and a question's set provenance is not
shown. Raters see what the model receives: the left, front and right cameras, the navigation instruction, the question and the same answer options as the model. The earlier frames (T$-$1.5\,s to T$-$0.5\,s) are available as a clip. As with the quality checks in Appendix \cref{sec:quality_checks}, the human scoring was carried out by two members of our research team, who took part voluntarily, knew how their scores would be used, and agreed to their use. No personal information about the participants was collected or reported.
Both raters score the same on the two sets, so the mean is 93.8 on each
and the human natural--balanced gap is 0.0 (95\% CI $\pm$3.7; intervals over questions, using the per-question mean of the two raters). Unlike the models in \Cref{tab:vqa_investigation}, humans gain nothing from the natural set's skewed answers. On 18 questions (3.6\%) neither of the raters answered correctly; on 17 of these both raters chose the same alternative. We take these 17 (3.4\%) as likely label errors, which bounds attainable accuracy at about 97\% and aligns with what our quality checks found (\cf Appendix \cref{sec:quality_checks}).

\section{Additional details on scaling experiments}
\label{sec:place_scaling_details}

\begin{figure*}[t]
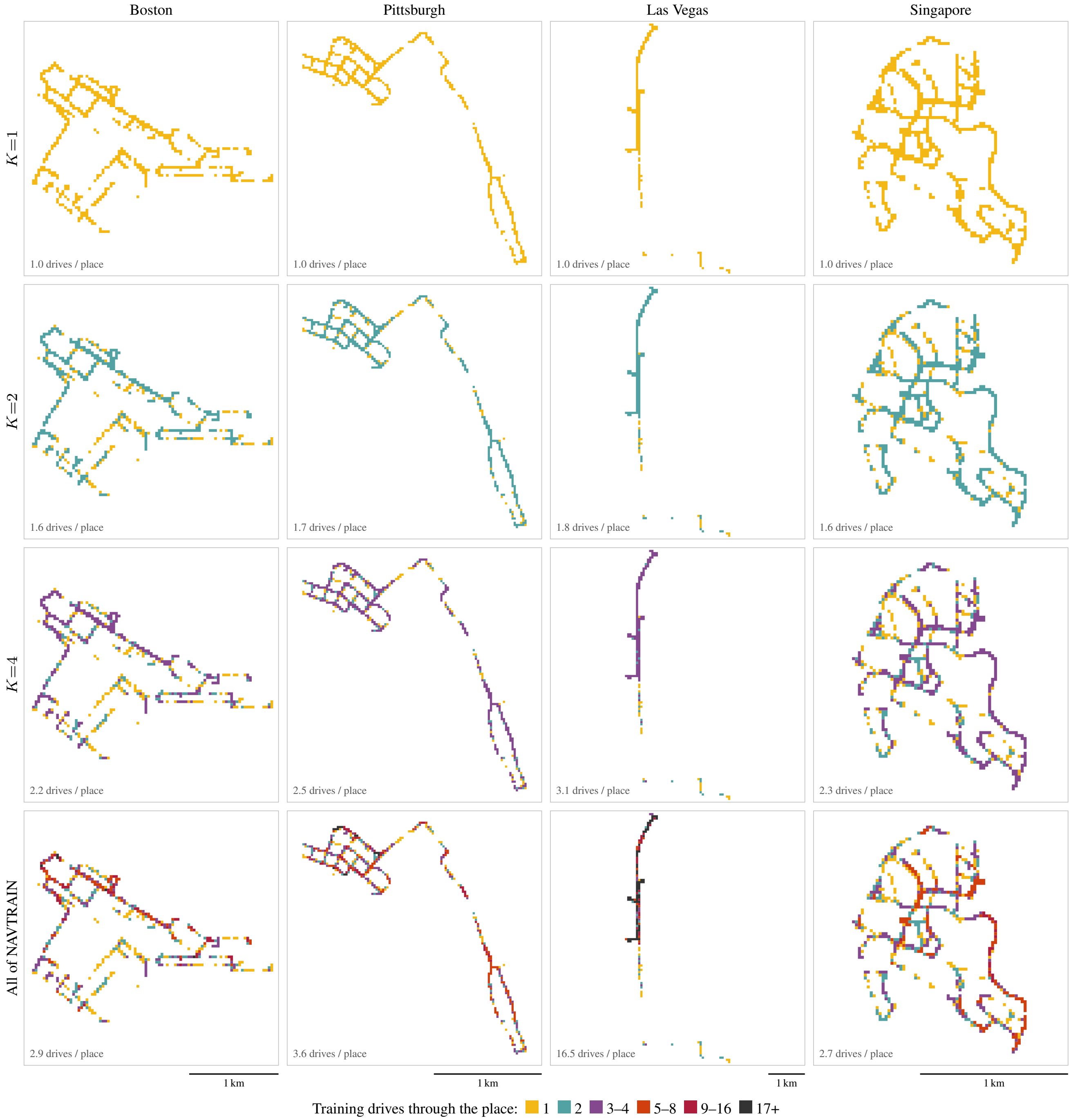

\centering
\providecommand{\scalingareasdir}{supplementary/figures/scaling_areas}%
\providecommand{\scalingareaswidth}{\linewidth}%
\colorlet{sca1}{_yellow}\colorlet{sca2}{_teal}\colorlet{sca3}{_purple}%
\colorlet{sca4}{_orange}\colorlet{sca5}{_red}\colorlet{sca6}{black!80}%
\tikzset{sca bin 1/.style={fill=sca1}, sca bin 2/.style={fill=sca2}, sca bin 3/.style={fill=sca3},
    sca bin 4/.style={fill=sca4}, sca bin 5/.style={fill=sca5}, sca bin 6/.style={fill=sca6},
    sca frame/.style={draw=black!25, line width=0.3pt},
    sca note/.style={font=\tiny, text=black!65, fill=white, fill opacity=0.8, text opacity=1, inner sep=1pt}}%
\def\scaset#1#2#3{\expandafter\def\csname sca@#1@#2\endcsname{#3}}%
\def\scaget#1#2{\csname sca@#1@#2\endcsname}%
\input{\scalingareasdir/meta}%
\begingroup
\def\scalabel{16pt}\def\scagap{4pt}%
\pgfmathsetlengthmacro{\scapanel}{(\scalingareaswidth - \scalabel - 3*\scagap) / 4}%
\pgfmathsetlengthmacro{\scabottom}{-(4*\scapanel + 3*\scagap)}%
\begin{tikzpicture}[font=\scriptsize]
\foreach \city/\cname [count=\j from 0] in {boston/Boston, pittsburgh/Pittsburgh, lasvegas/Las Vegas, singapore/Singapore}{
    \pgfmathsetlengthmacro{\scax}{\scalabel + \j * (\scapanel + \scagap)}
    \pgfmathsetlengthmacro{\scaunit}{\scapanel / \scaget{\city}{n}}
    \node[anchor=base, inner sep=0pt] at (\scax + 0.5 * \scapanel, 3pt) {\cname};
    \foreach \set/\sname [count=\i from 0] in {k1/{$K{=}1$}, k2/{$K{=}2$}, k4/{$K{=}4$}, all/{All of NAVTRAIN}}{
        \pgfmathsetlengthmacro{\scay}{-(\i + 1) * \scapanel - \i * \scagap}
        \begin{scope}[shift={(\scax, \scay)}, x=\scaunit, y=\scaunit]
            \input{\scalingareasdir/\city_\set}
        \end{scope}
        \draw[sca frame] (\scax, \scay) rectangle ++(\scapanel, \scapanel);
        \node[sca note, anchor=south west] at (\scax + 1.5pt, \scay + 1.5pt) {\scaget{\city-\set}{mean} drives / place};
        \ifnum\j=0
            \node[rotate=90, anchor=south, inner sep=1pt] at (\scalabel - 2pt, \scay + 0.5 * \scapanel) {\sname};
        \fi
    }
    \pgfmathsetlengthmacro{\scakm}{\scaget{\city}{km} * \scaunit}
    \draw[line width=0.6pt] (\scax + \scapanel - \scakm, \scabottom - 4pt) -- ++(\scakm, 0)
        node[midway, below=1pt, font=\tiny, inner sep=1pt] {1\,km};
}
\node[anchor=north, inner sep=0pt] at (\scalabel + 2 * \scapanel + 1.5 * \scagap, \scabottom - 16pt) {%
    Training drives through the place:\enspace
    \tikz\fill[sca1] (0,0) rectangle (6pt,6pt);~1\enspace
    \tikz\fill[sca2] (0,0) rectangle (6pt,6pt);~2\enspace
    \tikz\fill[sca3] (0,0) rectangle (6pt,6pt);~3--4\enspace
    \tikz\fill[sca4] (0,0) rectangle (6pt,6pt);~5--8\enspace
    \tikz\fill[sca5] (0,0) rectangle (6pt,6pt);~9--16\enspace
    \tikz\fill[sca6] (0,0) rectangle (6pt,6pt);~17+};
\end{tikzpicture}%
\endgroup
\caption{The repeated-visitation training sets on NAVTRAIN. Rows: at most $K=1, 2, 4$ drives per place, and all of NAVTRAIN. All sets cover the same 6{,}328 places; they differ only in how often each place is driven. For cells that have been driven in multiple heading directions, we plot the busiest place.}
\label{fig:supp_scaling_areas}
\end{figure*}

For the scaling experiments in \Cref{sec:place_scaling}, we define a place as a city, a 10\,m grid cell of the ego pose and one of four 90$^\circ$ heading bins; a drive is one recorded log. 
For each place we list its drives, shuffle the list once with a fixed seed and keep the first $K$ drives, together with all of their decision points at that place. Because the shuffle is shared, the sets are nested ($K{=}1 \subset K{=}2 \subset K{=}4$), and every set keeps every place; a place with fewer than $K$ drives keeps all of them, so the mean stays below $K$. On NAVTRAIN, $K=1, 2, 4$ keep 14{,}083, 23{,}965 and 37{,}321 decision points, with 1.00, 1.65 and 2.47 drives per place; on GEOTRAIN, 11{,}820, 20{,}581 and 33{,}428, with 1.00, 1.69 and 2.67 drives per place. \Cref{fig:supp_scaling_areas} shows the NAVTRAIN sets, where the number of drives through each place grows with $K$.

\paragraph{Matching the sample count}
Smaller $K$ leaves fewer decision points, so we fill the $K{=}1$ and $K{=}2$ sets up to the sample count of $K{=}4$ with extra samples at the same decision points. Each extra sample repeats the decision point's trajectory turn and asks a KNOW question it has not been asked yet, so $K{=}1$ trains on 2.7 samples per decision point on NAVTRAIN (2.8 on GEOTRAIN) and $K{=}2$ on 1.6. The extra questions are assigned greedily: visiting the decision points in a seeded random order, each extra sample takes, among the queries that apply there and have not been asked there, the one furthest below its target relative to the target, where the targets are the per-query question counts of the $K{=}4$ set. All three sets thus have the same number of samples and, as far as the decision points allow, the same number of questions per query. The runs of one experiment are compared at the same optimizer step with the global batch of 64 and 2{,}775 steps. As a reference, the RoadVLA-KNOW run on all of NAVTRAIN or GEOTRAIN (5 epochs, 8k steps) is shown after an axis break in \Cref{fig:place_scaling}, since it differs in both samples and steps. The two panels plot corresponding runs trained on NAVTRAIN and GEOTRAIN, respectively, using different horizontal-axis quantities: in (a), the number of training drives through the test place, which equals $K$ since every test place there is driven at least four times; in (b), where test places are unseen, the mean number of drives per training place under the cap $K$ (1.0, 1.7 and 2.7). The $K{=}4$ run thus appears at 4 in (a) and at 2.7 in (b).

\paragraph{Geographic diversity}
The diversity sets have as many decision points as $K{=}1$ on GEOTRAIN (11{,}820) but draw them from fewer places. Within each city, we take places one after another, with all decision points of all their drives, until every turn class of the recorded path (slow, straight, bend or turn) holds at least as many decision points as in $K{=}1$. The places come in a seeded random order for the set with 1{,}430 places (3.9 drives per place), and from the most-driven place downward for the set with 636 places (9.4 drives per place). From the decision points taken this way, we then sample at random exactly $K{=}1$'s count per city and turn class, so all three sets agree in city mix and turn mix and differ only in how many places they spread over. All three are filled with extra questions in the same way as above.

\end{document}